\documentclass[twoside,11pt]{article}

\usepackage{jmlr2e}

\usepackage{microtype}

\usepackage{amsmath,amsfonts,bm,bbm}

\newcommand{\bigO}{\mathcal{O}}
\newcommand{\RR}{\mathbb{R}}

\def\eqref#1{equation~\ref{#1}}

\def\ceil#1{\lceil #1 \rceil}
\def\floor#1{\lfloor #1 \rfloor}
\def\1{\bm{1}}

\DeclareMathAlphabet{\mathsfit}{\encodingdefault}{\sfdefault}{m}{sl}
\SetMathAlphabet{\mathsfit}{bold}{\encodingdefault}{\sfdefault}{bx}{n}

\def\gL{{\mathcal{L}}}

\def\sR{{\mathbb{R}}}

\newcommand{\R}{\mathbb{R}}

\usepackage{ragged2e}
\usepackage{hyperref}
\usepackage{url}

\usepackage{caption}
\usepackage{wrapfig}
\usepackage{subcaption}

\usepackage{xargs}                                      
\usepackage{xcolor}                  
\usepackage{enumitem}
\usepackage{array}
\usepackage{tabularx}
\usepackage{booktabs}
\usepackage{placeins}

\usepackage{tikz}
\def\checkmark{\tikz\fill[scale=0.4](0,.35) -- (.25,0) -- (1,.7) -- (.25,.15) -- cycle;} 
\usepackage{makecell}

\ShortHeadings{Insights into Ordinal Embedding Algorithms}{C Vankadara, Haghiri, Lohaus, Ul Wahab, and von Luxburg}
\firstpageno{1}
\begin{document}

\title{Insights into Ordinal Embedding Algorithms: \\ A Systematic Evaluation}

\author{\name Leena C Vankadara \email leena.chennuru-vankadara@uni-tuebingen.de \\
     \addr University of T{\"u}bingen
       
       \AND
       \name Michael Lohaus \email michael.lohaus@uni-tuebingen.de \\ 
        \addr University of T{\"u}bingen
     
     \AND
       \name Siavash Haghiri \email siyavash.haghiri@gmail.com \\
       \addr University of T{\"u}bingen
    
    \AND
       \name Faiz Ul Wahab \email fwahhab89@gmail.com \\
     \addr University of T{\"u}bingen
     
     \AND
       \name Ulrike von Luxburg \email ulrike.luxburg@uni-tuebingen.de \\
     \addr University of T{\"u}bingen
       }
       

\editor{tba}

\maketitle

\begin{abstract}
The objective of ordinal embedding is to find a Euclidean representation of a set of abstract items, using only answers to triplet comparisons of the form ``Is item $i$ closer to the item $j$ or item $k$?''. In recent years, numerous algorithms have been proposed to solve this problem. However, there does not exist a fair and thorough assessment of these embedding methods and therefore several key questions remain unanswered: Which algorithms perform better when the embedding dimension is constrained or few triplet comparisons are available? Which ones scale better with increasing sample size or dimension? In our paper, we address these questions and provide the first comprehensive and systematic empirical evaluation of existing algorithms as well as a new neural network approach. 
We find that simple, relatively unknown, non-convex methods consistently outperform all other algorithms, including elaborate approaches based on neural networks or landmark approaches. This finding can be explained by our insight that many of the non-convex optimization approaches do not suffer from local optima. Our comprehensive assessment is enabled by our unified library of popular embedding algorithms that leverages GPU resources and allows for fast and accurate embeddings of millions of data points. 
\end{abstract}


\section{INTRODUCTION}
We investigate the problem of representation learning given a set of items $X = \left \{x_1, x_2, \cdots, x_n \right \}$ with \textbf{neither an explicit input representation} nor a \textbf{similarity function} over pairs of items. Instead, we assume that we are provided with a set of triplet comparisons $(i,j,k)$, which encode the relationship that item $x_i$ is closer to item $x_j$ than to item $x_k$. A whole sub-community is dedicated to machine learning based on such triplet comparisons~\citep{heikinheimo2013crowd,kleindessner2014uniqueness,amid2015multiview,kleindessner2015dimensionality, balcan2016Learning,haghiri2017comparison,haghiri2018comparison, kleindessner2017kernel}. One approach that enables the application of machine learning methods to such data is through ordinal embedding. 
Formally, given a set of such triplet comparisons $\mathcal{T} =~ \{t_1,t_2,\ldots t_m\}$, the goal of ordinal embedding is to find a $d$-dimensional Euclidean representation $y_1,y_2, \ldots y_n \in \sR^d$ which satisfies as many triplet comparisons as possible. This can be expressed by the following optimization problem:
\begin{align}
\label{eqn:OEobjective}
\min_{y_1,\ldots, y_n \in \RR^d}{\sum_{t=(i,j,k)\in T}{\mathbbm{1}_{\Vert y_i - y_j \Vert^2 > \Vert y_i - y_k \Vert^2}}},
\end{align}
where $\mathbbm{1}_{E}$ is the indicator function, which equals one, if the condition $E$ is true and zero otherwise. The Euclidean representation of items, obtained from Ordinal embedding (OE) can, for instance, subsequently be used as input to classical machine learning approaches.

\subsection{Related work: What is missing?}
Many OE approaches have been proposed in the literature so far \citep{agarwal2007generalized, van2012stochastic, heikinheimo2013crowd, terada2014local, jain2016finite, anderton2019scaling, ghosh2019landmark}. In Table~\ref{tab:summary_of_methods}, we summarize the existing literature on Ordinal Embedding algorithms. Most of these algorithms are devoid of any theoretical support and, moreover, a thorough empirical assessment of these methods is completely missing from the literature. Existing empirical evaluations of these approaches are quite limited---from the sample sizes used to the number of algorithms evaluated. As shown in Table~\ref{tab:summary_of_methods}, none of the existing evaluations constitute more than three different algorithms and no experiments with ordinal embedding algorithms have been reported for more than $\sim 10,000$ items.
\newcolumntype{C}[1]{>{\centering\let\newline\\\arraybackslash\hspace{0pt}}m{#1}}
\begin{table*}[t]
    \centering
    {\renewcommand{\arraystretch}{1}
    \begin{tabular} { C{1.9cm} C{1.9cm} C{3cm} C{3.7cm} C{2.5cm}  } 
        {\small\bf Algorithm} & {\small\bf Compared Against} & {\small\bf Dataset Attributes} $(n, dim_{in}, dim_{out})$ & {\small\bf Evaluation Criteria} & {\small\bf Platform} \\ 
        \toprule
        \small{GNMDS} & \small{None} & \small{($100$, -, 2)} & \small{Visualization} & \small{-} \\ 
        \hline
        \small{CKL, CKLx} & \small{None} & \small{($500$, -, -)} & \small{Accuracy on a ranking task, Classification error, Visualization} & \small{-} \\ 
        \hline
        {\small{(t-)STE}} & \small{CKL, GNMDS} & \small{($5000$, $784$, $2$)} & \small{Triplet error, Nearest Neighbor error} & \small{MATLAB} \\ 
        \hline
        \small{SOE} & \small{GNMDS} & \small{($5000$, $2$, $2$)} & \small{Visualization} & \small{R} \\ 
        \hline
        \small{FORTE} & \small{None} & \small{($100$, $8$, $8$)} & \small{Triplet error} & \small{Python} \\ 
        \hline
        \small{LSOE} & \small{SOE, t-STE} & \small{($16000$, $30$, $30$)} & \small{Triplet error, Procrustes error, Nearest Neighbor error, CPU running time} & \small{C++} \\ 
        \hline
        \small{LLOE} & \small{None} & \small{($10^6$, $2$, $2$)} & \small{Same as LSOE} & \small{C++} \\ 
        \hline
        \small{LOE} & \small{GNMDS, STE} &  \small{($10^5$, $2$, $2$)} & \small{Procrustes error, CPU running time} & \small{MATLAB} \\ 
    \end{tabular}}
    \caption{Summary of the literature on Ordinal Embedding algorithms. We list details of the evaluation of each algorithm as follows: the competitors, the evaluation criteria, the maximum size of the datasets used, and the implementation platform/language used in the respective original publication. 
    }
    \label{tab:summary_of_methods}
    \vspace{-0.15in}
\end{table*}

One of the primary hurdles to a thorough evaluation stems from the issue of computational complexity of ordinal embedding: it is NP-Hard in the worst case~\citep{bower2018landscape}. In nearly all of the OE algorithms, first-order approaches are used to solve either a non-convex optimization problem or a semi-definite program with respect to either the embedding matrix $X \in \R^{n \times d}$ or the Gram matrix $K \in \R^{n \times n}$, where $n, d$ denote the number of samples and embedding dimension respectively. With a train triplet set $\mathcal{T}$ the computational complexity of \textit{each gradient update} scales as $\bigO(d |\mathcal{T}|)$ when the optimization is with respect to $X$ and as $\bigO(n |\mathcal{T}|)$ for optimization with respect to $K$. \citet{jain2016finite} show that at least $\Omega(n d \log n)$ \textit{actively chosen} triplets are needed to reconstruct the original embedding up to similarity transformations. Therefore, the computational complexity of each gradient update scales as $\bigO(n d^2 \log n)$ or as $\bigO(n^2 d \log n)$ for optimization with respect to $X$ and $K$ respectively. 

As a consequence, the computational complexity of OE algorithms quickly becomes prohibitive with increasing $n$ or $d$. There have been recent attempts to overcome the issue of scalability by asking for active triplet comparisons \citep{anderton2019scaling, ghosh2019landmark}, for example with respect to a fixed set of landmarks. However, a fair and comprehensive evaluation of these approaches is still missing in the literature as evidenced by Table~\ref{tab:summary_of_methods}. Due to lack of such a thorough and comprehensive evaluation, several key questions regarding the performance of ordinal embedding algorithms, in various constrained settings, remain unaddressed. 

\subsection{Our Contributions.}
\begin{itemize}
    \item \textbf{A First Comprehensive Evaluation.} In this paper, we provide the first systematic and thorough evaluation of all the existing ordinal embedding algorithms, as well as a new approach based on neural networks, across a spectrum of real and synthetic datasets using various evaluation criteria.
    
    \item \textbf{GPU-supported Library.} Our comprehensive assessment is enabled by our GPU-supported library containing standardized implementations of all the popular embedding algorithms that allows fast and accurate embeddings of millions of data points (or hundreds of millions of triplets).
    
    \item \textbf{Address Key Questions.} In our evaluation, we address several key questions concerning the performance of different OE algorithms. In particular, we focus on \textit{optimization hurdles} to non-convex OE methods, \textit{scalability} with respect to sample size or input dimension, and on performance of the methods under various \textit{constrained settings}, for example, noisy triplets, small embedding dimension or lower numbers of available triplets.
    
\end{itemize}
Our empirical evaluation reveals several interesting phenomena. We briefly present our most interesting findings here. 
\begin{itemize}
    \item Across a spectrum of experiments, we find that simple algorithms that employ gradient-based optimization of non-convex objectives directly over the embedding matrix consistently outperform the neural network, more elaborate landmark-based approaches as well as algorithms based on convex-relaxations or those that optimize over the gram matrix in terms of both the quality of the embedding as well as computation time. 
     \item In the low noise regime, the relatively unknown, soft-ordinal embedding algorithm \citep{terada2014local} consistently matches or outperforms all the other algorithms in faithfully reconstructing the original data.
     \item In addition, unlike the rest of the algorithms, performance of the soft-ordinal embedding algorithm remains largely unaffected by increasing sample size or the ambient dimension of the underlying data. 
    \item These findings can be explained by another interesting insight from our analysis: simple non-convex OE algorithms attain high-quality local optima. In particular, the optimization objective of the soft-ordinal embedding  algorithm consistently obtains the globally optimal solution.
\end{itemize}
\section{ORDINAL EMBEDDING APPROACHES}
\label{sec:ordinal_approaches}
Passive OE approaches admit any arbitrary set of triplets to generate an embedding of the items, often by optimizing a loss function to find an embedding that satisfies a given set of triplet constraints. Active approaches, on the other hand, require access to an oracle through which the algorithm can actively query any triplet comparisons. In this section, we discuss the existing ordinal embedding algorithms. In the appendix, we present them in greater detail. 

\subsection{Optimization over Gram Matrix.} Generalized Non-metric Multi-dimensional Scaling (GNMDS, \cite{agarwal2007generalized}), Crowd Kernel Learning (CKL, \cite{tamuz2011adaptively}), and Fast Ordinal Triplet Embedding (FORTE, \cite{jain2016finite}) consider optimization objectives over the Gram matrix $K$ with the following constraints: $K$ is positive semi-definite and has rank equal to that of the embedding dimension d. The final embedding $X$ of all items can then be obtained by a Cholesky decomposition of the optimal $K$.

\paragraph{GNMDS \citep{agarwal2007generalized}.}
Given a set of triplets $(i, j, k)$ GNMDS proposes a convex and semi-definite program (SDP) over the gram matrix $K$ of items which can be reformulated as in (\ref{eq:gnmds_quaraplet}).

\begin{equation}
\label{eq:gnmds_quaraplet}
    \begin{aligned}
     \min \limits_{K}  &\sum \limits_{(i, j, k)} \max \left \{0,  1 + K_{k,k} - 2 K_{k, i} + K_{i, i} - K_{i, i} + 2K_{i, j} - K_{j, j} \right \} + \lambda \textrm{Trace}(K) \\
     &\textrm{subject to } K \succeq 0.
    \end{aligned}
\end{equation}
 For better intuition for the optimization objective, it is helpful to write this objective as a function of $X$ (an embedding of items). As a result we see that the optimization objective corresponds to a $l_2$ regularized hinge loss over $X$, where $\lambda$ is the regularization parameter.
\begin{equation*}
    \sum \limits_{(i, j, k)} \max \left \{0,  1 + \vert \vert x_i - x_k \vert \vert^2 - \vert \vert x_i - x_j \vert \vert^2  \right \} + \lambda \sum \limits_{i} \vert \vert x_i \vert \vert^2
\end{equation*}

\paragraph{CKL \citep{tamuz2011adaptively}.}
 The optimization problem of CKL is defined based on a generative probabilistic model on triplets. The probability of triplets is defined based on the gram matrix $K$ of items. Let $\delta_{i,j} = \vert \vert x_i - x_j \vert \vert^2 = K_{i, i} + K_{j, j} - 2 K_{i, j}.$ Then the probability of a triplet being satisfied is given by (\ref{eq:ckl_prob}).
\begin{equation}
    \label{eq:ckl_prob}
    p_{i, j, k} = \frac{\mu + \delta_{i, k}}{2 \mu + \delta_{i, k} + \delta_{i, j}}.
\end{equation}
CKL proposes a non-convex, constrained optimization problem given in (\ref{eq:ckl_quaraplet}). 
 \begin{equation}
\label{eq:ckl_quaraplet}
    \begin{aligned}
     & \min \limits_{K}  \sum \limits_{(i, j, k)} p_{i, j, k} \textrm{ subject to } K \succeq 0.
    \end{aligned}
\end{equation}
For both GNMDS and CKL, we solve the optimization problems over $K$ iteratively using ADAM (to minimize the loss) followed by a projection on to the space of positive semi-definite matrices of rank $d$. 

\paragraph{FORTE, \citep{jain2016finite}.} FORTE also proposes a non-convex, constrained optimization problem given in (\ref{eq:forte_quaraplet}).
 \begin{equation}
\label{eq:forte_quaraplet}
    \begin{aligned}
     & \min \limits_{K}  \sum \limits_{(i, j, k)} \log (1 + \exp{(\delta_{i,j} - \delta_{i,k})}) \textrm{ subject to } K \succeq 0.
    \end{aligned}
\end{equation}
To optimize this objective, we use the Rank-d Projected Gradient Descent (PGD) with line search which is the optimization method prescribed by the authors in \citet{jain2016finite}.


\subsection{Optimization over the Embedding.} 
In contrast to methods which optimize over the Gram matrix, these algorithms directly optimize over the embedding space thus reducing the number of parameters from $n^2$ to $nd$. 
Stochastic Triplet Embedding (STE, \cite{van2012stochastic}), t-Stochastic Triplet Embedding (t-STE, \cite{van2012stochastic}), and a variant of CKL (CKLx) are based on generative probabilistic models. The probability of a triplet is defined based on kernel values of paired items using the Gaussian kernel (STE) or the Student-t kernel (t-STE). Optimizing the log-likelihood for a given set of triplets leads to the desired embedding. 

\paragraph{(t-)STE \citep{van2012stochastic}.} Similar to CKL, (t-)Stochastic triplet embedding (STE) define a generative probabilistic model for the triplets. The log-likelihood of a given set of triplets is optimized to obtain the desired embedding. Authors also propose a variant of the model with t-student kernel between the pair of items. The optimization objectives corresponding to STE and t-STE are given in (\ref{eq:ste}) and (\ref{eq:tste}) respectively.

\begin{equation}
    \label{eq:ste}
  \max \limits_{X} \sum \limits_{(i,j,k)} \log p_{i,j,k} \textrm{, where, } p_{i,j,k} = \frac{\exp{(-\delta_{i,j})}}{\exp{(-\delta_{i,j})} + \exp{(-\delta_{i,k})}}.
\end{equation}
\begin{equation}
    \label{eq:tste}
   \max \limits_{X} \sum \limits_{(i,j,k)} \log p_{i,j,k} \textrm{, where, } p_{i,j,k} = \frac{(1 +\frac{ \delta_{i,j}}{\alpha})^{\frac{-(\alpha + 1)}{2}}}{(1 +\frac{ \delta_{i,j}}{\alpha})^{\frac{-(\alpha + 1)}{2}} + (1 + \frac{\delta_{i,k}}{\alpha})^{\frac{-(\alpha + 1)}{2}}}.
\end{equation}
 In our implementation of the t-STE algorithm, consistent with the authors' recommendation, we set $\alpha$ as $d-1$, where $d$ is the embedding dimension. To solve both of these non-convex, unconstrained optimization problems, we use ADAM.

\paragraph{Soft ordinal embedding \citep{terada2014local}.}
Another successful approach is the Soft Ordinal Embedding method. The authors use an optimization objective based on the hinge triplet margin loss to satisfy the set of input triplet comparisons and the corresponding optimization problem is given in (\ref{eq:soe}).
\begin{equation}
    \label{eq:soe}
    \min \limits_{X} \sum \limits_{(i,j,k)} \max \left \{0, 1 + \sqrt{\delta_{i,j}} - \sqrt{\delta_{i,k}}\right \}
\end{equation}
The optimization problem is non-convex and consistent with the rest of the algorithms, we use ADAM for optimization.


\subsection{Landmark Approaches.}
There have been two recent landmark-based approaches that use active triplet queries in order to provide a large-scale OE algorithm. 

\paragraph{Large-scale landmark ordinal embedding (LLOE, \citet{anderton2019scaling}).}

Large-scale Landmark Ordinal Embedding constitutes of two phases. The first phase aims to embed a subset $(m)$ of items which serve as landmarks in the second phase of the algorithm. To obtain the embedding of this subset, a smaller set $(L)$ of items are first chosen as landmarks in Phase 1. The algorithm then queries $\bigO(L m \log m)$ triplets to find 1) orderings of all the points with respect to the distance to each landmark and 2) orderings of all landmarks with respect to the distance to each point. These triplets are then filtered through transitivity to obtain $\bigO(L m)$ triplets. SOE is then used to find an embedding of the subset $m$ of items that satisfies these $\bigO(L m)$ triplets. To choose the value of $m$, the method suggests that a successive doubling strategy is used until the loss of the SOE objective becomes larger than a given threshold. 

However, as observed in our experiments and also noted by the authors in \citet{anderton2019scaling}, the aforementioned triplet selection procedure suggested by the authors significantly increases the time for convergence of SOE compared to when triplets are chosen at random. Furthermore, our experiments show that sufficiently many randomly chosen triplets suffice for SOE to attain consistently superior performance both in absolute terms (as measured by the triplet error and the Procrustes error) as well as relative to all other methods. Therefore, we use SOE in Phase 1 of the algorithm to obtain an accurate embedding of a subset of items which are given as landmarks to the Phase 2.

Phase 2 of the algorithm is then used to embed the remaining points. To do this, for each point, anchors are chosen in a farthest first traversal order and binary search is used to insert this point into a ranking of the $m$ items. To do so, the method finds two points $p$ and $q$ for each anchor such that the point to be embedded lies in the spherical intersection of the two balls centered at the anchor and with radius equal to the distance between the anchor and $p$ and $q$ respectively. $2(d+1)$ such anchors are chosen for each point and a margin relaxed sphere intersection objective is minimized to find an embedding of each item. The full pseudo-code for the algorithm can be found in \citet{anderton2019scaling}.


\paragraph{Landmark ordinal embedding (LOE \citet{ghosh2019landmark}).} LOE first chooses $\bigO(d)$ landmark points. Then, it asks $dn\log (n)$ triplet queries to sort the distances of remaining points with respect to the landmarks. The ranking is estimated using the maximum likelihood estimator of the BTL ranking model. Using the acquired ranking and the properties of Euclidean distance matrices, it estimates the distance matrix of landmark points with respect to the full set of items. Consequently, it uses Landmark MDS~\citep{de2004sparse} to reconstruct the full Gram matrix of items, and the hence the embedding for items. 


\subsection{THE ORDINAL EMBEDDING NEURAL NETWORK}

\begin{figure}[t!]
    \centering
   \begin{subfigure}[b]{0.45\textwidth}
   \centering
         \includegraphics[width=0.7\textwidth, angle=90]{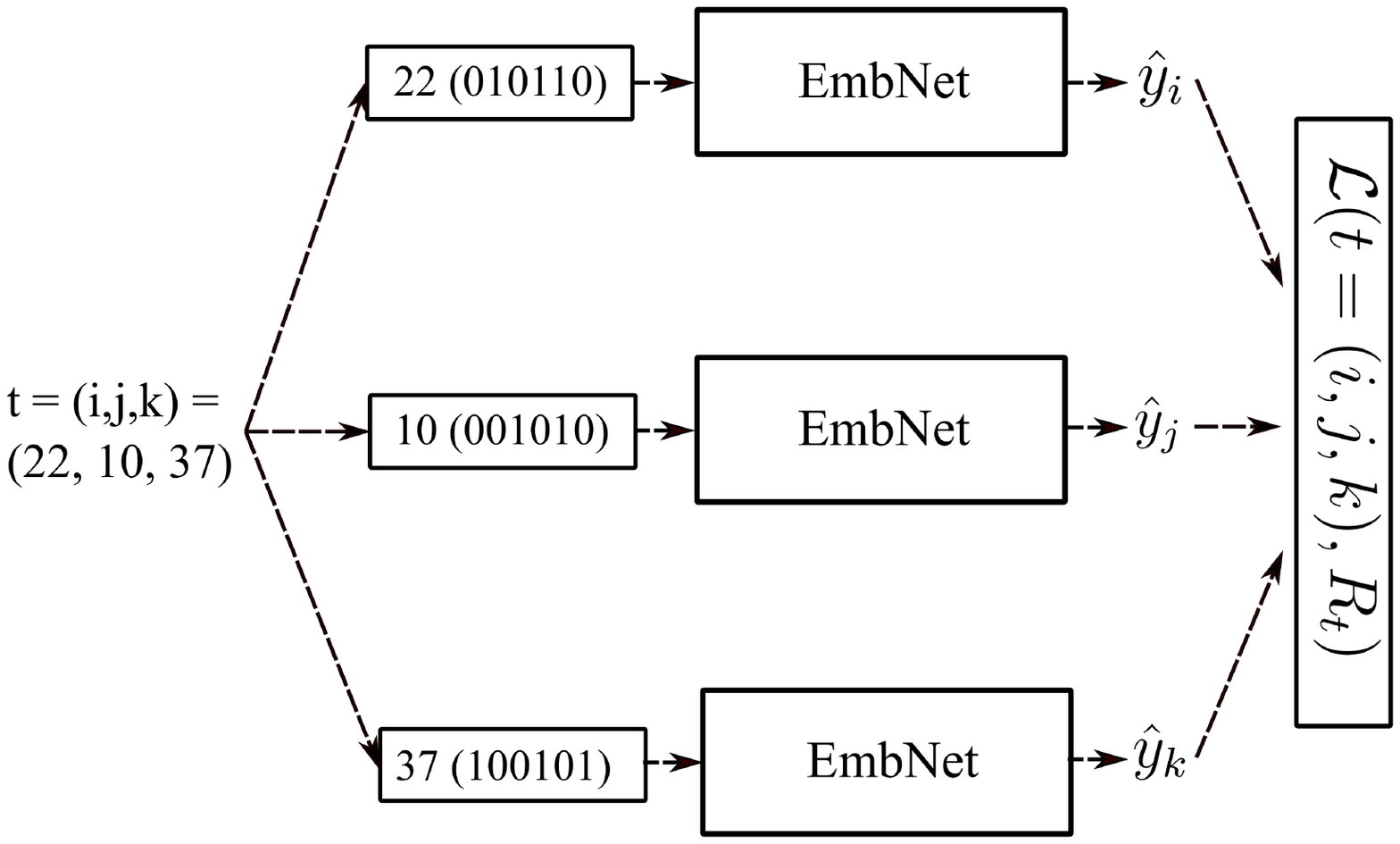}
         \caption{}
     \end{subfigure}
    ~ \hfill
    \begin{subfigure}[b]{0.45\textwidth}
         \centering
         \includegraphics[width=0.7\linewidth, angle =90]{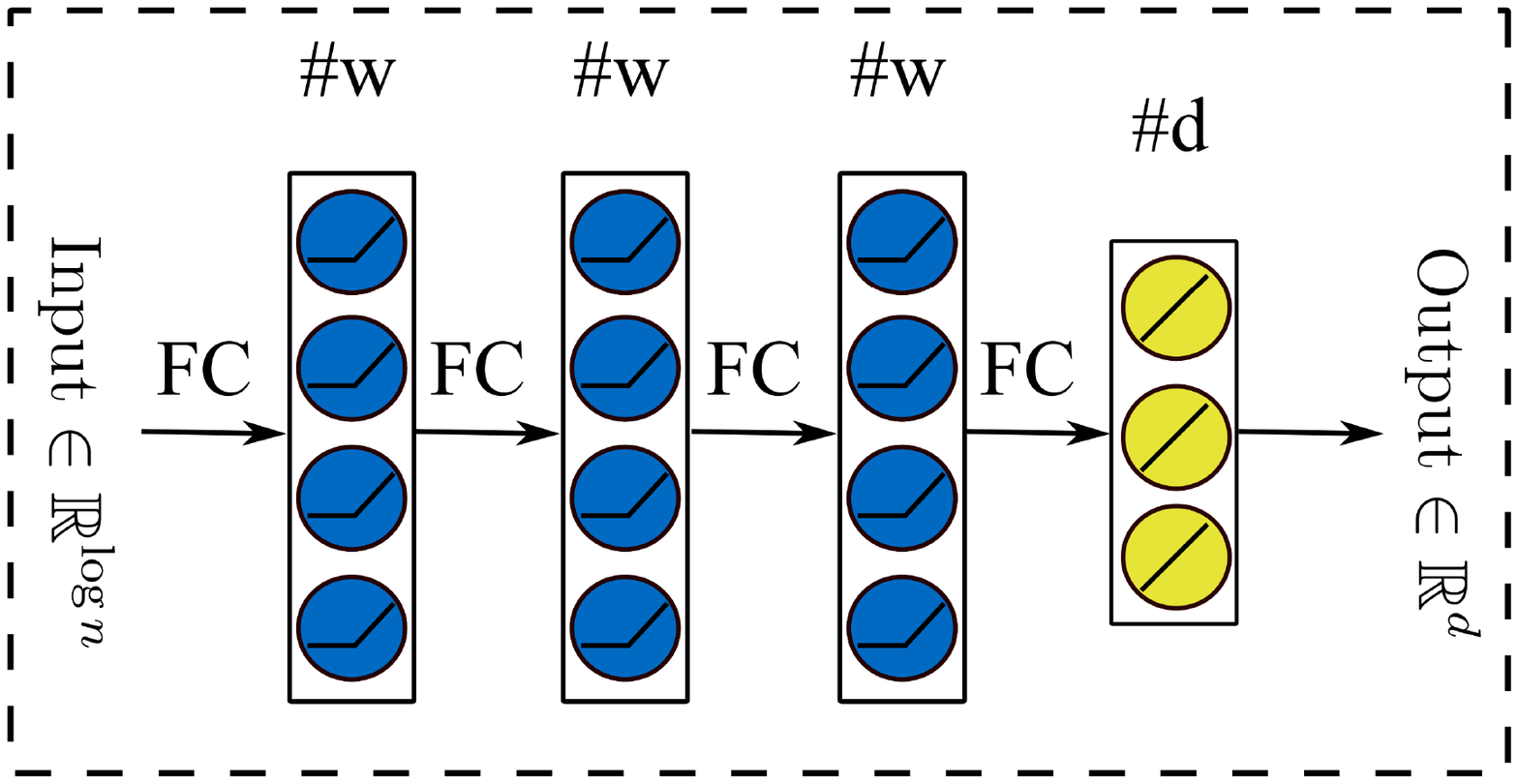}
         \caption{}
         \label{fig:EmbNet}
     \end{subfigure}
\vspace{-3mm}
\caption{The architecture of Ordinal Embedding Neural Network (OENN). An example triplet (22, 10, 37) and its answer $R_t$ are  fed to the architecture. \label{fig:NNEM}(b) The EmbNet module, which is used as a building block of the ordinal embedding architecture.
}
\vspace{-0.15in}
\end{figure}

The problem of ordinal embedding is NP-hard \citep{bower2018landscape} and naturally arising optimization objectives to OE are non-convex. It is widely believed amongst neural network practitioners that the non-convex landscape of deep and wide neural networks is primarily free of sub-optimal local minima. This theory is supported under simplified assumptions by various theoretical findings \citep{nguyen2017loss, choromanska2015loss, kawaguchi2019depth, kawaguchi2019effect}. Inspired by this line of work, we postulated that a neural network-based approach to OE could potentially overcome the issue of local optima (if it exists). Therefore, we designed a 3-layer, feedforward neural-network architecture for OE which we refer to as Ordinal Embedding Neural Network(OENN). 

\textbf{\textit{On the choice of input representation:}}
Since our abstract items are devoid of any meaningful input representation, the choice of input representations presents a challenge to this approach. We leverage the expressive power of neural networks~\citep{leshno1993multilayer,barron1993universal} and their ability to fit random labels to random inputs~\citep{zhang2016understanding} to motivate our choice of input encoding. Since our main goal is to find representations that minimize the training objective, we believe that completely arbitrary input representations are a viable choice. 

One such input representation could be the one-hot encoding of the index (where point $i$ is encoded by a string $\hat{x}_{i}$ of length $n$ such that $\hat{x}_{i}(l) = 1$ if $l = i$ and $0$ otherwise). The advantage of choosing such a representation is that it is memory efficient in the sense that there is no need to additionally store the representations of the items. However, under this choice of representation the length of the input vectors grows linearly with the number $n$ of input items. 
For computational efficiency, we consider a more efficient way: we represent each item by the binary code of its index, leading to a representation length of  $\log{n}$.  Such a representation retains the memory efficiency of the one-hot encoding (in the sense as discussed above) but improves the length of the input representation from $n$ to  $\log{n}$. As we will see below, this representation works well in practice. However, note that there is nothing peculiar about this choice of binary code. Our simulations (Subsection~\ref{sec:simWidth} in the supplementary) suggest that unique representations for items generated uniformly, randomly from a unit cube in $\mathbb{R}^{\alpha = \Omega(\log n)}$ can also be used as the input encoding without any significant effect on the performance.

\textbf{Architecture and loss.}
Figure~\ref{fig:NNEM} shows a sketch of the network architecture which is inspired from ~\cite{wang2014learning,schroff2015facenet,cheng2016person,hoffer2015deep}. The central sub-module of the architecture is what we refer to as the embedding network (\textbf{\textit{EmbNet}}), which is a three layer, fully connected neural network with Rectified linear unit (ReLU) activation functions. One such network takes a certain encoding of a single data point $x_i$ as input and outputs a $d$-dimensional representation $\hat{y_i}$ of data point $x_i$. The EmbNet is replicated three times with shared parameters. The overall OENN network now takes the \textbf{\textit{indices}} $(i,j,k)$ corresponding to a triplet $(x_i,x_j,x_k)$ as an input. It routes each of the indices $i,j,k$ to one of the copies of the EmbNet, which then return the $d$-dimensional representations $\hat{y}_i, \hat{y}_j , \hat{y}_k$, respectively (cf. Figure~\ref{fig:NNEM}). The three sub-modules are trained jointly using the triplet hinge loss, as described by the following objective function:
\begin{equation*}
    \label{eq:TripletLoss} 
    \gL(T)=  \frac{1}{\vert T \vert}\sum_{(i,j,k)\in T} {\max \left \{ \Vert \hat{y}_i - \hat{y}_j \Vert^2  - 
\Vert \hat{y}_i - \hat{y}_k \Vert^2 +1, 0 \right \}}.
\end{equation*}
\textbf{On the choice of parameters.}
 We run an extensive set of simulations to find good rules of thumb to determine the parameters of the network architecture and the input encoding. We provide the results of these simulations in the supplementary (see Section \ref{sec:method_details}).
 
\textbf{Difference to previous contrastive learning approaches.}
As described earlier, our architecture is inspired by ~\citet{wang2014learning,hoffer2015deep}. However, note that we have no access to representations for the input items $x_1,..,x_n$ and our network takes completely arbitrary representations for the input items. Additionally, in this line of work, triplets are typically contrastive, which is of the form $(x, x^+, x^-)$ which indicates that $x$ and $x^+$ belong to the same \textit{class} and $x^{-}$ is sampled from a different class. In contrast, we randomly sample triplets from the set of items.
 



\section{EXPERIMENTS AND FINDINGS}
\label{sec:experiments}
In this section, we provide the first thorough and comprehensive evaluation of all the OE algorithms presented in Section~\ref{sec:ordinal_approaches}. 
To perform this evaluation, we implemented 10 different algorithms (including OENN) with both CPU and GPU support.\footnote{Private, anonymous link to the code:\url{ https://rb.gy/kuofed}. Code will be made public on github after publication.} We conducted around 500 experiments using 15 different datasets and evaluated them using 5 different evaluation criteria.
\subsection{Experimental Setup}
\textbf{Triplet Generation.} Given a ground-truth dataset, we generate triplets based on the Euclidean distances of the data. To generate a triplet, we sample the indices $i, j, k$ of 3 items $x_i, x_j, x_k$ uniformly from the dataset and evaluate whether $ \vert \vert x_i -  x_j \vert \vert^2 < \vert \vert x_i - x_k \vert \vert^2$ or vice versa. Accordingly, either the triplet $(i,j,k)$ or $(i,k,j)$ is added to the training set. For each dataset and experiment, unless specified otherwise, we generate $\lambda n d \log n$ triplets based on Euclidean distances, where $d$ refers to the dimension of the embedding space for some $\lambda \in \mathbb{N}$. We refer to $\lambda$ as the triplet multiplier. This is a known theoretical lower bound 
on the number of \textit{active} triplets required to reconstruct the original embedding up to similarity transformations \citep{jain2016finite}. In addition, our extensive empirical investigation also reveals that $2 n d \log n$ passively generated triplets also suffice to achieve good reconstructions of the original embedding (for instance, see the experiments on increasing number of triplets in Section \ref{sec:increasing_triplets}). Therefore, unless specified otherwise, we simply set $\lambda = 2$. Active algorithms have access to an oracle which can provide any queried triplet. 

\textbf{Datasets.} We use a variety of datasets to generate training triplets for the ordinal embedding task. The datasets range from a few hundred items to millions of items and from 2 dimensions to 784 dimensions. To enable better visual evaluation of the embeddings generated by the various algorithms, we consider a variety of 2D datasets from \citet{ClusteringDatasets}. For a thorough evaluation, we also consider a range of high-dimensional real (CHAR \citep{char_covtype}, USPS \citep{usps}, MNIST \citep{mnist}, Fashion MNIST \citep{fmnist}, KMNIST \citep{kmnist}, Forest Covertype \citep{char_covtype}) and synthetic datasets (Uniform, Mixture of Gaussians). In the supplementary (\ref{sec:description_datasets}) we summarize the properties of the datasets in detail and describe in which experiment each of the datasets is used. 

\textbf{Algorithms and Implementation.} Our library consists of all nine OE methods summarized in Table~\ref{tab:summary_of_methods} and the new OENN method. We implement all algorithms in Python 3.7 using the PyTorch library~\citep{NEURIPS2019_9015}, which allows us to use GPU resources for training. We describe our hardware settings in the supplementary (Section \ref{sec:experimental_setup}). We optimize all methods with a stochastic gradient descent using adaptive learning rates (ADAM~\cite{kingma2014adam}), with the exception of FORTE, which uses a line-search based on the implementation by~\cite{jain2016finite}. The parameter selection for each method is described in the supplementary (Section \ref{sec:experimental_setup}).  

%

\textbf{Evaluation.} In our experiments, we use the following criteria to evaluate the embeddings obtained by the algorithms.

\begin{enumerate}
 \item \textit{Qualitative Assessment.} In the case of two-dimensional embeddings, we simply plot the embedding output. In case of higher dimensional embeddings, we visualize the high-dimensional embedding in two dimensions using t-SNE~\citep{maaten2008visualizing}.
Previous work in OE also report visualisation as an evaluation criteria~\citep{agarwal2007generalized,tamuz2011adaptively,van2012stochastic}.
\item \textit{Triplet Error.} Train triplet error is the proportion of the input triplets that are not satisfied by the estimated embedding. To evaluate the \textbf{test triplet error}, we randomly sample $10,000$ triplets, independent of the set of training triplets, based on the ground-truth data. This metric measures generalization of the obtained embedding to held-out triplets.
\item \textit{Procrustes Disparity.} Measures the distance between an orthogonal transformation of the output embedding and the ground-truth~\citep{gower2004procrustes}. We minimize the Frobenius norm between the original embedding $X^\ast$ and an orthogonal transformation of the embedding output $X$: $\min_{U} \left\Vert XU - X^\ast \right\Vert_F^2$. If the ordinal embedding has a lower dimension, we consider the embedding in the original dimension by adding zero columns to $X$. Note that this metric can only be utilized when a ground truth embedding is available which is indeed the case for all of our datasets.
\item \textit{kNN Classification Error.} In order to capture how well the local structure is preserved, we train a $k$-nearest neighbour classifier on 70\% of the dataset and measure the error of the remaining 30\% of the data. We fix the parameter $k = \floor{\log n}$. Therefore, we can only report the performance of the algorithms via this evaluation metric on labelled datasets.
\item \textit{Running Time.} The running time is measured as long as the change in train triplet error on a randomly chosen subset of training triplets is at least $0.005$.
\end{enumerate}
  \begin{figure*}[!htb]
{\centering
\includegraphics[width=0.95\textwidth]{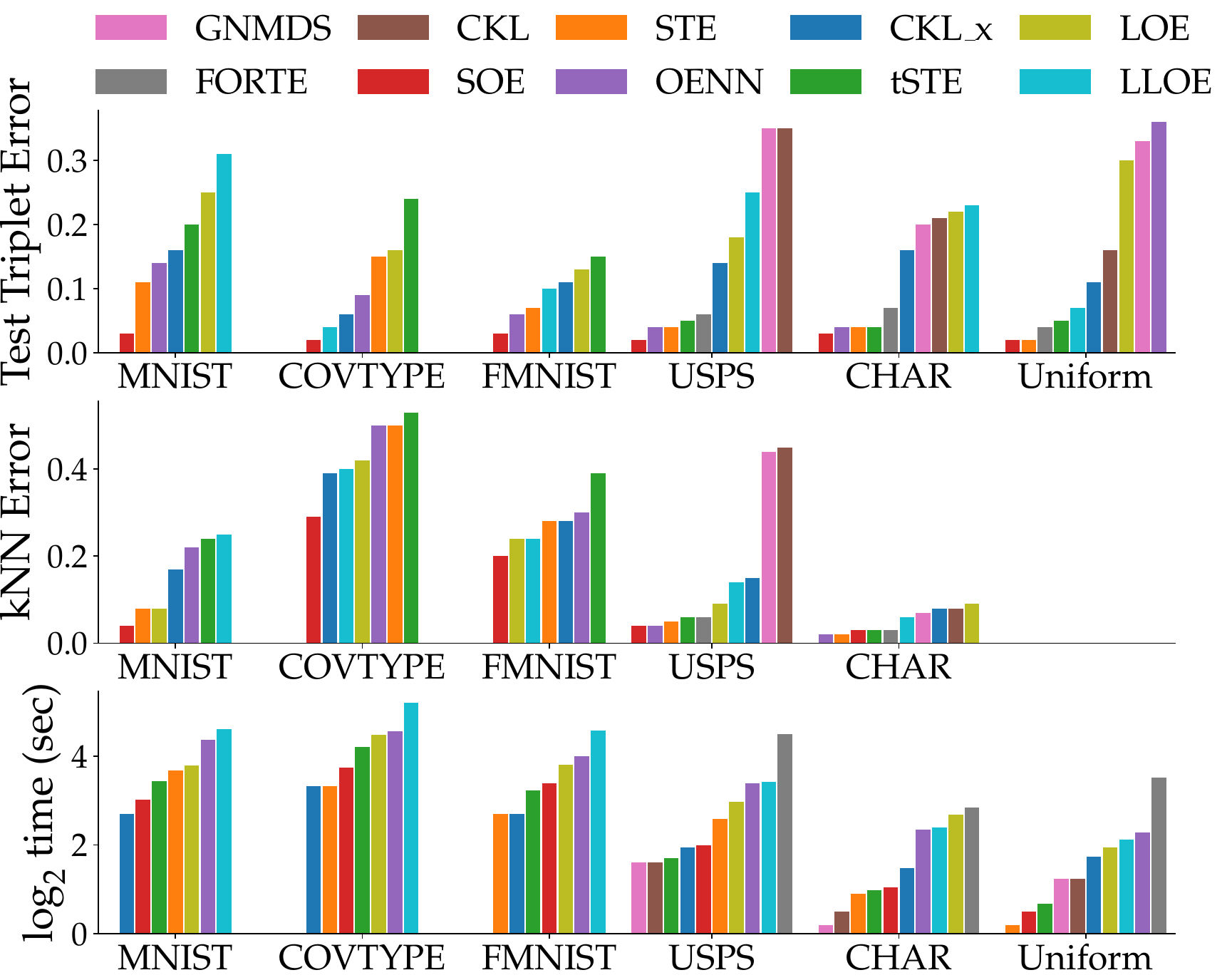}
}
\caption{General Experiments\label{subfig:general_experiment}. The rows represent test triplet error, kNN error, and running time of all the algorithms on 6 datasets, respectively. Uniform data does not have labels for computing the kNN error, hence the corresponding plot is missing. We omit Procrustes error since the results are qualitatively similar to those of test triplet error.}
\end{figure*}

\subsection{For a General Ordinal Embedding Task, which Algorithm should you use?}
We first investigate the general performance of the algorithms on six different datasets with a wide range of sample sizes ($1.8K$ to $581K$) and input dimensions ($2$ to $784$). We evaluate the performance of each algorithm on all datasets using our evaluation criteria. Algorithms which optimize over the Gram matrix, that is, CKL, GNMDS, and FORTE cannot be evaluated on datasets over $10,000$ items since the Gram matrix becomes too large to fit in the GPU and the optimization is computationally prohibitive on a CPU. Therefore, we exclude these methods in evaluations with larger datasets. We present the results of our evaluation in Figure~\ref{subfig:general_experiment} and summarize some of the key findings here.

\begin{itemize}
    \item In terms of reconstruction of the original embedding, without exception, SOE is the best performing algorithm. This can be observed in Figure~\ref{subfig:general_experiment} which shows that SOE achieves the lowest test triplet error compared to all the other methods. 
    
    \item SOE is also consistently among the best performing methods for preserving the local neighborhood as measured by the kNN error. Contrary to the popular belief that t-STE better preserves the local neighborhood, our experiments suggest that it can often be among the poorly performing algorithms in this respect. 
    
    \item Across the datasets, SOE incurs median running time, but among the algorithms which have good reconstruction performance, there is no algorithm with clear advantage in running time over the others.
    
    \item Perhaps surprisingly, CKL and GNMDS have the lowest running times of all the methods: even though each gradient step scales as $\bigO(n^2)$ the number of iterations required for convergence of these methods is significantly lower than that of the other methods. However, among the passive methods, they demonstrate the worst performance which is consistent with similar observations in the literature~(\cite{van2012stochastic,ghosh2019landmark}). 
    
    \item Consistently across all the experiments, algorithms that directly optimize over the embedding matrix appear to outperform those that consider optimization over the gram matrix. Among all Gram matrix based methods, only FORTE is somewhat competitive with the others with respect to triplet error and kNN error, although, the running time is significantly higher. 
    
    \item Active, landmark-based approaches: LOE and LLOE do not demonstrate consistent superior performance in either running time or accuracy. In particular, the running time of LLOE is consistently amongst the worst, since, unlike the other methods, LLOE cannot be fully parallelized over GPU cores. 

\end{itemize}


\begin{figure}[!htb]
\centering
\subcaptionbox{MNIST}[.49\textwidth]{
	\centering
	\includegraphics[width=.49\textwidth]{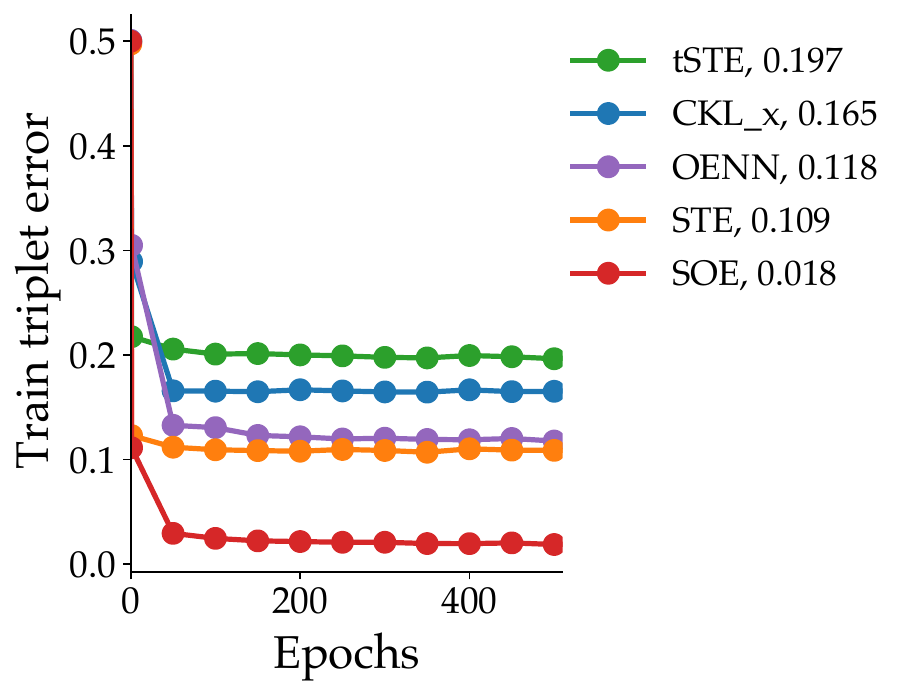}
}
\subcaptionbox{Covertype}[.49\textwidth]{
	\centering
	\includegraphics[width=.49\textwidth]{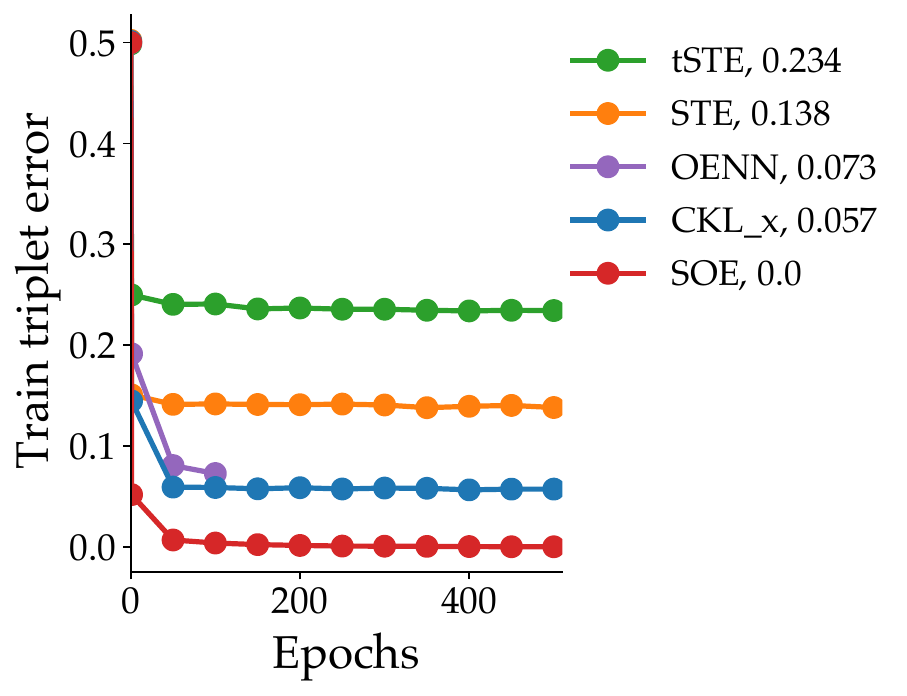}
}
\hspace*{\fill}

\subcaptionbox{USPS}[.49\textwidth]{
	\centering
	\includegraphics[width=.49\textwidth]{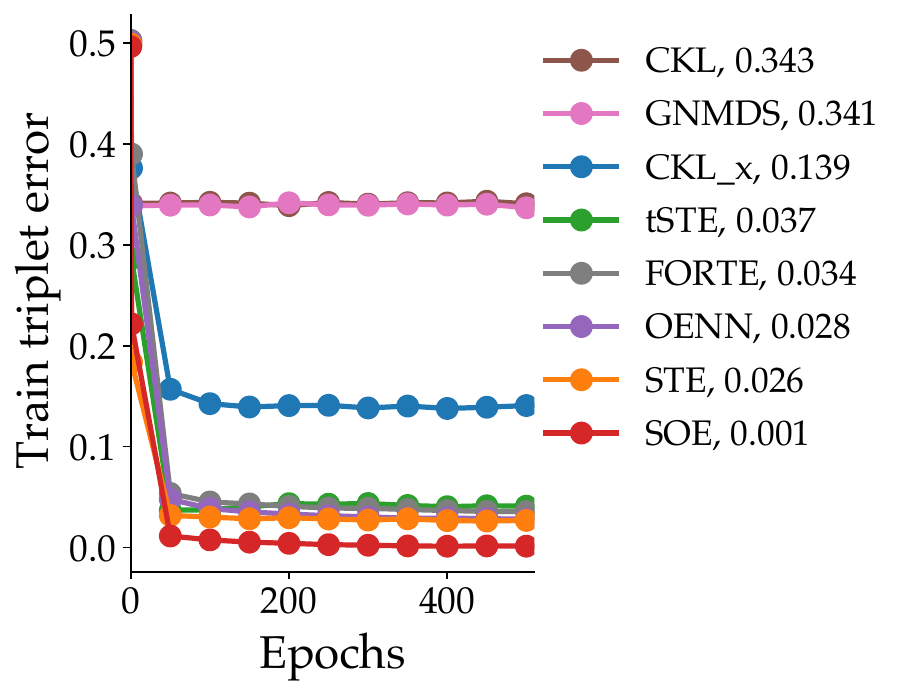}
}
\subcaptionbox{Char}[.49\textwidth]{
	\centering
	\includegraphics[width=.49\textwidth]{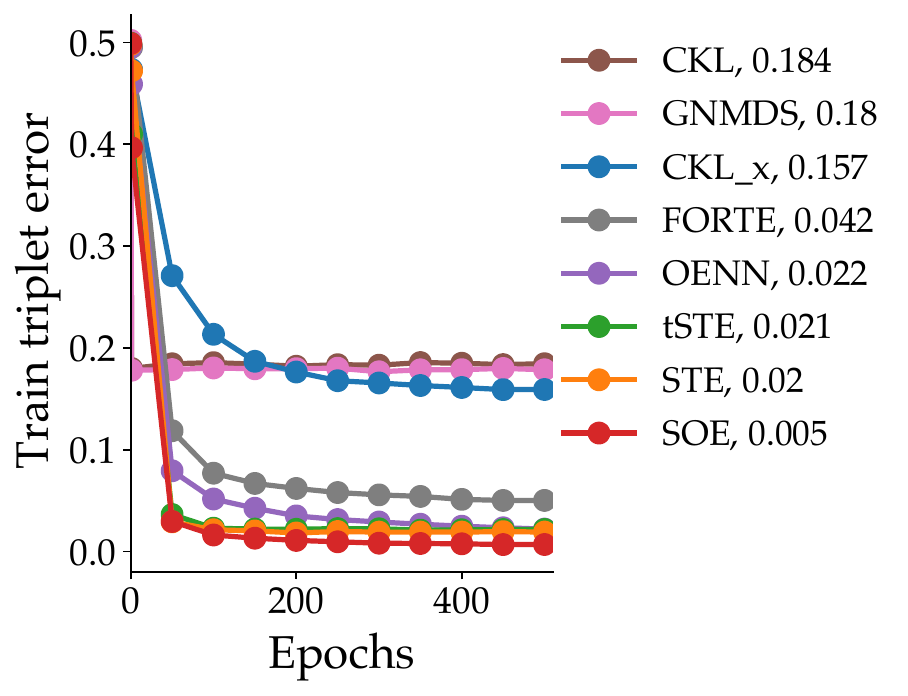}
}
\hspace*{\fill}
\caption{\label{fig:convergence}Train triplet error of the OE algorithms with increasing number of training epochs. We display the final triplet error against each method in the legend. For CKL, GNMDS, and FORTE, the MNIST dataset is discarded due to computational and memory issues as discussed earlier. Several methods converge to the global optimum, while on MNIST the only method to converge is SOE.}
\end{figure}
\subsection{Do the Non-Convex Approaches to OE suffer from Local Optima?}
Finding a feasible solution to the OE problem constitutes optimizing over a non-convex objective function and hence some ordinal embedding algorithms use convex relaxations or majorizing functions to make the optimization problem easier to handle \citep{agarwal2007generalized, terada2014local}. Other OE algorithms involve optimizing over non-convex landscapes by means of first order methods. A natural question to ask here would be ``Are local optima a hindrance to these OE algorithms due to the non-convexity of their objectives?'' In this section, we provide extensive empirical evidence that suggests otherwise. Moreover, we show that certain algorithms with non-convex objectives obtain globally optimal solutions to the problem. 

 To test this hypothesis, we train all the algorithms over a selection of datasets (see Table~1 in the supplementary) and plot the train triplet error with increasing number of epochs. The main objective of this experiment is to verify whether the non-convex approaches are able to achieve \textit{near zero} train triplet error which would indicate obtaining a globally optimal solution. Note that the loss functions used by different algorithms vary both in nature and in scale, therefore the train triplet error is used as a unifying metric for convergence. Train triplet error cannot be meaningfully defined for active approaches since unlike passive methods, they query specific triplets from an oracle and are not provided a set of training triplets. Therefore, we exclude them from the current experiment. Figure~\ref{fig:convergence} depicts the train triplet error of eight OE methods on three datasets. The final value of the train triplet error after $500$ epochs is reported in the legend of each plot. As it is evident from plots, SOE is the only algorithm that consistently, across all the datasets, obtains the globally optimal solution, that is, attains nearly zero train triplet error. OENN, STE, and t-STE attain relatively small training error. Interestingly, even though the learning objective of OENN and SOE are near identical, we find that on MNIST, OENN indeed appears to \textit{get stuck in a local optimum} while SOE does not. We infer this since the train triplet error attained by OENN plateaus after around 200 epochs and remains high thereafter. In line with the rest of our experiments, CKL and the convex approach GNMDS attain high train triplet error. To conclude, most of the non-convex approaches to OE do not seem to suffer from local optima. In particular, given sufficiently many epochs and with an appropriate learning rate, the optimization of SOE always finds a globally optimal solution. Additional results from this experiment are provided in the Supplementary (Section \ref{sec:convergence_supp})

\begin{figure}[!htb]
\centering
\subcaptionbox{\label{fig:increasing_triplets_aggregation}Aggregation}[.8\textwidth]{
	\centering
        \includegraphics[width=.7\textwidth]{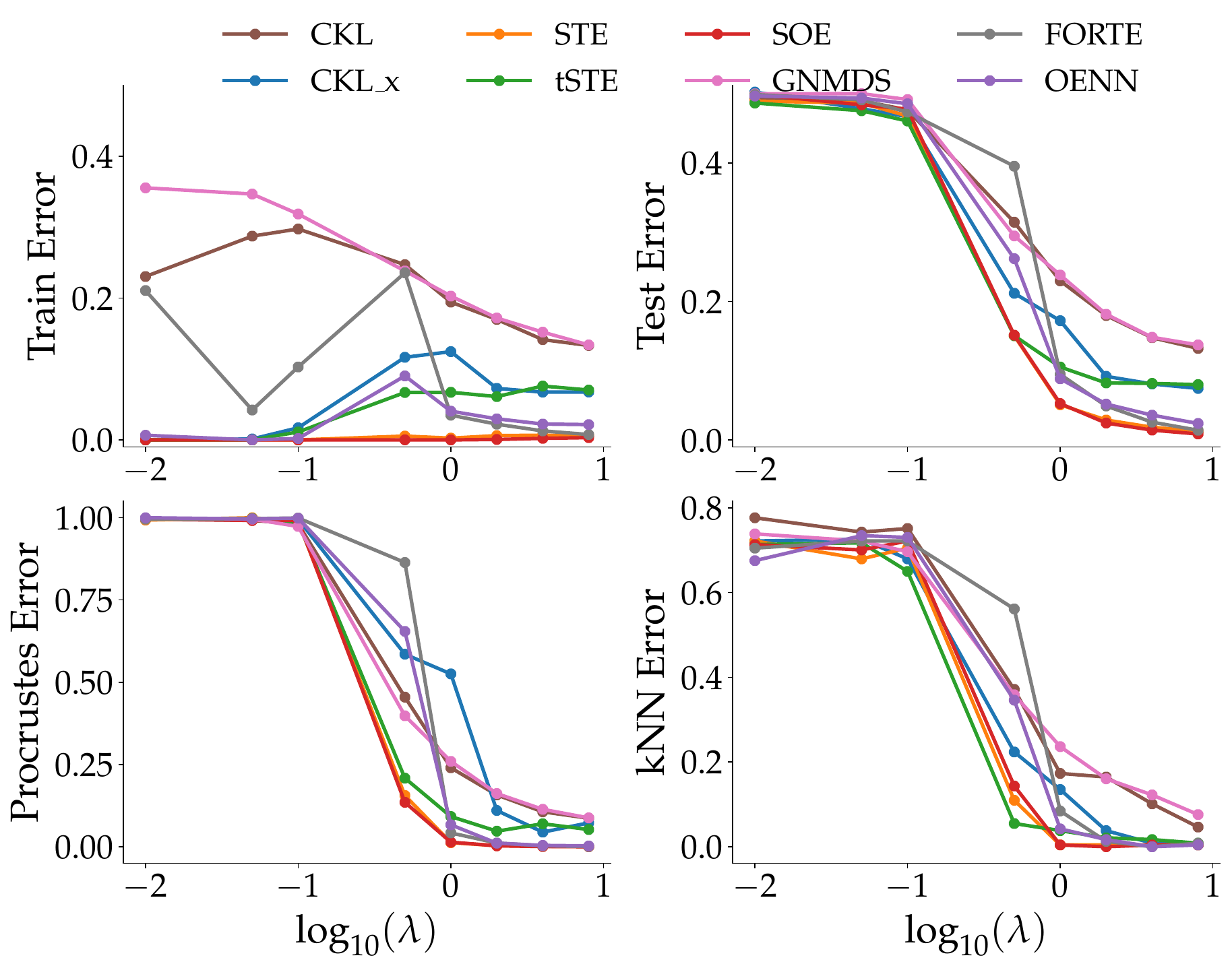}
 }
 
 \vspace{3mm}
 
 \subcaptionbox{\label{fig:increasing_triplets_mnist}MNIST}[.8\textwidth]{
	\centering
        \includegraphics[width=.7\textwidth]{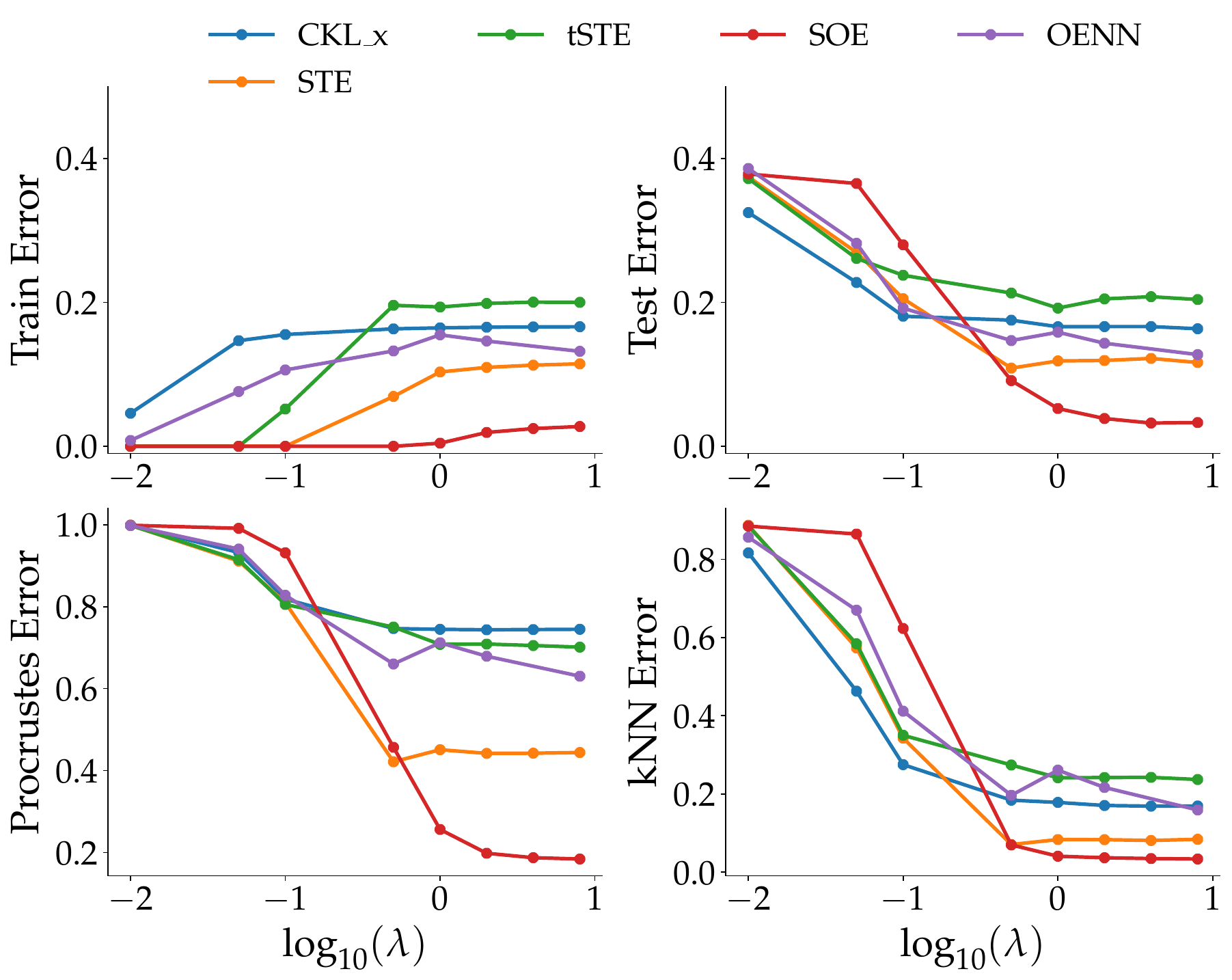}
 }
 \caption{Train, Test, Procrustes and kNN Errors for algorithms on two datasets(Aggregation, MNIST) with increasing number of triplets specified by the triplet multiplier $\lambda$. More results on other datasets, appear in the supplement.  \label{fig:increasing_triplets_all}}
\end{figure}
\begin{figure}[!htb]
   \centering

	\includegraphics[width=.3\textwidth]{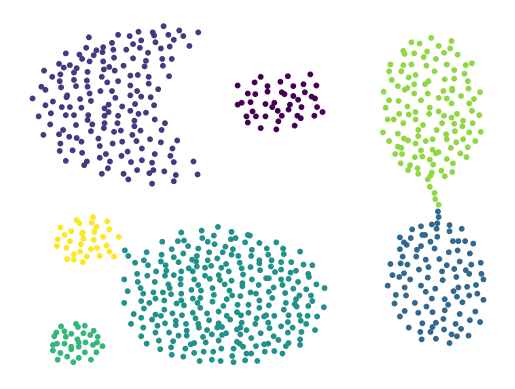}
    \caption{Original 2d visualization of the \textbf{Aggregation} dataset.}
    \label{fig:original_aggregation}
\end{figure}

\begin{figure*}[!htb]

\centering
\includegraphics[width=.95\textwidth]{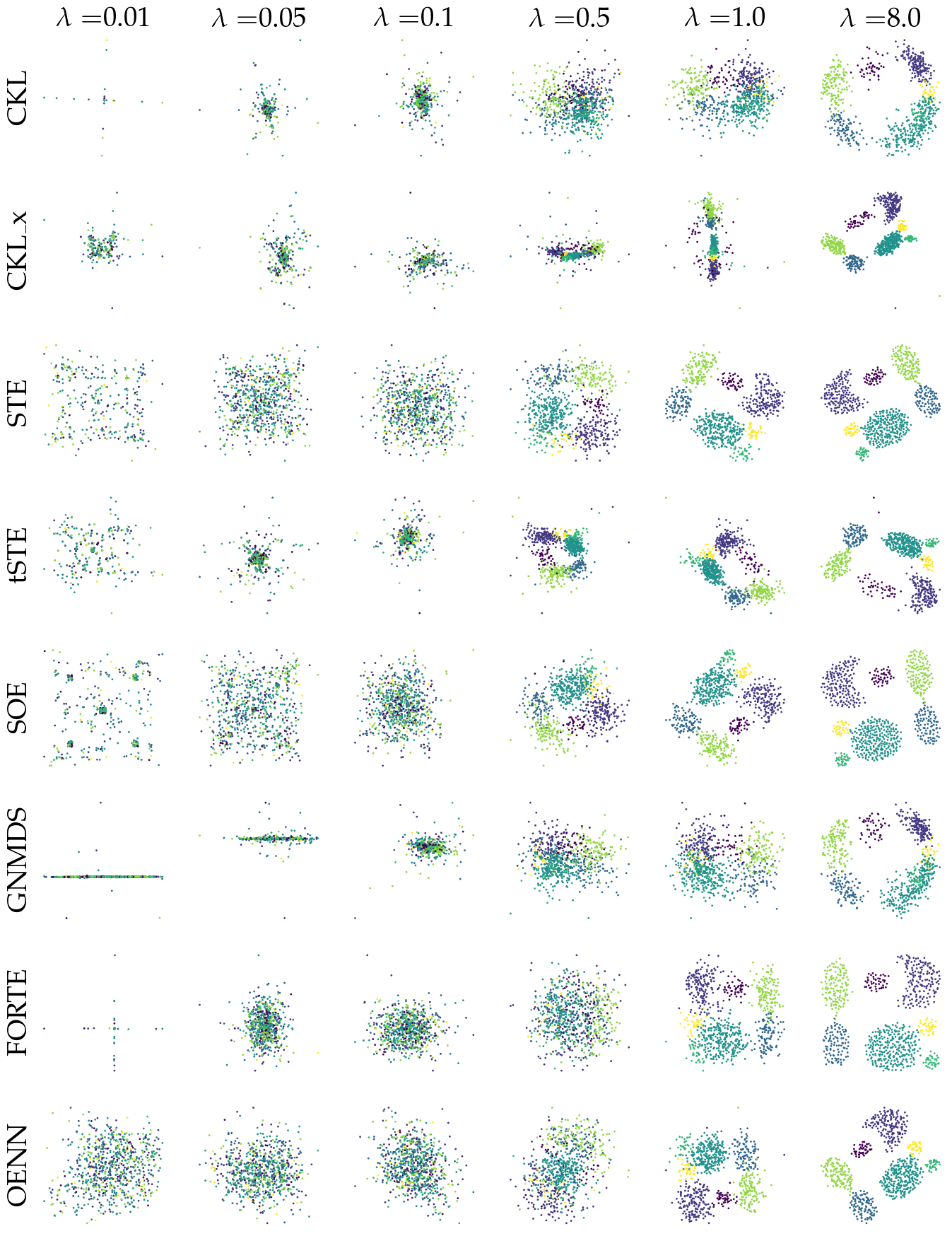}
\caption{For each algorithm we show the embeddings which are obtained when learning with a varying number of train triplets. This is controlled with the triplet multiplier $\lambda$ yielding $\lambda d n \log n$ triplets. }
\end{figure*}

\subsection{Which Algorithms perform Better in Different Triplet Regimes?}
\label{sec:increasing_triplets}
In this experiment, we vary the number of triplets as a factor of $n d \log n$ and examine the performance of all methods. Since the number of available triplets cannot be fully controlled for the active approaches, namely, LOE and LLOE, we exclude them in our analysis. In this experiment, we vary the number of triplets $\lambda n d \log n$ with the triplet multiplier $\lambda$ ranging from $0.01$ to $8$ for three high-dimensional datasets and seven 2D datasets. We primarily focus on two aspects of the obtained embeddings: How well does the obtained embedding reconstruct the original embedding, as measured by the Procrustes and test triplet errors and how well does it preserve the local neighborhoods which is quantified using kNN error.


We observe two different regimes in the parameter space of the triplet multiplier $\lambda$: the low triplet regime where the number of available triplets are less than $n d \log n$ and the large triplet regime, where the number of triplets are larger than $ n d \log n.$ We make this distinction following our observation in our experiments that there is a relatively sharp transition in the Procrustes error (or test error) when the number of available triplets is between $0.5 \cdot n d \log n$ to $1 \cdot n d \log n.$ This can, for instance, be clearly observed in Figure~\ref{fig:increasing_triplets_all}.

\paragraph{Small Triplet Regime.}
Intuitively speaking, when the number of available triplets is small it should be easier to produce an embedding that does not violate triplet constraints. On the other hand, a small set of triplets does not uniquely specify an embedding up to similarity transformations. Therefore, without additional constraints, the Procrustes error or the test triplet error which indicate the quality of reconstruction of the original embedding can be rather high. This is reflected for  $\log_{10}(\lambda) < 0$ in Figure~\ref{fig:increasing_triplets_all}, where we plot the Procrustes error and kNN error on two different datasets for all the embedding methods.

For most datasets, when the number of triplets are less than $0.5 \cdot n d \log n$, none of the algorithms achieve better reconstruction error than a randomly initialized embedding, that is a Procrustes error of $1.0$ (see, for example, Figure~\ref{fig:increasing_triplets_all}).

\paragraph{Large Triplet Regime.} When the number of triplet constraints are large enough, embeddings that can satisfy all the constraints are unique up to similarity transformations \citep{kleindessner2014uniqueness, jain2016finite}. Therefore, the train triplet error often faithfully reflects the test and the Procrustes error in this regime. Since SOE always achieves the best train triplet error, as expected, in the large triplet regime it is also the best performing method in terms of reconstructing the original embedding as well as in preserving the local neighborhood (see Figure~\ref{fig:increasing_triplets_all}) . Some of the other OE approaches, namely STE, OENN, and FORTE achieve good generalization in this regime for a majority of the datasets. However, unlike SOE, on a small fraction of the datasets, they fail to achieve good training as well as test errors providing more evidence supporting the hypothesis that SOE does not suffer from local optima unlike the other non-convex approaches. 

\paragraph{Other findings.}

\begin{itemize}
    \item  Preliminary evaluation in \citet{van2012stochastic}, which compared t-STE with STE, seemed to suggest that t-STE better preserves the local neighborhood. Our extensive evaluation finds some evidence to the contrary. For seven out of the ten datasets, in both the small and the large triplet regimes, STE outperforms t-STE on both Procrustes error as well as on kNN classification error.
    \item In the large-triplet regimes, across most of the datasets, the worst performing methods are CKL and GNMDS. This trend can be observed in other experiments as well: Gram matrix approaches are the  worst performing methods with the exception of FORTE.
 \end{itemize}

\begin{figure}[!htb]
\centering
\subcaptionbox{USPS}{  \includegraphics[width=0.65\textwidth]{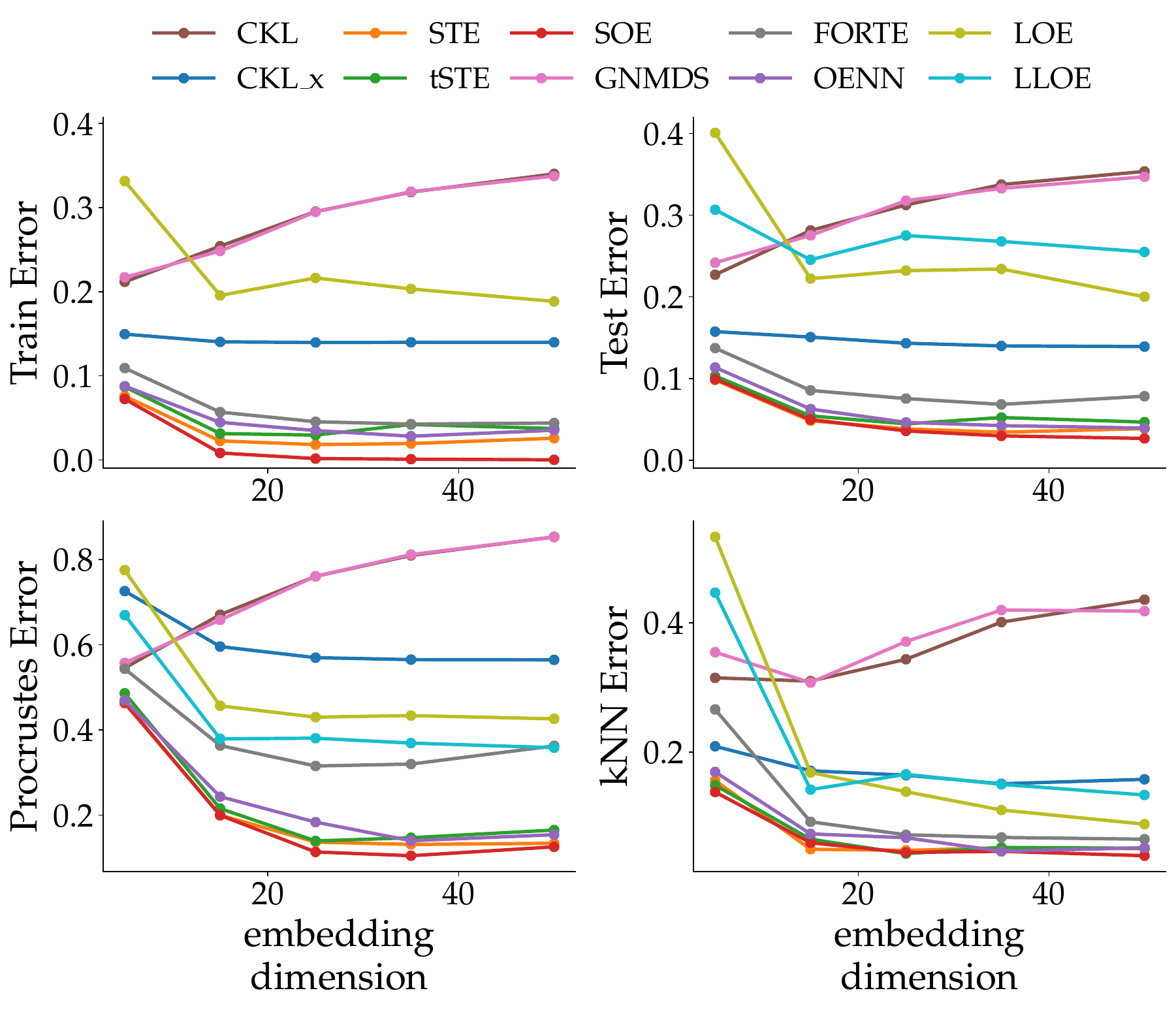}}
\subcaptionbox{KMNIST}{
    \includegraphics[width=0.65\textwidth]{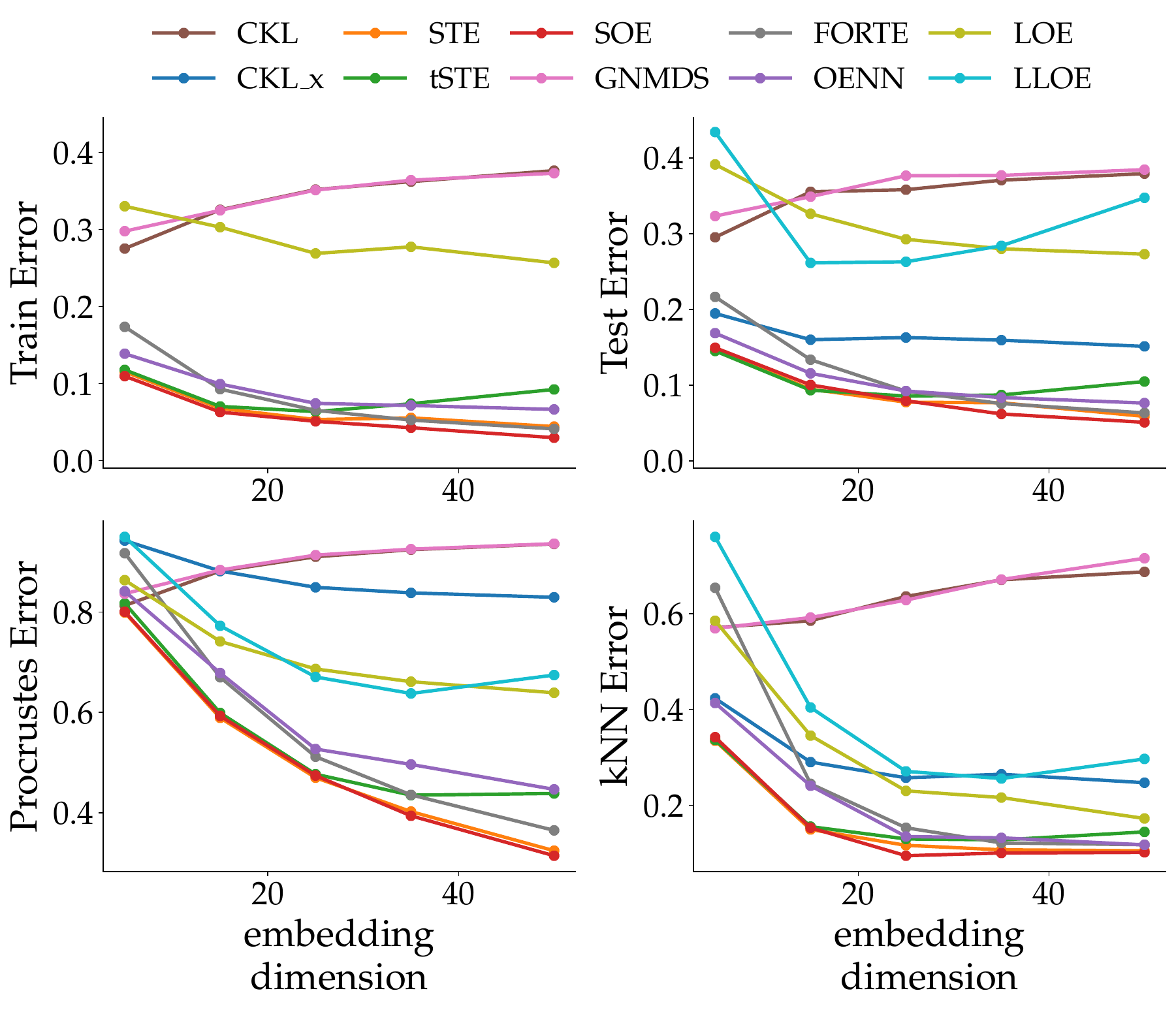}}
\caption{\label{fig:increasing_d}Increasing embedding dimension (USPS, KMNIST). Train, Test, Procrustes and kNN errors of  all algorithms as we increase the embedding dimension from $5$  to $50$ for the USPS dataset. The number of triplets are kept constant at $2dn \log n$.}
\end{figure}
\begin{figure}[!htb]
\begin{center}
\begin{subfigure}[b]{0.19\textwidth}
     	\includegraphics[width=1.0\textwidth]{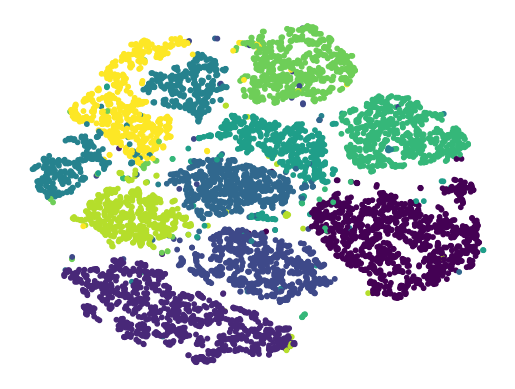}
     	\caption{USPS: Original.}
\end{subfigure}\end{center}

{\begin{center}
\begin{subfigure}[b]{0.74\textwidth}
\includegraphics[width=1.0\textwidth]{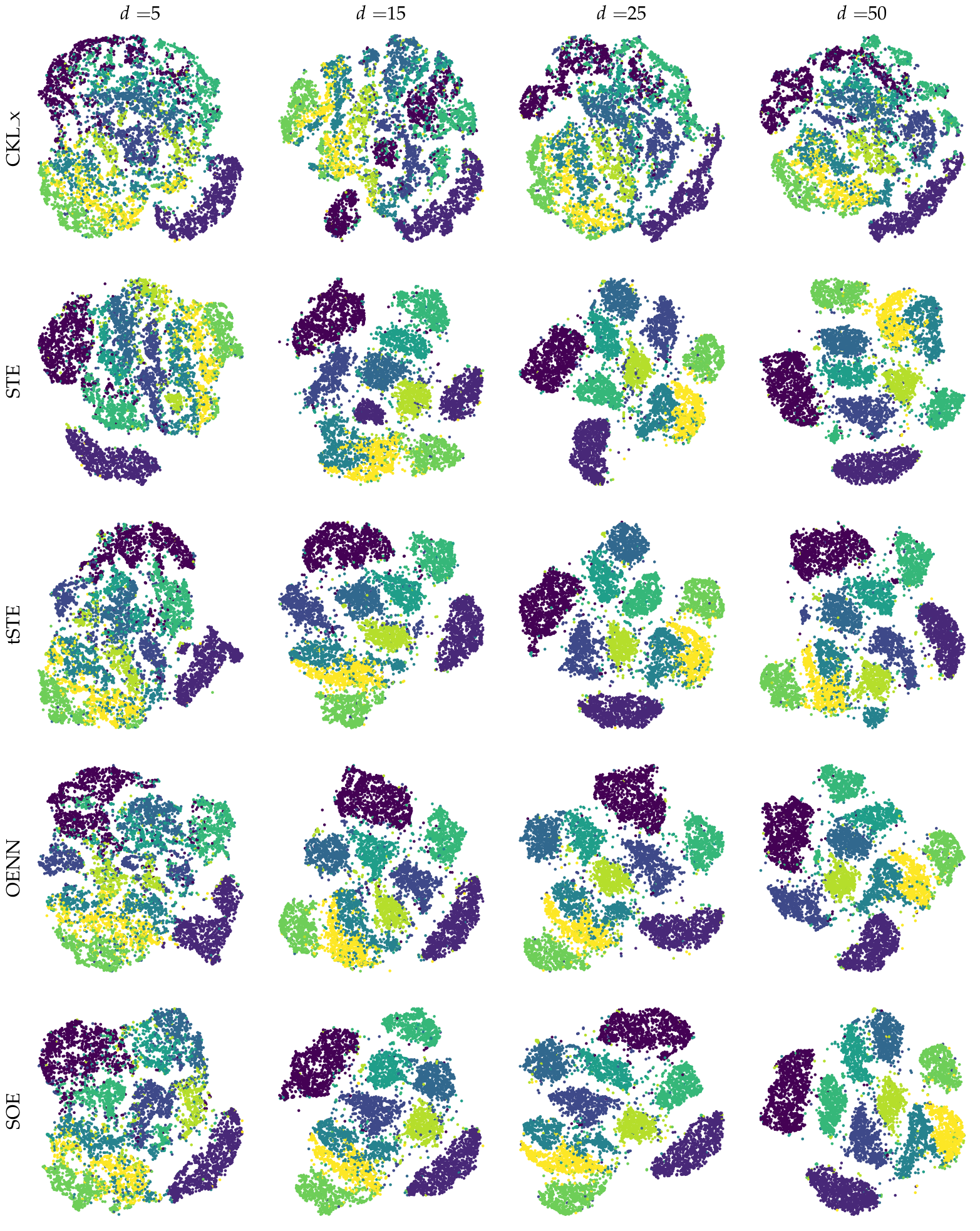}
\caption{\textbf{USPS: Reconstructions}}
\end{subfigure} \end{center}}
\caption{(a) A t-SNE visualization of the original USPS dataset. (b) t-SNE visualizations of embeddings created by SOE, OENN, tSTE, STE, and CKLx with increasing embedding dimension. The colors represent digits of USPS. Note that the labels are only used for the purpose of visualization and are not utilized to generate triplets.}
\end{figure}

\begin{figure}[!htb]
\begin{center}
\begin{subfigure}[b]{0.19\textwidth}
     	\includegraphics[width=1.0\textwidth]{img/increasing_d/usps/2d_emb/original.png}
     	\caption{USPS: Original.}
\end{subfigure}\end{center}

{\begin{center}
\begin{subfigure}[b]{0.73\textwidth}
\includegraphics[width=1.0\textwidth]{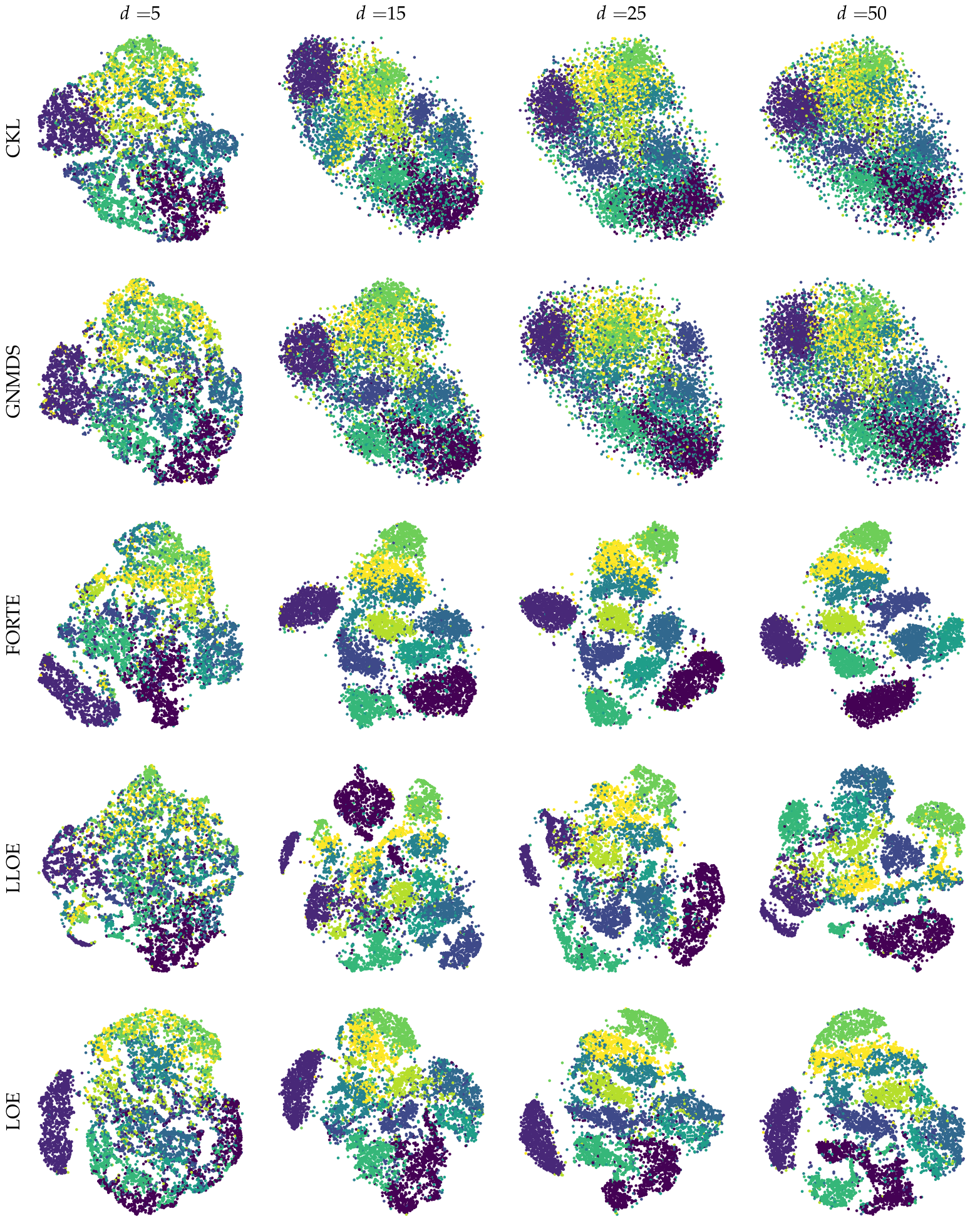}
\caption{\textbf{USPS: Reconstructions}}
\end{subfigure} \end{center}}
\caption{(a) A t-SNE visualization of the original USPS dataset. (b) t-SNE visualizations of embeddings created by LOE, LLOE, FORTE, GNMDS, and CKL. The colors represent digits of USPS. Note that our triplets are not generated based on class labels and the labels are only used for the purpose of visualization.}
\end{figure}
%

\subsection{Which Algorithms perform Better for Low Embedding Dimensions?}
In this experiment, we evaluate the performance of the embedding algorithms by varying the embedding dimension. The number of triplets used are kept constant at $2dn \log n$. A low embedding dimension is a hard constraint on the OE problem since we reduce the degrees of freedom to fulfill all triplet constraints. On the other hand, with higher embedding dimensions it is easier to find an embedding that satisfies a larger number of triplet constraints. Here, since the embedding dimension is always chosen to be smaller than the input dimension, we expect the train error, as well as reconstruction errors, to decrease as the embedding dimension increases. Further, it would be of interest to see if some algorithms perform better than the others in the low dimensional setting in terms of either reconstruction or in preserving the local neighborhood.
In Figure~\ref{fig:increasing_d}, we observe that most methods obtain lower Procrustes and kNN error with higher embedding dimension. SOE, STE, tSTE, and OENN exploit the higher dimensions the most as evidenced by the relatively sharp decrease in both kNN and Procrustes errors. Interestingly, CKL and GNMDS, which optimize over the Gram matrix and, hence, do not require the dimension $d$ as a parameter, worsen in performance with increasing embedding dimension. In this experiment, we do not observe a favourable inductive bias for any algorithm when the embedding dimension is small.

\begin{figure}[!htb]
\centering
\subcaptionbox{{\bfseries USPS:} Increasing $p$ of Bernoulli Noise.}{\includegraphics[width=.7\textwidth]{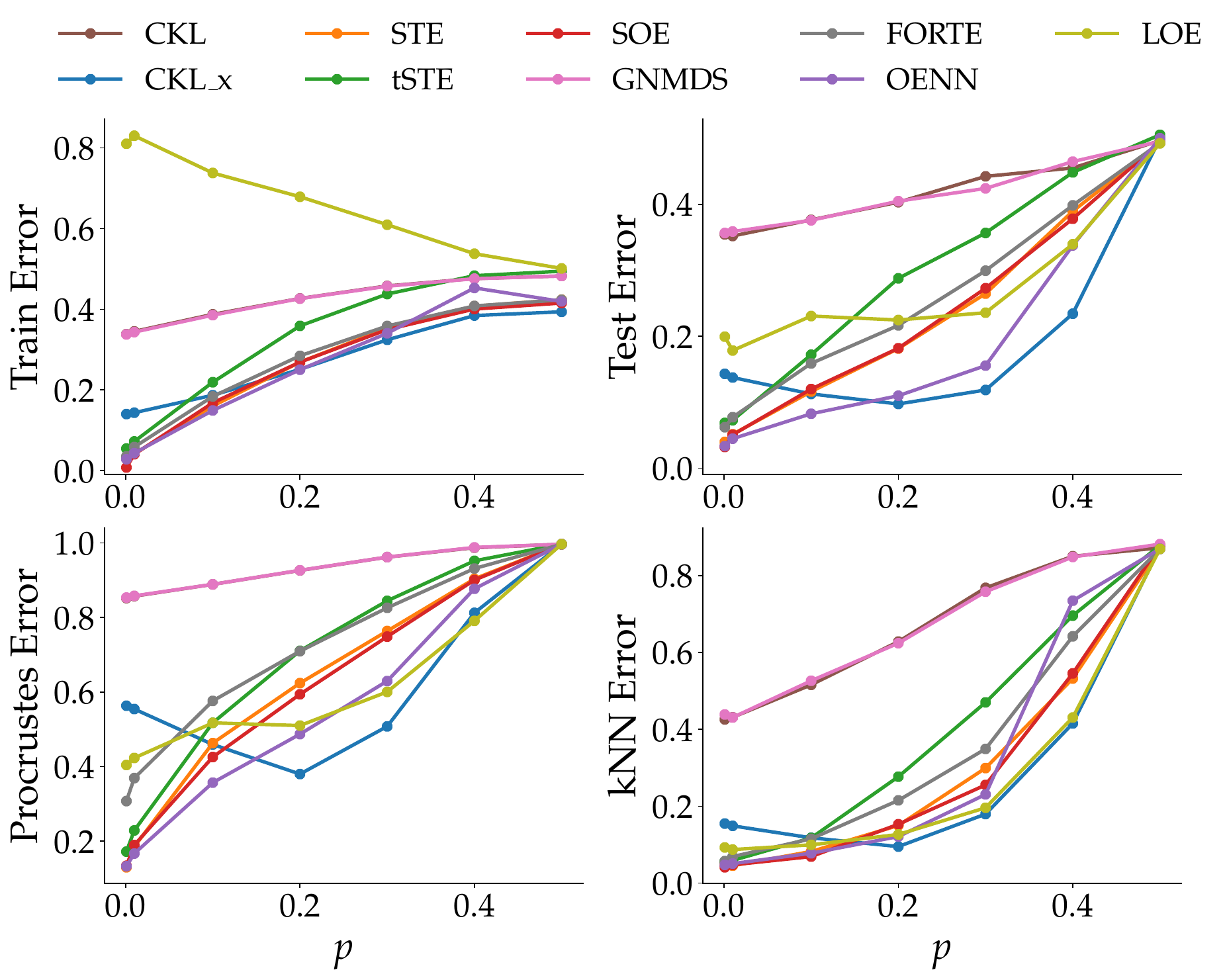}}
\subcaptionbox{{\bfseries CHAR:} Increasing $\sigma$ of Gaussian Noise.} {\includegraphics[width=.7\textwidth]{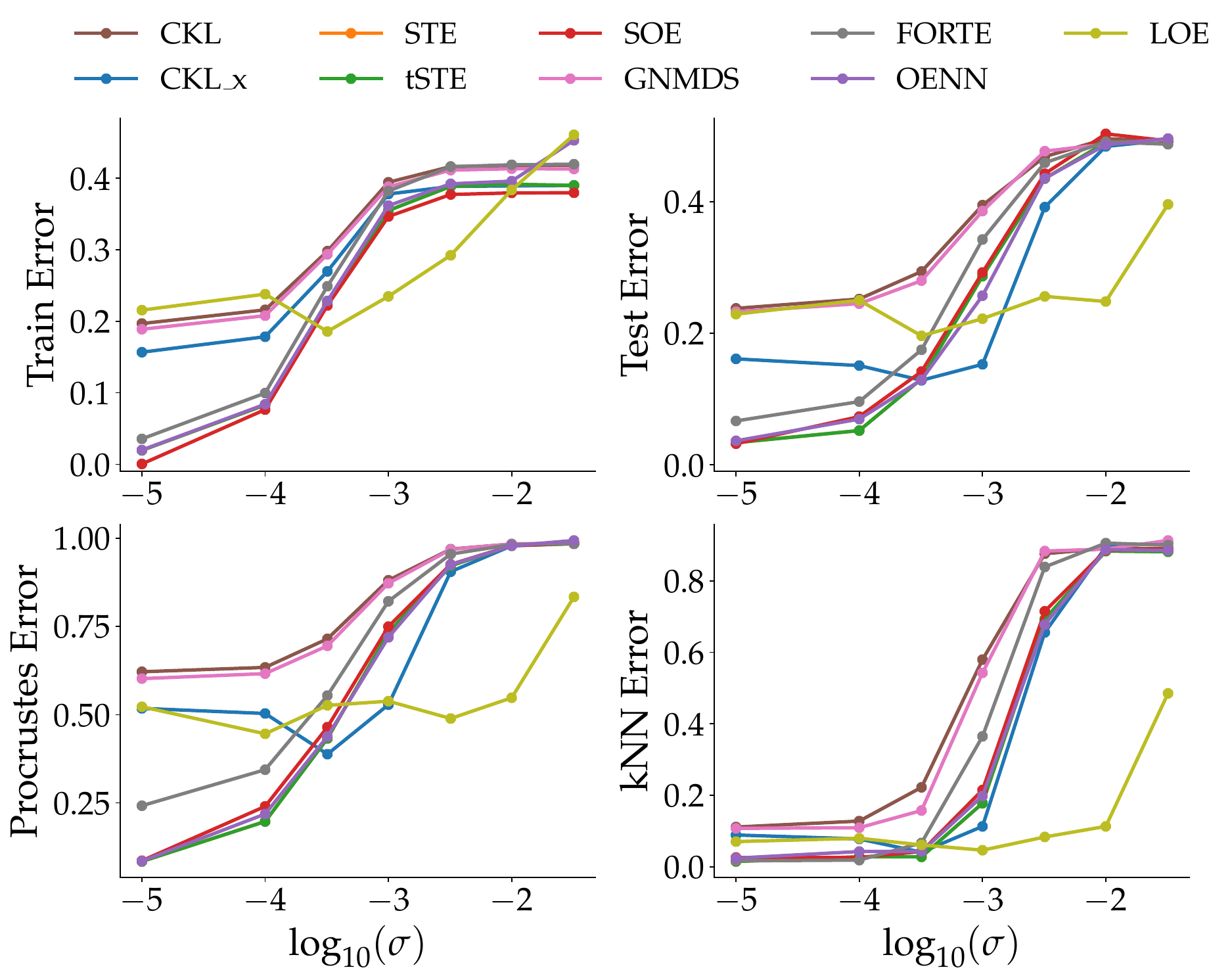}}
\caption{\label{fig:increasing_noise}  Train, Test, Procrustes and kNN error of algorithms with (a) increasing with increasing Bernoulli Noise on USPS and (b) increasing Gaussian Noise on CHAR datasets. The value of triplet multiplier is kept constant at $\lambda$ is $2$.}
\end{figure}

\FloatBarrier
\subsection{What is the effect of noise on the performance of OE algorithms?}
Here, we evaluate the effect of noise on the performance of OE algorithms. To this end, we consider two noise models on our training triplets.
 \paragraph{Bernoulli Noise Model.} In this model, we independently flip a true triplet $(x,y,z)$ to $(x,z,y)$ with probability $p$. This model is fairly intuitive and the probability $p$ naturally determines the level of noise.
 \paragraph{Gaussian Noise Model.} When creating triplets from the ground truth, we need to evaluate two distances: $d(x,y)$ and $d(x,z)$. In this noise model, we independently add a centered Gaussian $\mathcal{N}(0,\sigma)$ to the distances. We then generate triplets by comparing the noisy distances. The variance $\sigma$ represents the noise level. 
 
 \par
 To conduct the experiment, we vary the noise level and evaluate the performance of the algorithms for each of the noise models. The number of triplets used are kept constant at $2dn \log n.$ The performance of all the OE methods naturally deteriorates with increasing noise levels under both noise models (see Figure~\ref{fig:increasing_noise}). Consistent with the other experiments, under low levels of noise, SOE clearly matches or outperforms the rest of the algorithms. Under moderate to high levels of noise, all the simple distance-based methods: SOE, STE, t-STE, and CKLx perform similarly and in line with the rest of the experiments, they outperform the gram-matrix based approaches as well as the more elaborate neural network or landmark-based approaches.

\begin{figure*}[!htb]
\centering
\subcaptionbox{\textbf{MNIST}}{
\includegraphics[width=.725\textwidth]{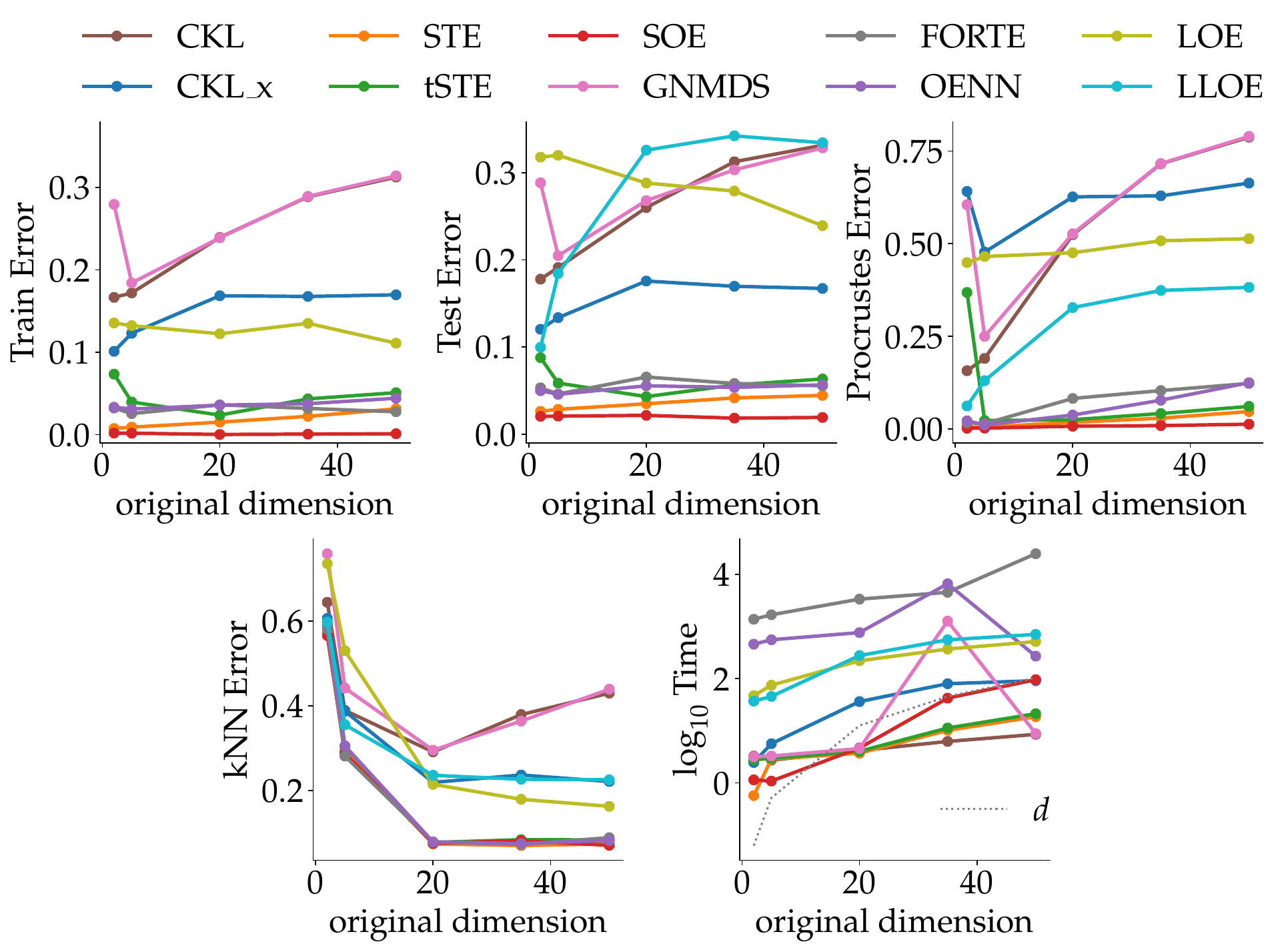}}
\vspace{3mm}
\centering
\subcaptionbox{\textbf{Gaussian Mixture}}{
\includegraphics[width=.725\textwidth]{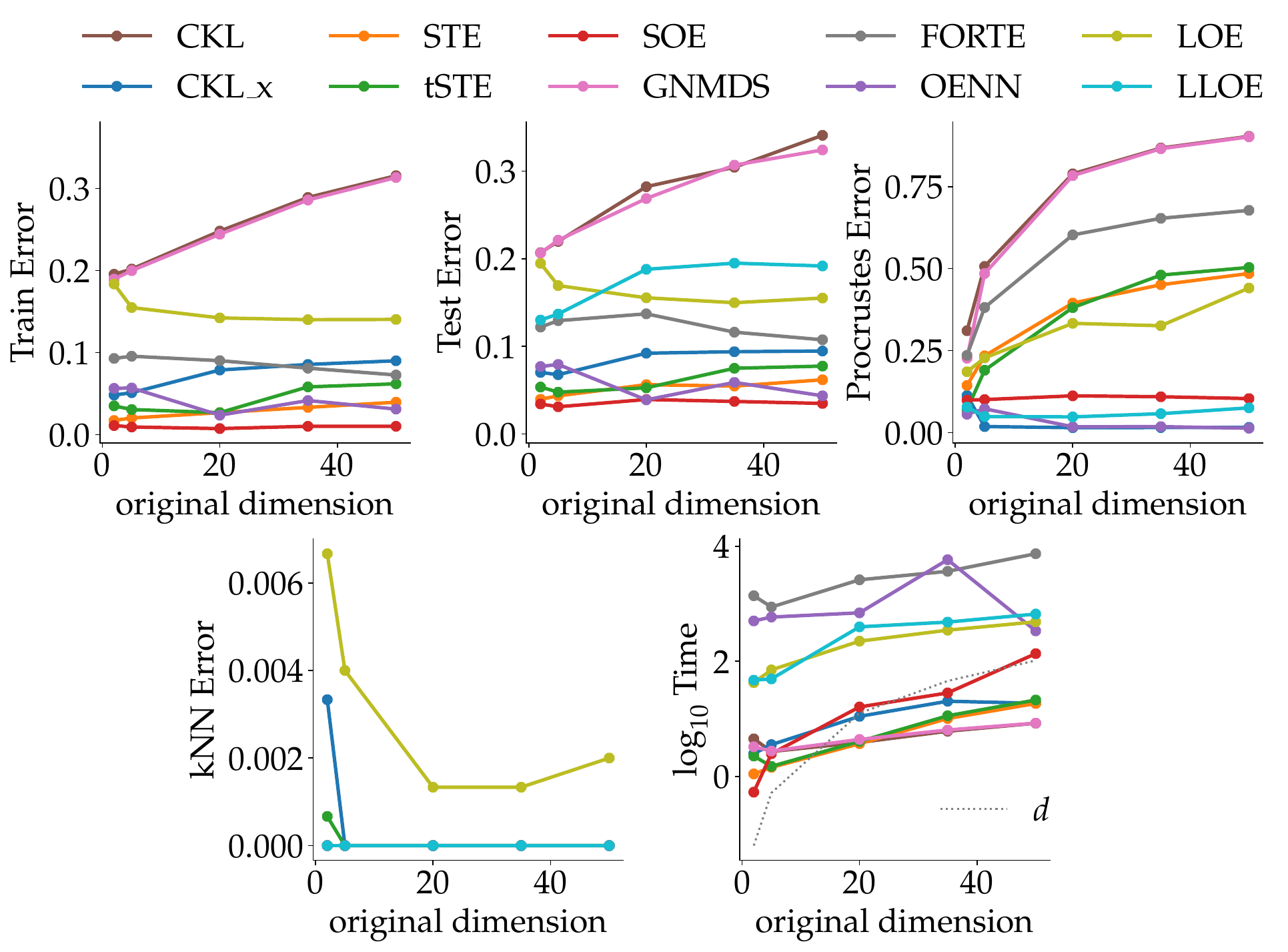}}
\caption{\label{fig:increasing_orig_d}Performance of algorithms with increasing original dimension. (a) \textbf{MNIST:} For this experiment, we use a PCA projection of $5,000$ MNNIST points into the corresponding dimension as the ground-truth to generate the train triplets. (b) \textbf{Gaussian Mixture:} Surprisingly, LOE performs very badly on this dataset. Recall, that this dataset consists of two well separable normal distributions.}
\end{figure*}

\subsection{Which Methods scale Better with Sample Size and Input Dimension?}
\label{sec:various_n}

In this experiment, we investigate the running time of the algorithms with varying input dimension and with varying sample size. Recall that the running time of \textit{each gradient update} scales as $\bigO(n^2 d \log n)$ and as $\bigO(n d^2 \log n)$ for algorithms optimizing over the Gram matrix and the data matrix respectively. 
To create datasets with increasing original dimension, we consider two different datasets: Gaussian mixture models (GMM) and PCA MNIST. To generate data from GMM, we sample from Gaussian mixtures in increasing input dimension. For PCA MNIST, we project the original MNIST data using PCA into Euclidean space with increasing dimensions. We keep the embedding dimension for ordinal embedding the same as the dimension of the generated datasets.

\paragraph{Increasing Input Dimension d.} As expected, the running time of all methods increases with input dimension. As anticipated, Figure~\ref{fig:increasing_orig_d}a shows that running time of SOE, STE, t-STE and CKLx, which are methods optimizing over the data matrix, roughly scales as $d^2$ while methods optimizing over the Gram matrix roughly scale as $d$. The running time of FORTE, OENN, and LLOE is much higher than that of the simple, non-convex approaches: SOE, STE, and t-STE.
The increasing input dimension also has an impact on the methods' performance. Only SOE consistently achieves a low procrustes error, while the procrustes error of STE, tSTE, and OENN decreases slightly. 

\begin{figure}[!htb]
\centering
\subcaptionbox{\textbf{Covertype}}{
\includegraphics[width=.725\textwidth]{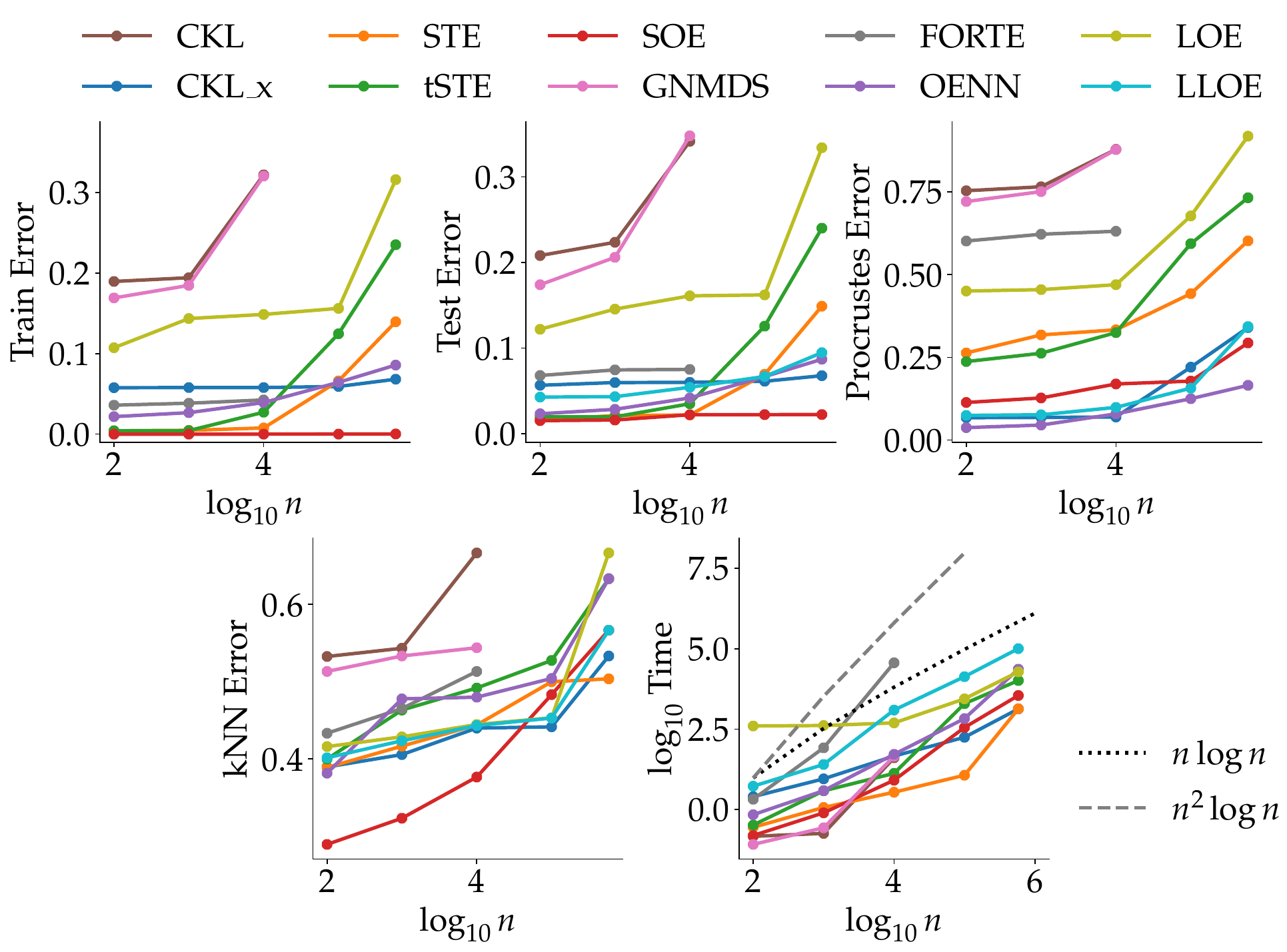}}
\centering
\subcaptionbox{\textbf{Uniform}}{
\includegraphics[width=.725\textwidth]{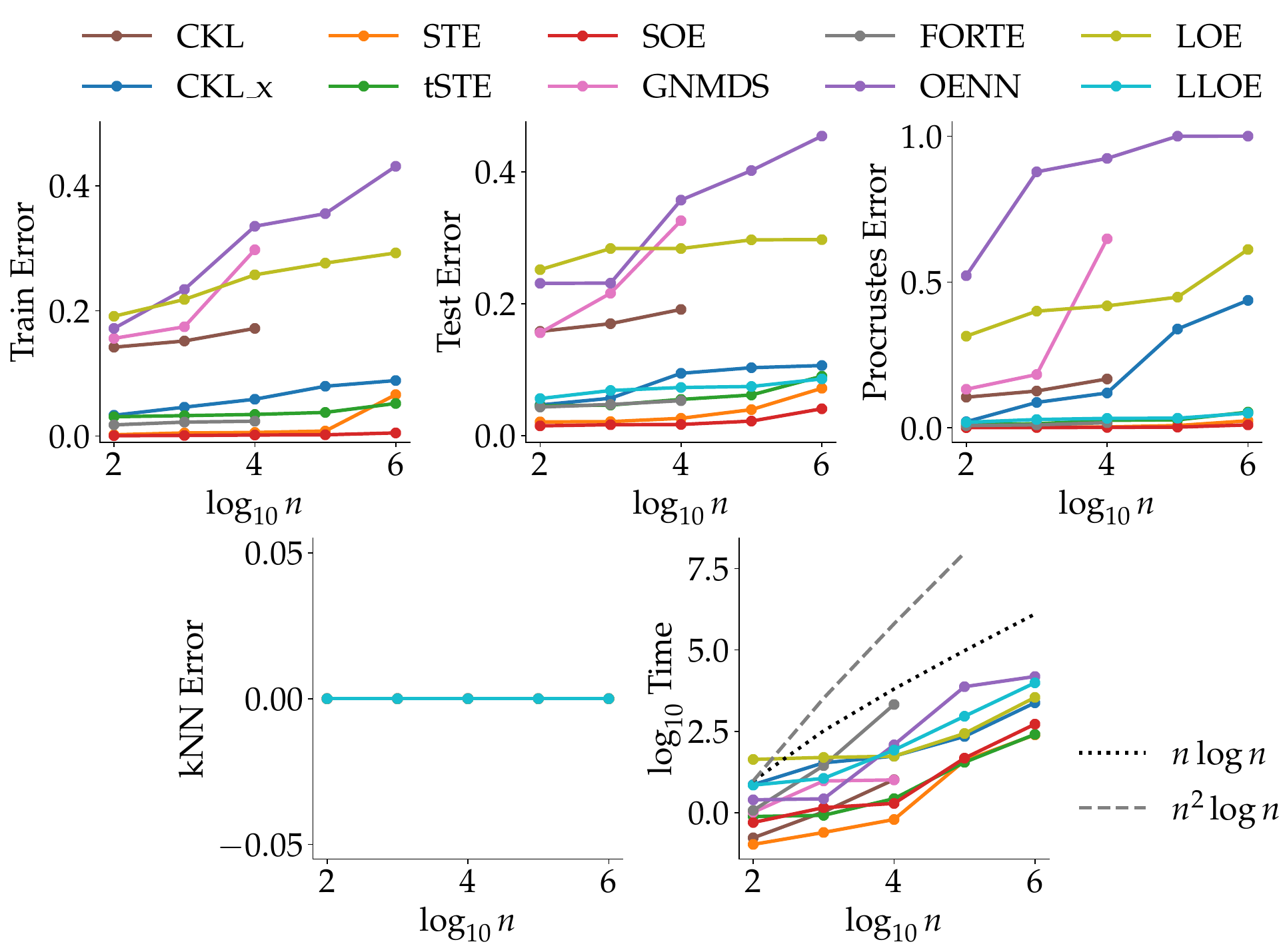}}
\caption{\label{fig:increasing_n} Train, Test, Procrustes, kNN errors and Time for algorithms with increasing sample size on two datasets: (a) Covertype, and (b) Uniform. kNN Error is specified as zero for uniform, since its an unlabelled dataset.}
\end{figure}

\paragraph{Increasing Sample Size n.} As discussed earlier, we cannot evaluate the Gram matrix approaches on datasets with more than $10K$ items and therefore, we clip the sample size at this limit for these approaches. Again, as expected, algorithms optimizing over the data matrix and Gram matrix exhibit a growth rate of $n\log n$ and $n^2 \log n$ respectively. OENN and LOE are an order of magnitude slower than the rest although the test error of OENN is considerably lower than that of LOE. We can observe in Figure~\ref{fig:increasing_n} that SOE has nearly constant test error with increasing number of points while the other methods worsen in performance. STE and t-STE, which are generally well performing methods, also seem to show a sharp increase in test error with increasing sample size.


\subsection{How do Methods scale on CPU vs GPU?}

\begin{figure}[!htb]
\centering
\subcaptionbox{\textbf{MNIST.} (Left) CPU Time. (Right) GPU Time.}[.8\textwidth]{
	\centering
	\includegraphics[width=.8\textwidth]{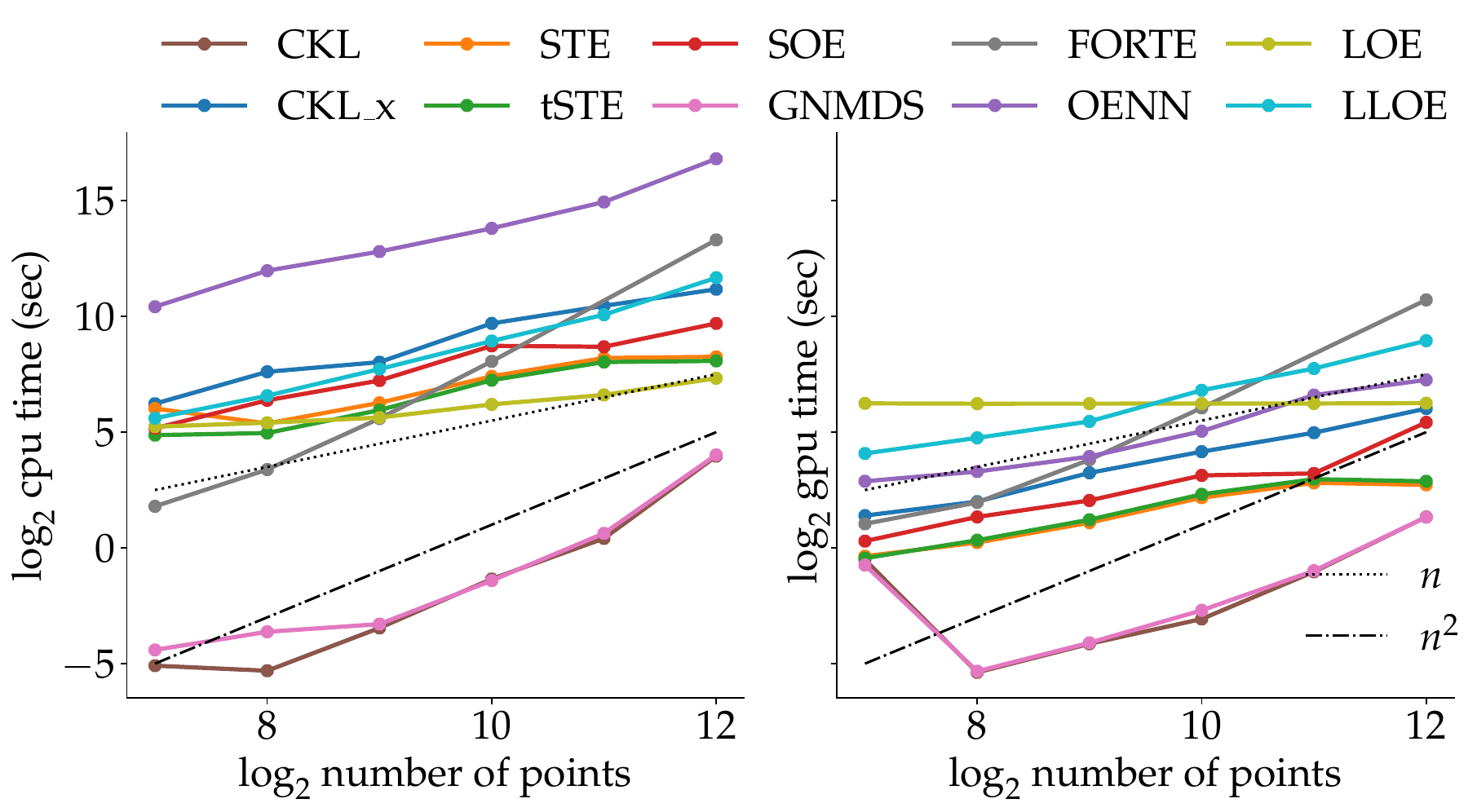}
}
\subcaptionbox{\textbf{Gaussian Mixture.} (Left) CPU Time. (Right) GPU Time.}[.8\textwidth]{
	\centering
	\includegraphics[width=.8\textwidth]{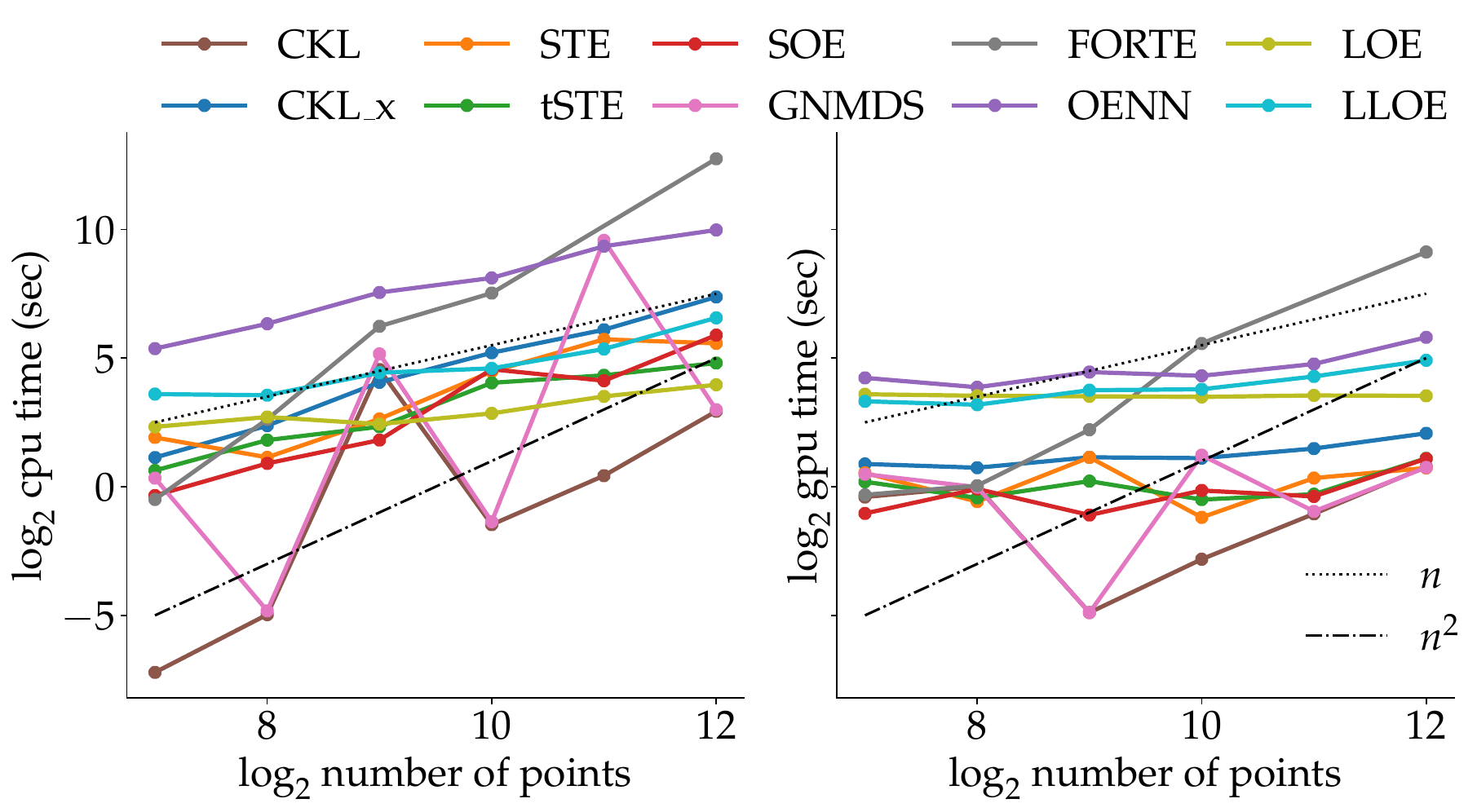}
}

\caption{\label{fig:cpu_gpu}GPU vs CPU running time comparison for two datasets.}
\end{figure}

We compare the running time of various OE algorithms on CPU versus GPU with increasing sample size. We choose small sample sizes in this experiment to avoid prohibitive CPU computational cost. Figure~\ref{fig:cpu_gpu} left and right respectively show the CPU and GPU times of embedding methods for the MNIST dataset. In the log-log plots shown in Figure~\ref{fig:cpu_gpu}, note that the slope of the line determines the growth rate of the running time with increasing $n$. Consistent with theoretical complexity, running time of approaches that optimize over the data matrix and Gram matrix have linear and quadratic growth rate in $n$ respectively. GNMDS and CKL do have low running times since the number of iterations required for convergence to a solution are typically low, which can also be seen in our convergence experiment (see Figure~\ref{fig:convergence}).

The running time of all methods drops drastically on GPUs and it is clear that OE methods can utilize parallel computation in their optimization procedure. Landmark-based approaches, namely LOE and LLOE claim to have better running times, but in our experiments on both GPU and CPU we do not find sufficient evidence to substantiate this claim. LOE appears to have nearly constant running time on GPU. However, this is an artifact of the small sample size used for this experiment to allow of reasonable CPU computation time. This can be observed in our experiment (Section~\ref{sec:various_n}) which investigates the running time of methods with increasing sample size.
\section{DISCUSSION}
We start off by summarizing our key findings as \textbf{rules of thumb} for OE practitioners. In the low noise regime, the clear method of choice given sufficiently many triplets is SOE since it scales comparably in time to other well performing methods and consistently demonstrates superior reconstruction performance. For small number of points and dimensions, you can use any of these non-convex approaches: SOE, STE, T-STE, and OENN. If computational efficiency is a high priority, then on both CPU as well as GPU, unless one can considerably compromise on the quality, SOE, STE, and t-STE are the best candidates. Given sufficiently many triplets, for good reconstruction and for preserving the local neighborhood, SOE is again the top choice. In the large-sample regime or in the high-dimensional regime, SOE is a particularly good choice of OE algorithms since it appears to attain nearly constant test error with increasing sample sizes and dimensions.

When the number of triplets are sufficiently large, it has been theoretically established that the original data can be exactly reconstructed up to similarity transformations \citet{kleindessner2014uniqueness,jain2016finite}. Therefore, with sufficiently many triplets, an algorithm that can achieve near zero train triplet error would also obtain low reconstruction errors. \citet{bower2018landscape} \textbf{conjectures} that the optimization of the hinge triplet loss may not suffer from local optima and that every local optima is also a global optima. The optimization objective used in SOE \citep{terada2014local} is closely related to the hinge triplet loss and we conjecture that this might indeed hold for the optimization objective of SOE as well. Our experiments provide an abundance of empirical evidence to support this hypothesis. Interestingly, even though OENN uses the hinge triplet loss suggested by \citet{bower2018landscape}, unlike SOE, OENN occasionally appears to suffer from local optima. Investigating this more formally could provide valuable insights into understanding non-convex optimization landscapes where local optima are also globally optimal. We leave this for future work.

\newpage
\acks{
This work has been supported by the German Research Foundation
through the Cluster of Excellence
“Machine Learning – New Perspectives for Science" (EXC 2064/1 number
390727645), the Baden-Württemberg
Stiftung through the BW Eliteprogramm for Postdocs, the BMBF Tübingen AI Center (FKZ: 01IS1803
9A), and the International Max Planck Research School for Intelligent Systems (IMPRS-IS).

}


\newpage

\appendix
\section*{Appendix}
\label{app:theorem}



In the appendix materials, we present results and details that were omitted from the main paper. In Section~\ref{sec:description_datasets}, we describe the datasets and present in which experiments they were used. In Section~\ref{sec:experimental_setup}, we provide further details on the experimental setup. Sections \ref{sec:gen_exps_supp} to \ref{supp:increasing_n} contain more results and findings from our experiments. In Section~\ref{sec:method_details}, we formulate the methods and their objectives in more detail. A detailed description of our neural network approach OENN and the simulations used to choose the parameters of OENN are provided in Section~\ref{sec:method_details}.

\section{DESCRIPTION OF DATASETS}
\label{sec:description_datasets}

The following, we describe the datasets that we use in our experiments. In Table~1, we visualize in experiments they were employed.

\paragraph{2D datasets.} For the reconstruction of 2D datasets with ordinal embedding methods, we use various datasets collected from~\citet{ClusteringDatasets}:
\begin{itemize}
    \item \textbf{Birch1} and \textbf{Birch3}~\citep{Birchsets}, with $n=10,000$ each, are artificial datasets with cluster structure: Birch1 has clusters in a regular grid, and Birch3 contains random sized clusters in random locations. However, the clusters are unlabeled.
    \item We use a collection of shape sets, containing \textbf{Aggregation}~\citep{Aggregation} with $n=788$ and $k=7$ clusters, \textbf{Compound}~\citep{Compound} with $n=399$ and $k=6$ clusters, \textbf{Path Based}~\citep{Path_and_Spiral} with $n=300$ and $k=3$ clusters, and \textbf{Spiral}~\citep{Path_and_Spiral} with $n=312$ and $k=3$ clusters. 
    \item \textbf{Worms}~\citep{Worms} with $n=105,600$, is a synthetic dataset containing shapes reminiscent of worms (unlabeled). 
\end{itemize}

\paragraph{CHAR~\citep{char_covtype}.} This dataset contains $8\times 8$ images of $10$ digits. Here, we use the test set where $n=1,797$.

\paragraph{USPS~\citep{usps}.} This dataset contains $16\times 16$ handwritten images of $10$ digits. We use the train set with $n=7,291$ points.

\paragraph{MNIST~\citep{mnist}.} The classic MNIST dataset contains $28\times 28$ handwritten images of $10$ digits. Again, we use the train set with $n=60,000$.

\paragraph{FMNIST~\citep{fmnist}.} The FMNIST dataset contains $28\times 28$ greyscale images of fashion items. Just as MNIST, it contains $n=60,000$ points in the train set, and consists of $10$ classes.

\paragraph{KMNIST~\citep{kmnist}.} The Kuzushiji-MNIST train set contains $n=60,000$ $28\times 28$ images of $10$ Kuzushiji (cursive Japanese) characters. For the experiment with increasing embedding dimension, we use a random subset of $10,000$ points. 

\paragraph{COVTYPE~\citep{char_covtype}.} This dataset contains $54$ cartographic variables to predict one of $4$ forest cover types. We use the whole dataset of $n=581,012$ observations.

\paragraph{Gaussian Mixture, GMM.} This datasets consists of two clearly separable normal distribution with unit variance, $\mathcal{N}(\mu_1,1)$ and $\mathcal{N}(\mu_2,1)$, where $\mu_1= 0$ is the zero vector, and $\mu_2 = 100\times\mathbf{1}$. The dimension of the dataset can be chosen. 

\paragraph{Uniform.} For this dataset, we draw the required number of points uniformly from $[0,10]^d$, where the dimension $d$ can be chosen. 

\begin{table*}[h]
    \caption{The datasets we use to evaluate all the algorithms in our experiments. Ticks mark the inclusion of a dataset in a particular experiment. The tuple \textbf{(n, org\_d, emb\_d)} summarizes the \textbf{number of points n}, the \textbf{original dimension org\_d}, and the default \textbf{embedding dimension emb\_d}. \label{tab:datasets}}
    \centering
    \resizebox{\textwidth}{!}
    {\renewcommand{\arraystretch}{2.3}
    \begin{tabular}{ m{1.8cm} 
     >{\centering\arraybackslash} m{2.65cm}
    >{\centering\arraybackslash} m{1.4cm} >{\centering\arraybackslash} m{1.4cm} >{\centering\arraybackslash} m{1.4cm} >{\centering\arraybackslash} m{1.4cm} >{\centering\arraybackslash} m{1.4cm} >{\centering\arraybackslash} m{1.7cm} >{\centering\arraybackslash} m{1.4cm} } 
        {\bf Dataset} & \makecell{\small\bf (n, org\_d, emb\_d) \\ {\small \bf n in 1000's}} & \makecell{{\small\bf General} \\ {\small\bf Exp.}} & \makecell{{\small\bf Increase} \\ {\small \bf of}\\ {\small \bf triplets}} & \makecell{{\small\bf Increase} \\ {\small \bf of}\\ {\small \bf points}} & \makecell{{\small\bf Increase} \\ {\small \bf of}\\ {\small \bf emb\_d}} & \makecell{{\small\bf Increase} \\ {\small \bf of}\\ {\small \bf org\_d}}  & \makecell{{\small\bf Convergence} \\ {\small \bf Exp.}}  & \makecell{{\small\bf CPU} \\ {\small \bf vs.} \\ {\small\bf GPU}}  \\
        \toprule
        {\small \bf 2D Data } $\times7$& {{($\sim$0.3-106, 2, 2}) } & &  \checkmark & & & & & \\ 
        \hline
        {\bf CHAR } & {($\sim$1.8, 64, 30) } & \checkmark & \parbox{1.9cm}{\centering \checkmark \\ {\scriptsize emb\_d = 50}} & & & & \checkmark & \\ 
        \hline
        {\bf GMM } & {(5, 2, 2) } & & & & & \checkmark & & \checkmark \\ 
        \hline
        {\bf USPS} & {($\sim$7.3, 256, 50) } & \checkmark & \checkmark & & \checkmark & & \checkmark & \\ 
        \hline
       {\bf MNIST } & {(60, 784, 50) } & \checkmark & \checkmark & & & \checkmark & \checkmark & \parbox{1.9cm}{\centering \checkmark \\ {\scriptsize emb\_d = 30}}\\
        \hline
       {\bf FMNIST } & {(60, 784, 50) } & \checkmark & & & & & \checkmark & \\ 
        \hline
        {\bf KMNIST } & {(10, 784, 50) } & & & & \checkmark & & & \\ 
        \hline
        { \bf COVTYPE} & {($\sim$581, 54, 30) } & \checkmark &  & \checkmark & & & \checkmark & \\ 
        \hline
        {\bf Uniform} & { ($10$, 2, 2) } & \checkmark &  \checkmark & \checkmark & \parbox{1.9cm}{\centering \checkmark \\ {\scriptsize org\_d = 10}} & &  & \\ 
        \bottomrule
    \end{tabular}}
\end{table*}

\section{EXPERIMENTAL SETUP}
\label{sec:experimental_setup}

\subsection{Hardware.}

All our experiments are conducted on one of three hardware settings. Most methods are provided with 4 CPUs and 1 GPU, while OENN requires 4 GPUs for all experiments. Due to the high number of points and the high embedding dimension, all experiments for the Covtype dataset are run on GPUs with more memory. 

\textbf{Hardware Setting 1 for most experiments:} Intel Xeon Gold 6240 @ 2.60GHz and NVIDIA GeForce RTX 2080-Ti 11 GB.

\textbf{Hardware Setting 2 for Covtype experiments:} Intel Core Processors (Broadwell) @ 2GHz and NVIDIA Tesla V100 SXM2 32GB.

\textbf{Hardware Setting 3 for CPUvsGPU experiments:} Intel Xeon E5-2650 v4 @ 2.2 GHz and NVIDIA GTX 1080-Ti 11 GB.

For the details on the hyperparameters selected in each method, we refer to the descprition of each method in Section~\ref{sec:method_details}.

\subsection{Hyperparamters.}

For all methods except OENN, we fixed the batch size to $\min(\textrm{\#triplets}, \textrm{1 million})$ and we performed a grid search for the learning rate and other parameters on $1000$ points of uniform data. We shortly summarize the results here. Due to the more complex choice of hyperparameters for OENN, we describe them in detail in Section~\ref{sec:simWidth}. 
\begin{itemize}
    \item \textbf{GNMDS.} For GNMDS we use a regularization of $\lambda = 0$ and a learning rate of $10$.
    \item \textbf{FORTE.} The learning rate is $100$. Similar to the implementation of \citet{jain2016finite}, we use the backtracking linear search parameter $\rho = 0.5$  and the Amarijo stopping condition parameter $c1 = 0.0001$ for the line-search optimization.
    \item \textbf{CKL.} For CKL we use a regularization of $\lambda = 0$ and a learning rate of $100$.
    \item \textbf{SOE, STE, tSTE, and CKL\_x} use a learning rate of $1$. 
    \item \textbf{LOE.} For LOE, the learning rate is $0.0001$.
    \item \textbf{LLOE.} The phase one learning rate is $1$, and the phase two learning rate is $0.5$. 
\end{itemize}

\newpage
\section{GENERAL EXPERIMENTS RESULTS}
\label{sec:gen_exps_supp}
\begin{table*}[!htb]
    \caption{Detailed results of the general experiments. Each cell shows \textbf{(test triplet error, kNN error)}, and below the running time. The Gram matrix based methods were not used on large datasets due to memory issues. \label{tab:general_experiments}}
    \centering
    {\renewcommand{\arraystretch}{1.8}
    \begin{tabular}{ m{1.0cm} 
    >{\centering\arraybackslash} m{2cm}
    >{\centering\arraybackslash} m{2cm} >{\centering\arraybackslash} m{2cm} >{\centering\arraybackslash} m{2cm} >{\centering\arraybackslash} m{2cm}
    >{\centering\arraybackslash} m{2cm}} 
        {\small\bf Dataset} & \makecell{\small\bf MNIST} & \makecell{{\small\bf COVTYPE}} & \makecell{{\small\bf FMNIST}} & \makecell{{\small\bf USPS}} & \makecell{{\small\bf CHAR }} & \makecell{{\small\bf UNIFORM}}\\
        \toprule
{\small \bf SOE}      & \makecell{ \small (0.03, 0.04) \\ \small11.2min}   & \makecell{ \small (0.02, 0.29) \\ \small58.3min}   & \makecell{ \small (0.03, 0.2) \\ \small26.1min}      & \makecell{ \small (0.02, 0.04) \\ \small61s}      & \makecell{ \small (0.03, 0.03) \\ \small6s}       & \makecell{ \small (0.02, 0.0) \\ \small1s}\\ \hline
{\small \bf OENN}    & \makecell{ \small (0.13, 0.13) \\ \small165.0min}      & \makecell{ \small (0.02, 0.13) \\ \small6.7h}     & \makecell{ \small (0.06, 0.28) \\ \small6.2h}  & \makecell{ \small (0.04, 0.05) \\ \small16.6min}    & \makecell{ \small (0.04, 0.02) \\ \small206s}     & \makecell{ \small (0.36, 0.0) \\ \small122s}\\ \hline
{\small \bf STE}      & \makecell{ \small (0.11, 0.08) \\ \small51.3min}    & \makecell{ \small (0.15, 0.5) \\ \small22.5min}   & \makecell{ \small (0.07, 0.28) \\ \small5.2min}     & \makecell{ \small (0.04, 0.05) \\ \small247s}      & \makecell{ \small (0.04, 0.02) \\ \small4s}     & \makecell{ \small (0.02, 0.0) \\ \small0.6s}\\ \hline
{\small \bf tSTE}      & \makecell{ \small (0.2, 0.24) \\ \small29.5min}  & \makecell{ \small (0.24, 0.53) \\ \small170.0min}  & \makecell{ \small (0.15, 0.39) \\ \small17.6min}      & \makecell{ \small (0.05, 0.06) \\ \small31s}      & \makecell{ \small (0.04, 0.03) \\ \small5s}       & \makecell{ \small (0.05, 0.0) \\ \small2s}\\ \hline
{\small \bf CKL\_x}    & \makecell{ \small (0.16, 0.17) \\ \small5.2min}   & \makecell{ \small (0.06, 0.39) \\ \small22.3min}   & \makecell{ \small (0.11, 0.28) \\ \small5.2min}      & \makecell{ \small (0.14, 0.15) \\ \small55s}     & \makecell{ \small (0.16, 0.08) \\ \small18s}      & \makecell{ \small (0.11, 0.0) \\ \small34s}\\ \hline
{\small \bf LOE}      & \makecell{ \small (0.26, 0.08) \\ \small65.2min}   & \makecell{ \small (0.24, 0.51) \\ \small42.4min}  & \makecell{ \small (0.16, 0.29) \\ \small66.7min}  & \makecell{ \small (0.22, 0.12) \\ \small10.7min}  & \makecell{ \small (0.23, 0.07) \\ \small5.2min}       & \makecell{ \small (0.19, 0.0) \\ \small6s}\\ \hline
{\small \bf LLOE}        & \makecell{ \small (0.31, 0.25) \\ \small7.2h}      & \makecell{ \small (0.04, 0.4) \\ \small27.9h}      & \makecell{ \small (0.1, 0.24) \\ \small6.8h}  & \makecell{ \small (0.25, 0.14) \\ \small27.5min}    & \makecell{ \small (0.23, 0.06) \\ \small153s}      & \makecell{ \small (0.07, 0.0) \\ \small84s}\\ \hline
{\small \bf GNMDS}                                                   & -                                                  & -                                                 & -      & \makecell{ \small (0.35, 0.44) \\ \small25s}     & \makecell{ \small (0.2, 0.07) \\ \small0.7s}      & \makecell{ \small (0.33, 0.0) \\ \small10s}\\ \hline
{\small \bf FORTE}                                                   & -                                                  & -                                                 & -     & \makecell{ \small (0.06, 0.06) \\ \small5.5h}  & \makecell{ \small (0.07, 0.03) \\ \small7.3min}  & \makecell{ \small (0.04, 0.0) \\ \small35.4min}\\ \hline
{\small \bf CKL }                                                    & -                                                  & -                                                 & -      & \makecell{ \small (0.35, 0.45) \\ \small25s}      & \makecell{ \small (0.21, 0.08) \\ \small1s}      & \makecell{ \small (0.16, 0.0) \\ \small10s}\\ \hline
    \end{tabular}}
\end{table*}

\newpage
\FloatBarrier
\section{CONVERGENCE EXPERIMENTS}
\label{sec:convergence_supp}
\begin{figure}[!htb]
\centering
\subcaptionbox{FMNIST10k}[.32\textwidth]{
	\centering
	\includegraphics[width=.32\textwidth]{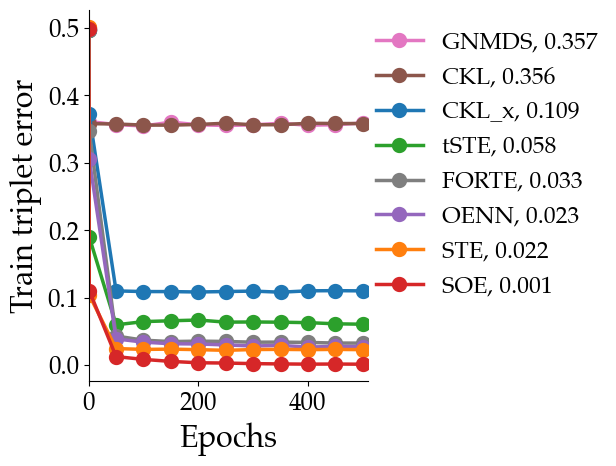}
}
\subcaptionbox{FMNIST}[.32\textwidth]{
	\centering
	\includegraphics[width=.32\textwidth]{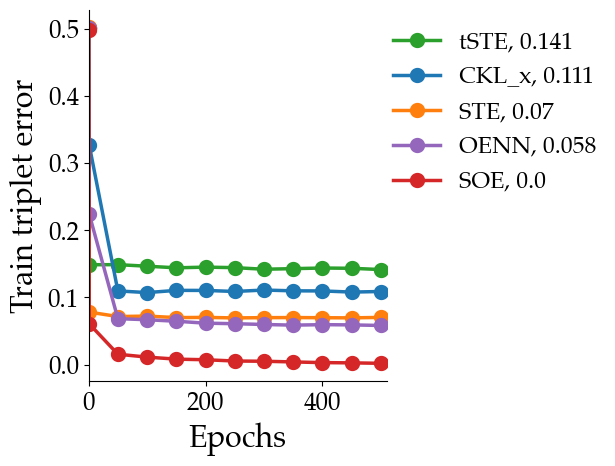}
}
\subcaptionbox{Char}[.32\textwidth]{
	\centering
	\includegraphics[width=.32\textwidth]{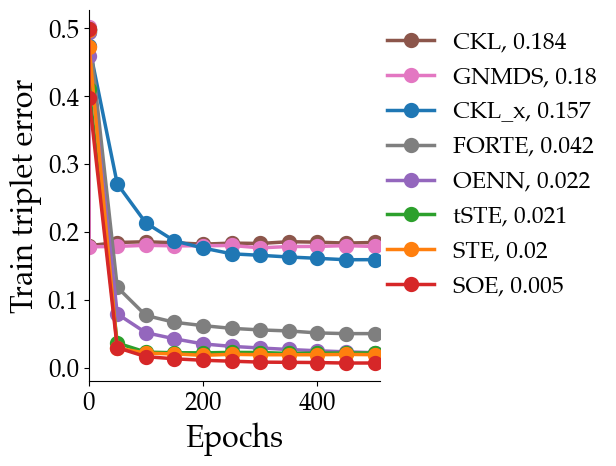}
}\hspace*{\fill}

\subcaptionbox{MNIST}[.32\textwidth]{
	\centering
	\includegraphics[width=.32\textwidth]{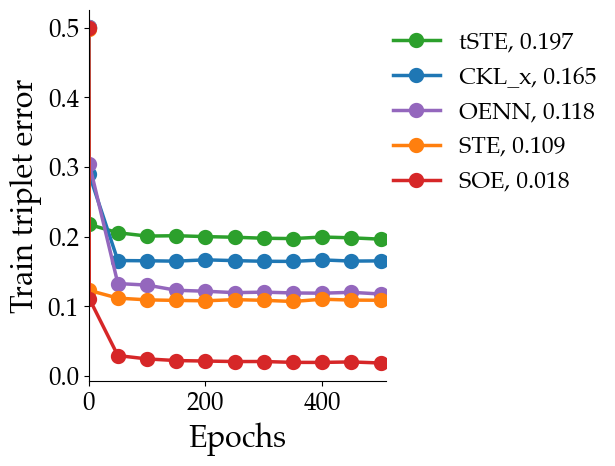}
}
\subcaptionbox{USPS}[.32\textwidth]{
	\centering
	\includegraphics[width=.32\textwidth]{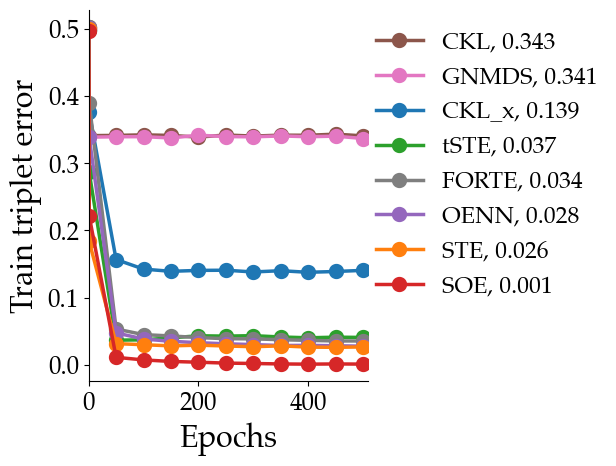}
}
\subcaptionbox{Covertype}[.32\textwidth]{
	\centering
	\includegraphics[width=.32\textwidth]{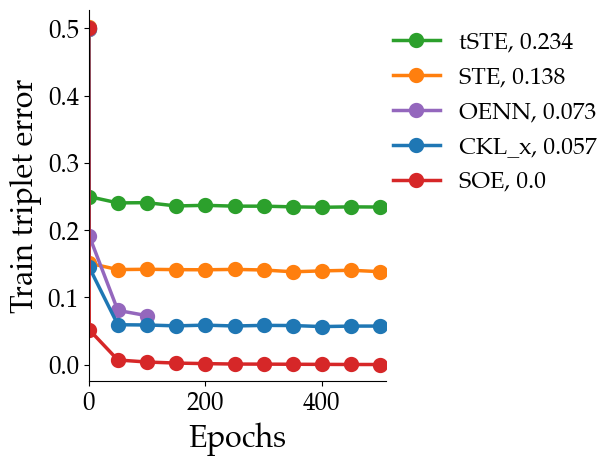}
}\hspace*{\fill}
\caption{\textbf{Convergence Experiments.} The train triplet error of various methods is plotted with respect to the running epoch. The final train triplet error value of each method is shown in the legend of each plot.}
\end{figure}

\FloatBarrier
\section{INCREASING TRIPLETS}
\label{sec:increasing_trips_supp}
\textbf{Other findings from this experiment.}
\begin{itemize}
    \item Most algorithms demonstrate consistently poor generalization performance in the low triplet regime as measured by the Procrustes error as well as the kNN classification error even though they attain good training triplet errors. For instance, FORTE is the only gram matrix based approach that is competitive with the well-performing algorithms that optimize over the gram matrix. However, in the low triplet regime, it consistently is among the methods with the worst generalization performance even though it achieves good training triplet error.
    \item  Preliminary evaluation in \citet{van2012stochastic}, which compared t-STE with STE, seemed to suggest that t-STE better preserves the local neighborhood. Our extensive evaluation finds some evidence to the contrary. For seven out of the ten datasets, in both the small and the large triplet regimes, STE outperforms t-STE on both Procrustes error as well as on kNN classification error.
  
    \item Across most of the datasets, the worst performing methods are CKL and GNMDS. This trend can be observed in other experiments as well: Gram matrix approaches are worst performing methods with the exception of FORTE. Interestingly, the variant of CKL that optimizes over the embedding directly, CKL\_x, seems to be much better at preserving local neighborhoods than at reconstructing the embedding.
 \end{itemize}
 
\begin{figure*}[!htb]
\centering
\includegraphics[width=.95\textwidth]{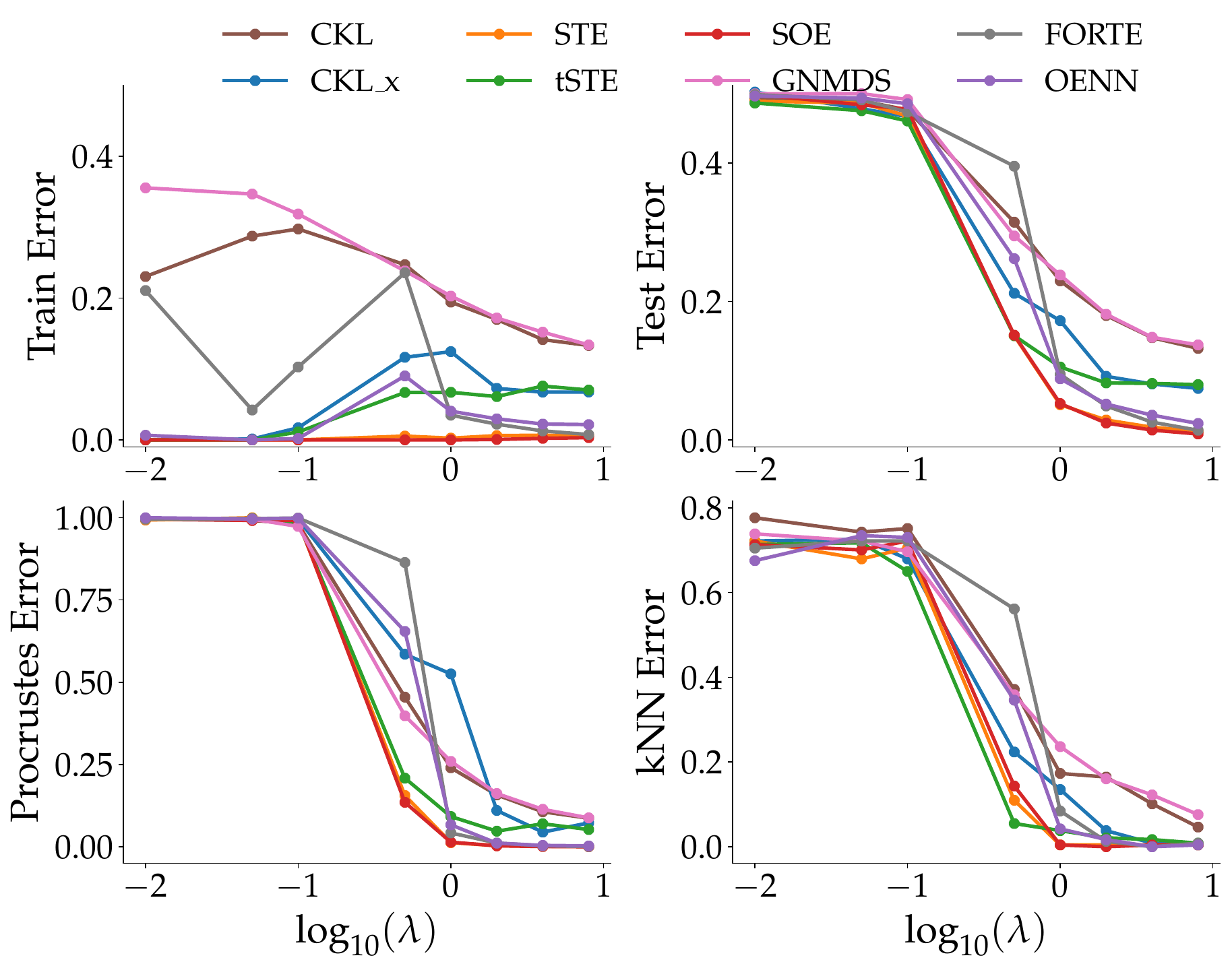}
\caption{{\bfseries Aggregation: }Increasing number of triplets. }
\end{figure*}

\begin{figure*}[!htb]
\centering
\subcaptionbox*{}[.4\textwidth]{
	\centering
	\includegraphics[width=.4\textwidth]{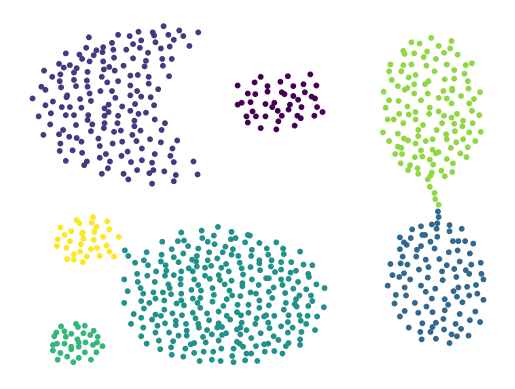}
}
\caption{Aggregation Original. }
\end{figure*}

\begin{figure*}[!htb]
\centering
{\bfseries Aggregation}\par\medskip
\includegraphics[width=.99\textwidth]{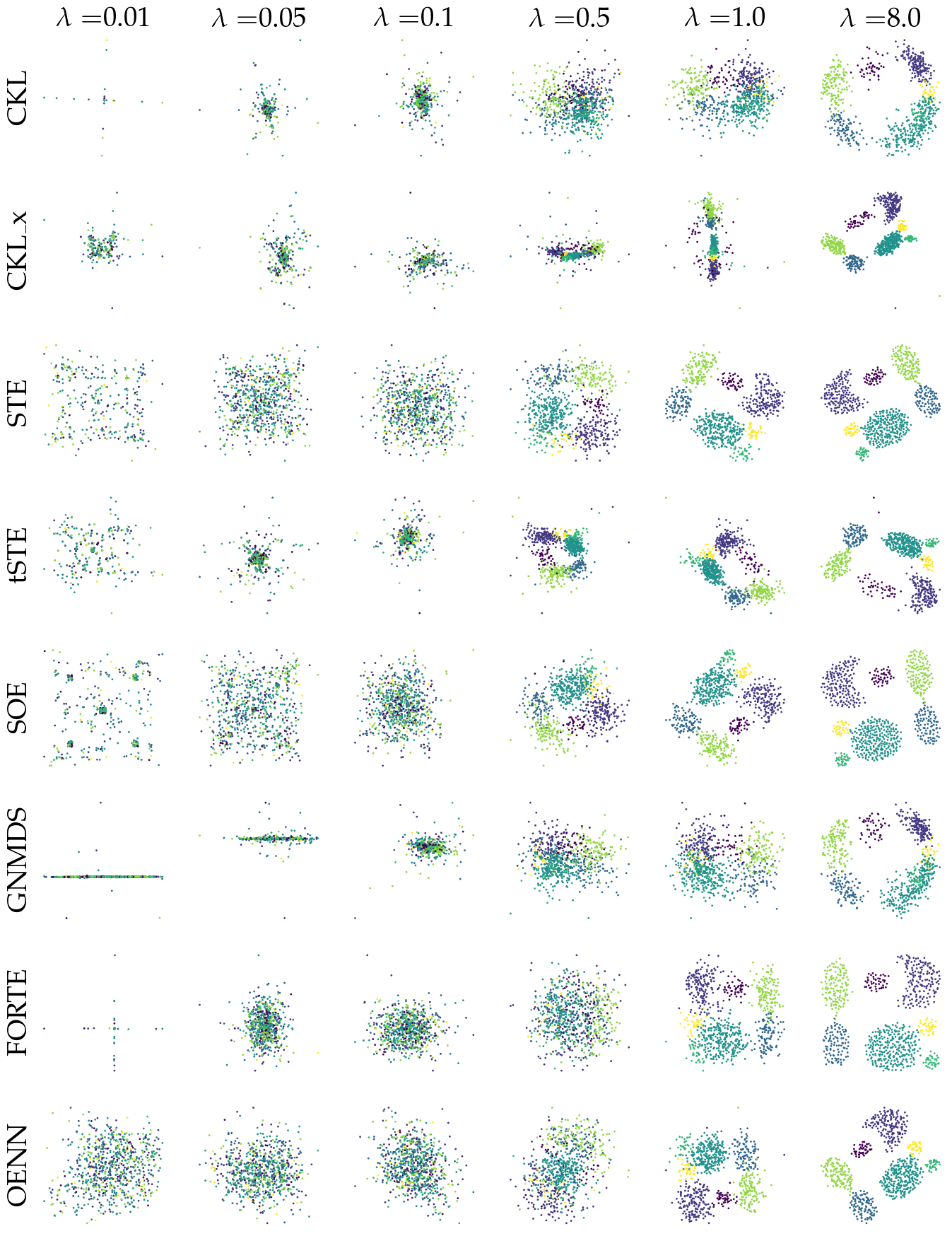}
\caption{\textbf{Aggregation: }Increasing the number of triplets with a triplet multiplier $\lambda$ yieding $\lambda d n \log n$ triplets. }
\end{figure*}

\begin{figure*}[!htb]
\centering
\includegraphics[width=.95\textwidth]{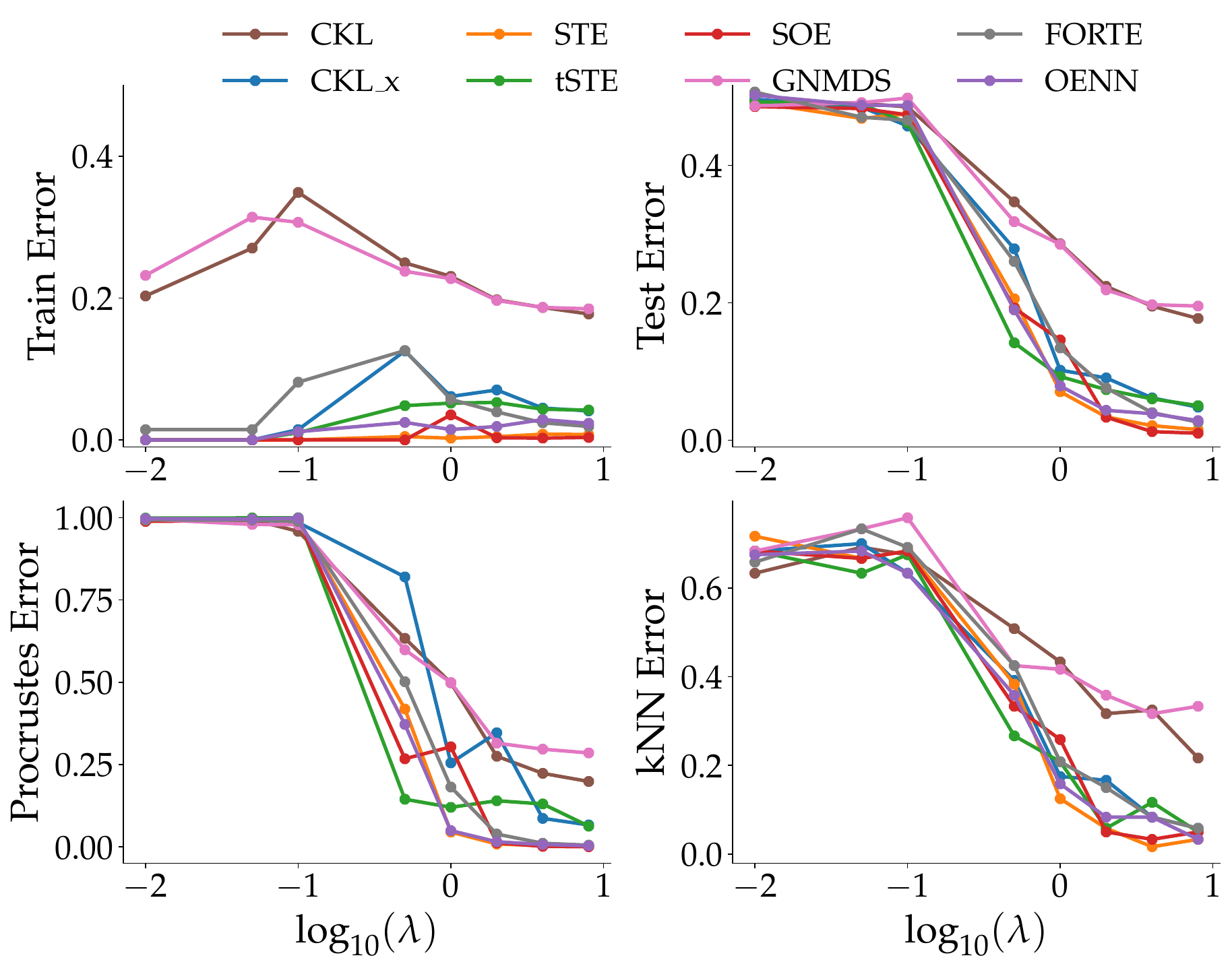}
\caption{{\bfseries Compound: }Increasing number of triplets. }
\end{figure*}

\begin{figure*}[!htb]
\centering
\subcaptionbox*{}[.4\textwidth]{
	\centering
	\includegraphics[width=.4\textwidth]{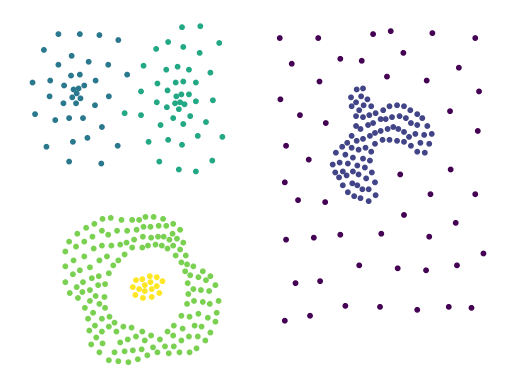}
}
\caption{Compound Original. }
\end{figure*}

\begin{figure*}[!htb]
\centering
{\bfseries Compound}\par\medskip
\includegraphics[width=.99\textwidth]{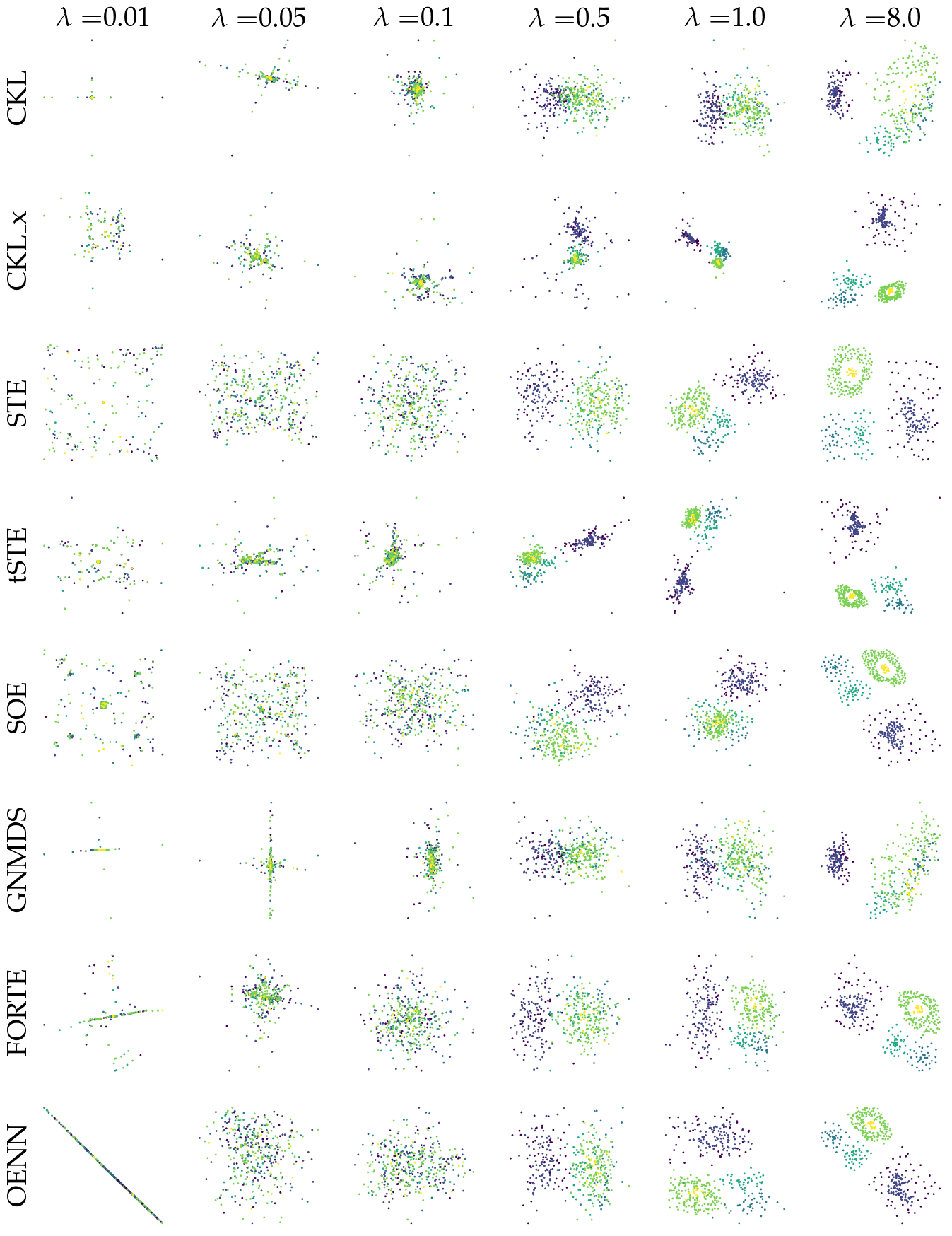}
\caption{\textbf{Compound: }Increasing the number of triplets with a triplet multiplier $\lambda$ yieding $\lambda d n \log n$ triplets. }
\end{figure*}

\begin{figure*}[!htb]
\centering
\includegraphics[width=.95\textwidth]{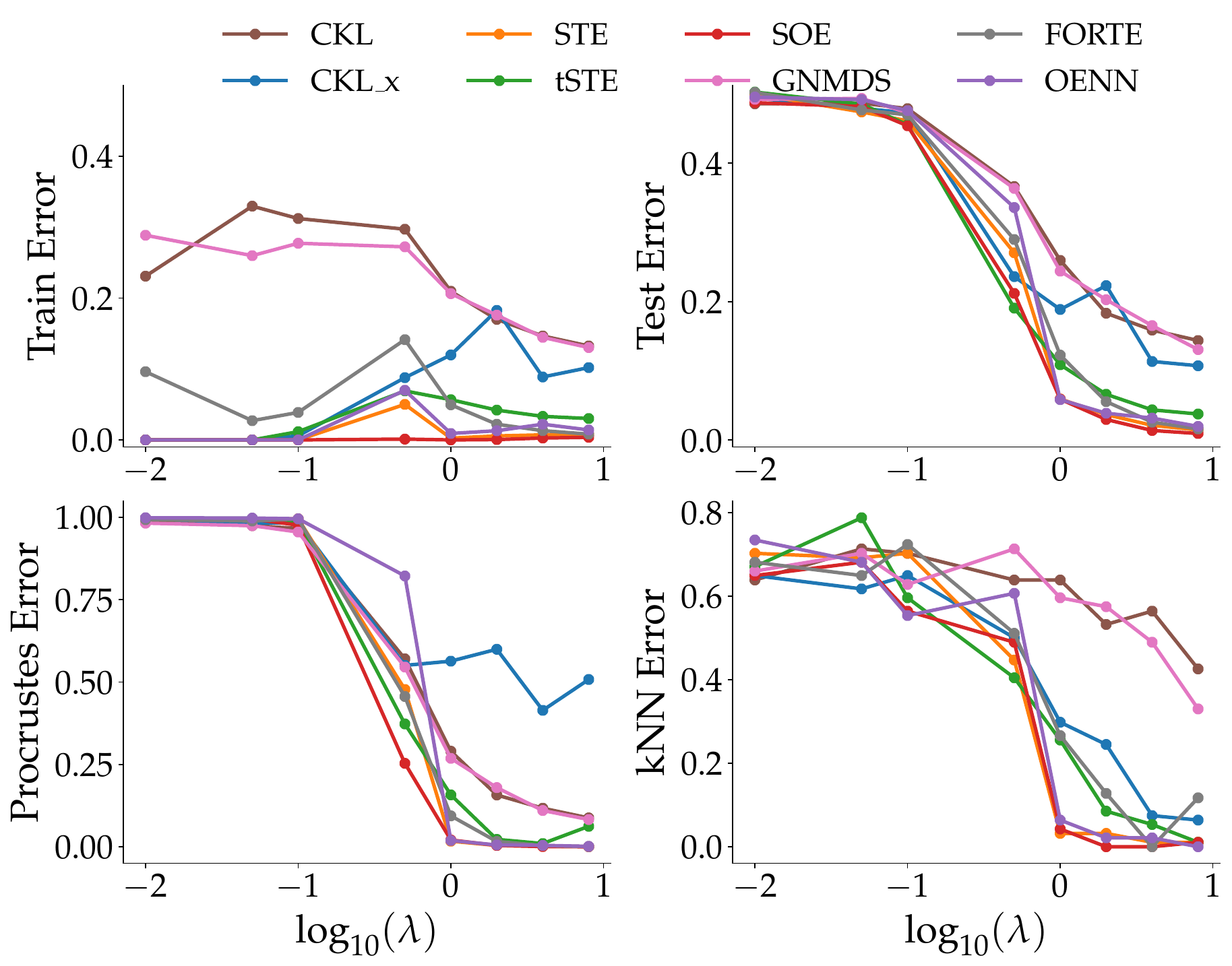}
\caption{{\bfseries Spiral: }Increasing number of triplets.}
\end{figure*}

\begin{figure*}[!htb]
\centering
\subcaptionbox*{}[.4\textwidth]{
	\centering
	\includegraphics[width=.4\textwidth]{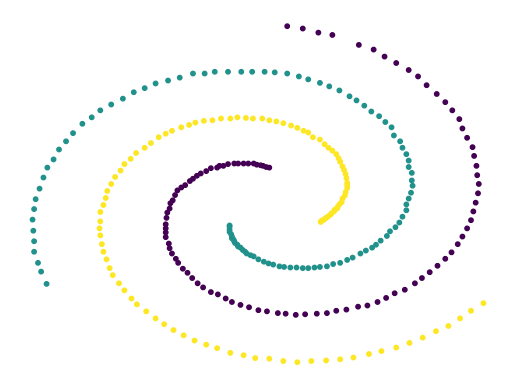}
}
\caption{Spiral Original. }
\end{figure*}

\begin{figure*}[!htb]
\centering
{\bfseries Spiral}\par\medskip
\includegraphics[width=.99\textwidth]{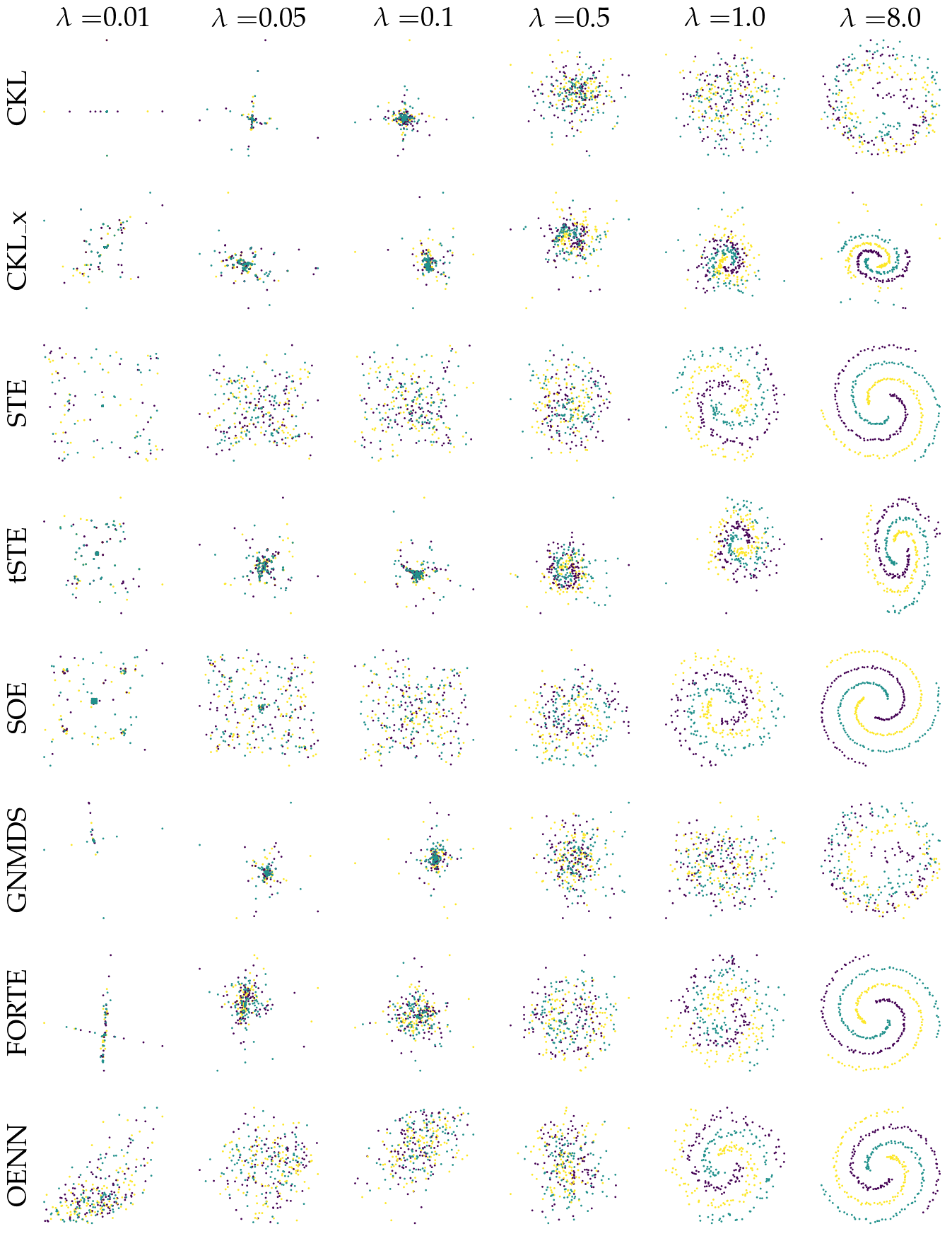}
\caption{\textbf{Spiral: }Increasing the number of triplets with a triplet multiplier $\lambda$ yieding $\lambda d n \log n$ triplets. }
\end{figure*}

\begin{figure*}[!htb]
\centering
\includegraphics[width=.95\textwidth]{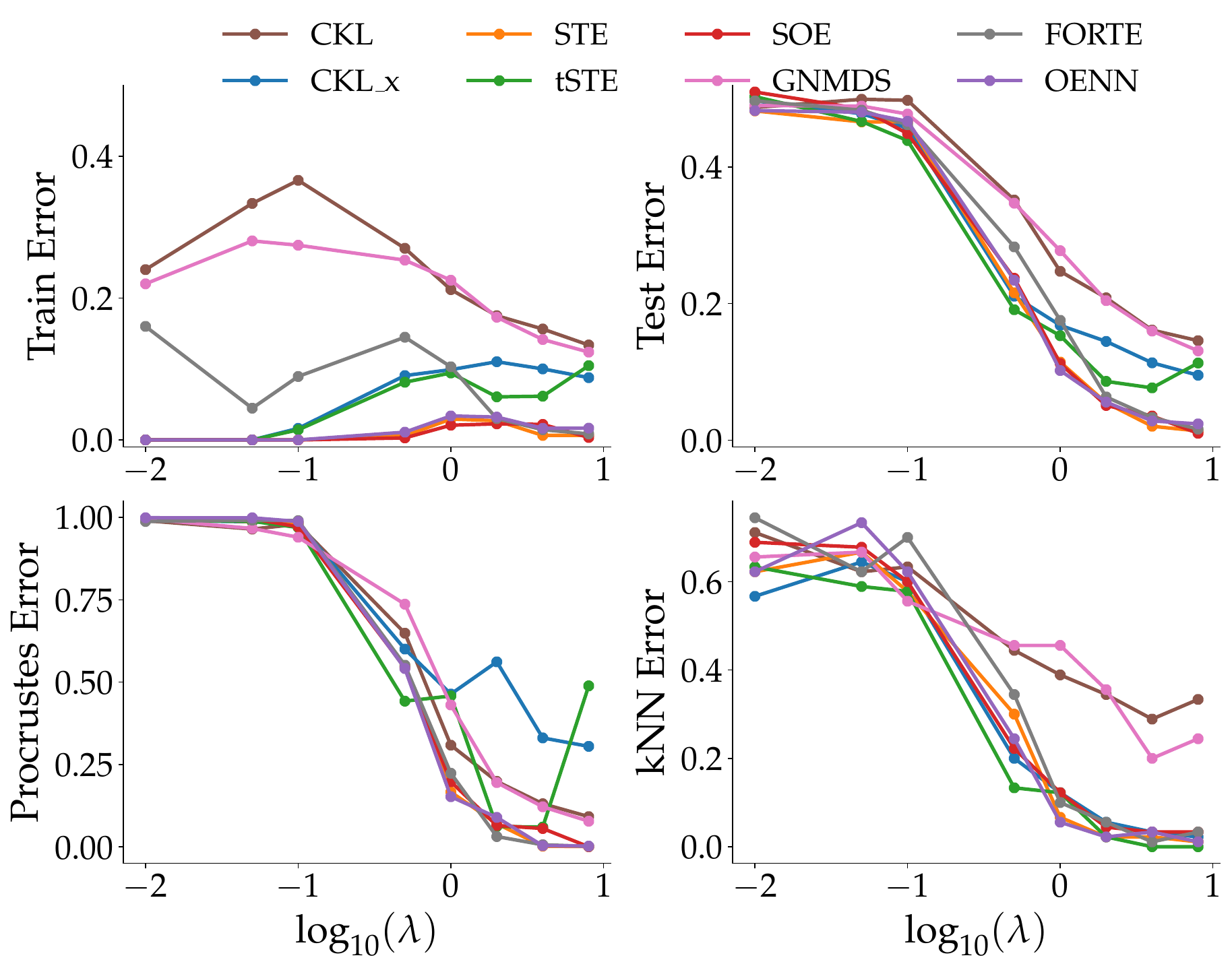}
\caption{{\bfseries Path-based: }Increasing number of triplets. }
\end{figure*}

\begin{figure*}[!htb]
\centering
\subcaptionbox*{}[.4\textwidth]{
	\centering
	\includegraphics[width=.4\textwidth]{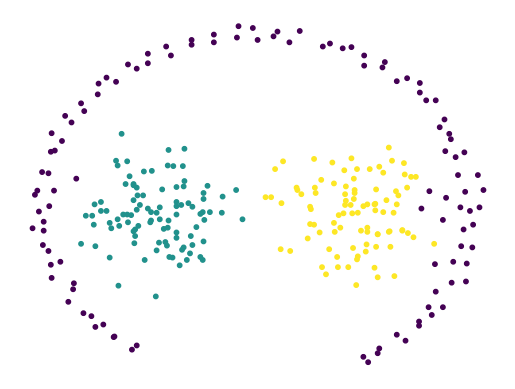}
}
\caption{Path-based Original. }
\end{figure*}

\begin{figure*}[!htb]
\centering
{\bfseries Path-based}\par\medskip
\includegraphics[width=.99\textwidth]{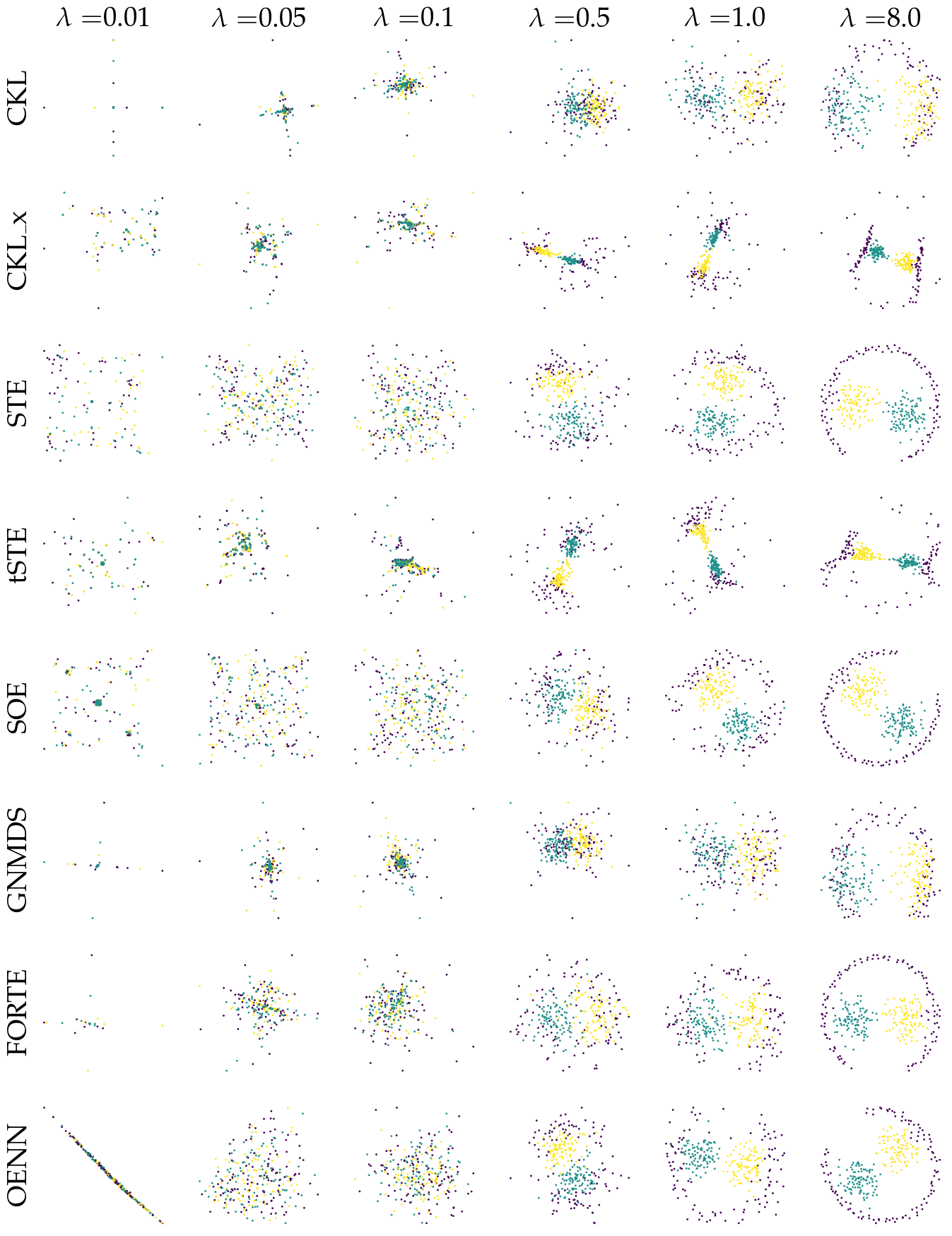}
\caption{\textbf{Path-based: }Increasing the number of triplets with a triplet multiplier $\lambda$ yieding $\lambda d n \log n$ triplets. }
\end{figure*}

\begin{figure*}[!htb]
\centering
\includegraphics[width=.95\textwidth]{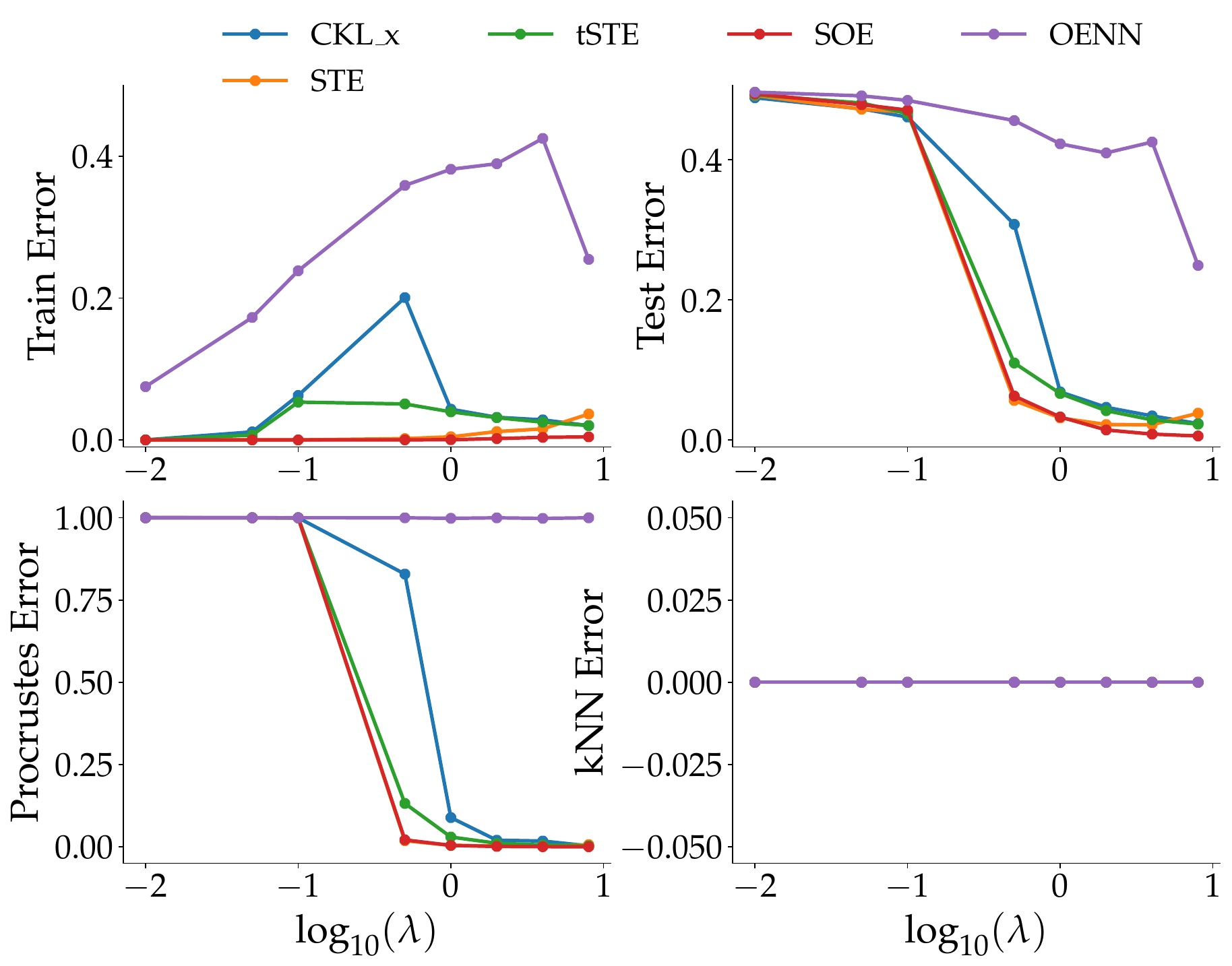}
\caption{{\bfseries Birch1: }Increasing number of triplets. }
\end{figure*}

\begin{figure*}[!htb]
\centering
\subcaptionbox*{}[.4\textwidth]{
	\centering
	\includegraphics[width=.4\textwidth]{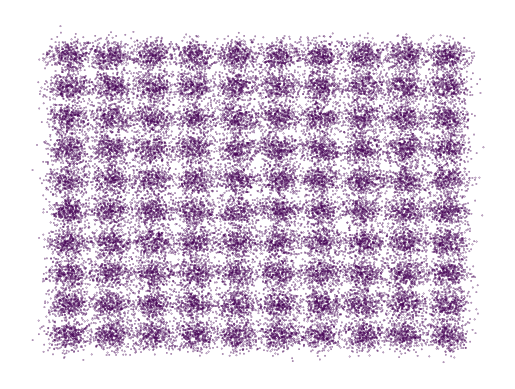}
}
\caption{Birch1 Original. }
\end{figure*}

\begin{figure*}[!htb]
\centering
{\bfseries Birch1}\par\medskip
\includegraphics[width=.99\textwidth]{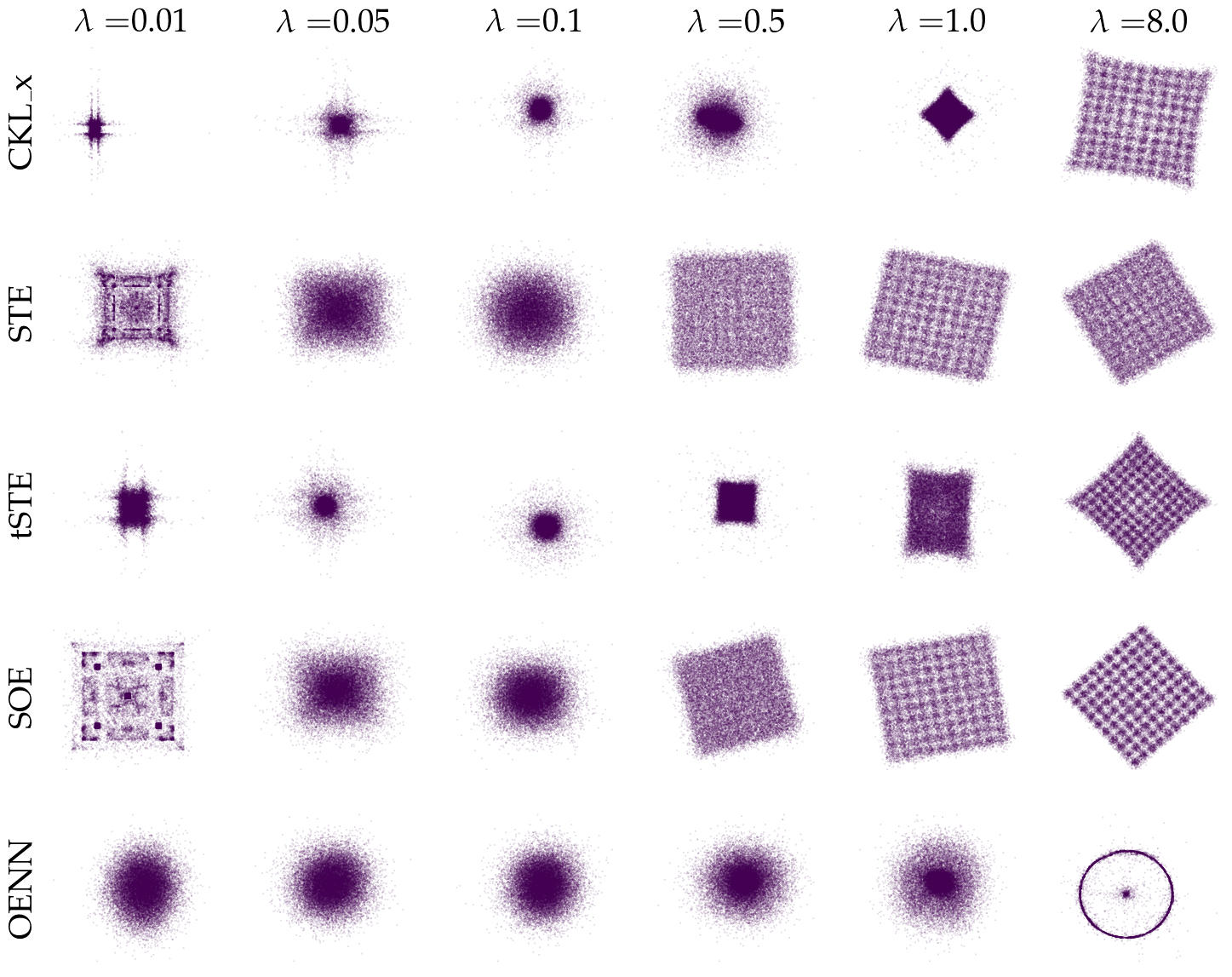}
\caption{\textbf{Birch1: }Increasing the number of triplets with a triplet multiplier $\lambda$ yieding $\lambda d n \log n$ triplets. }
\end{figure*}

\begin{figure*}[!htb]
\centering
\includegraphics[width=.95\textwidth]{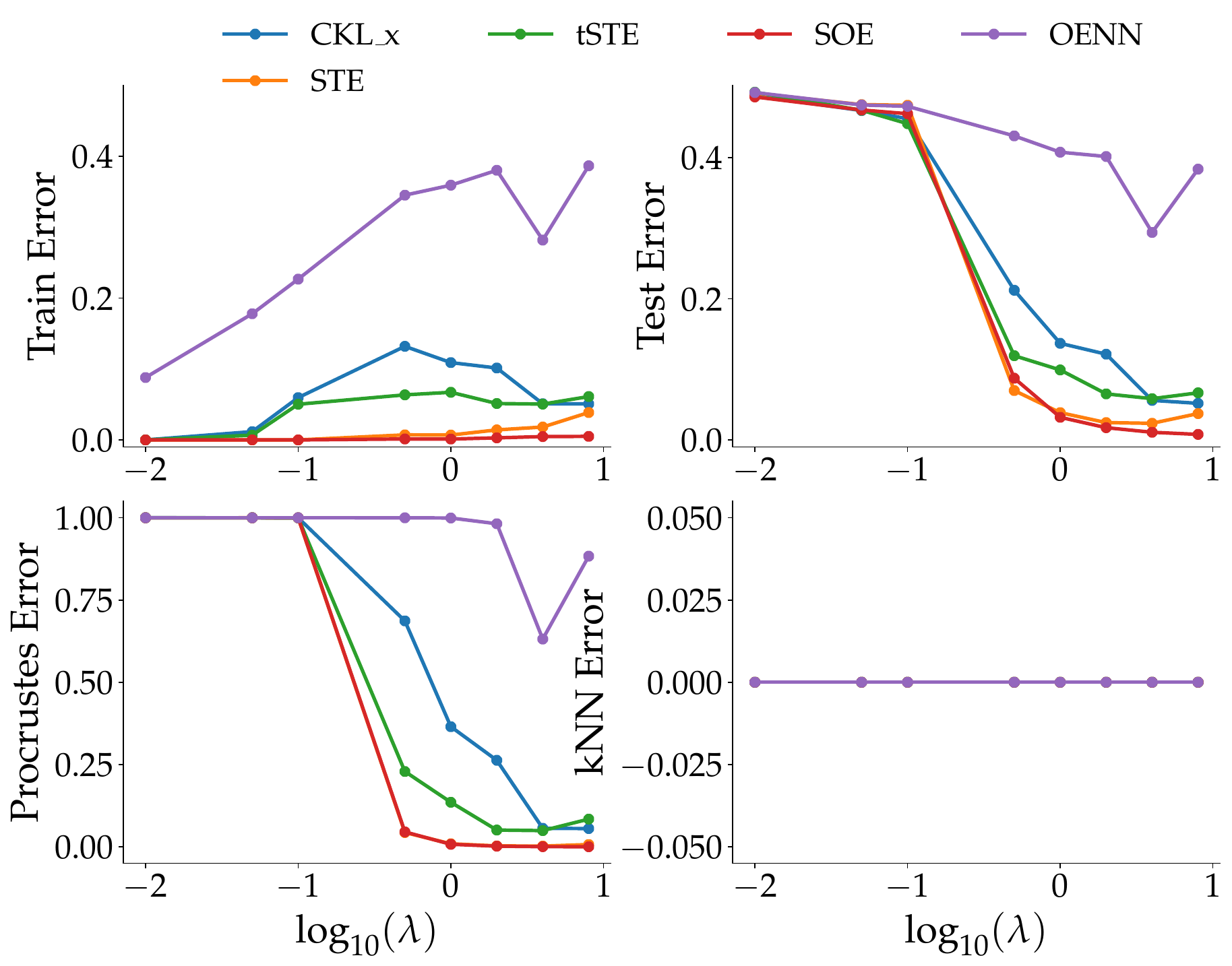}
\caption{{\bfseries Birch3: }Increasing number of triplets. }
\end{figure*}
\begin{figure*}[!htb]
\centering
\subcaptionbox*{}[.4\textwidth]{
	\centering
	\includegraphics[width=.4\textwidth]{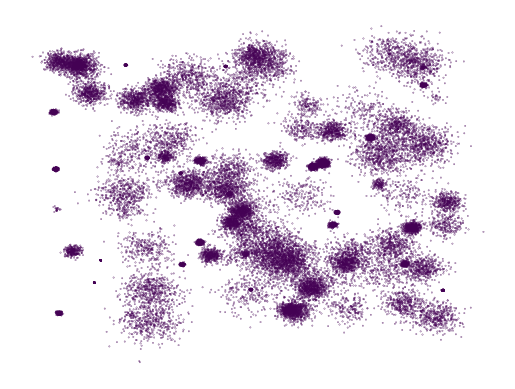}
}
\caption{Birch3 Original.}
\end{figure*}

\begin{figure*}[!htb]
\centering
{\bfseries Birch3}\par\medskip
\includegraphics[width=.99\textwidth]{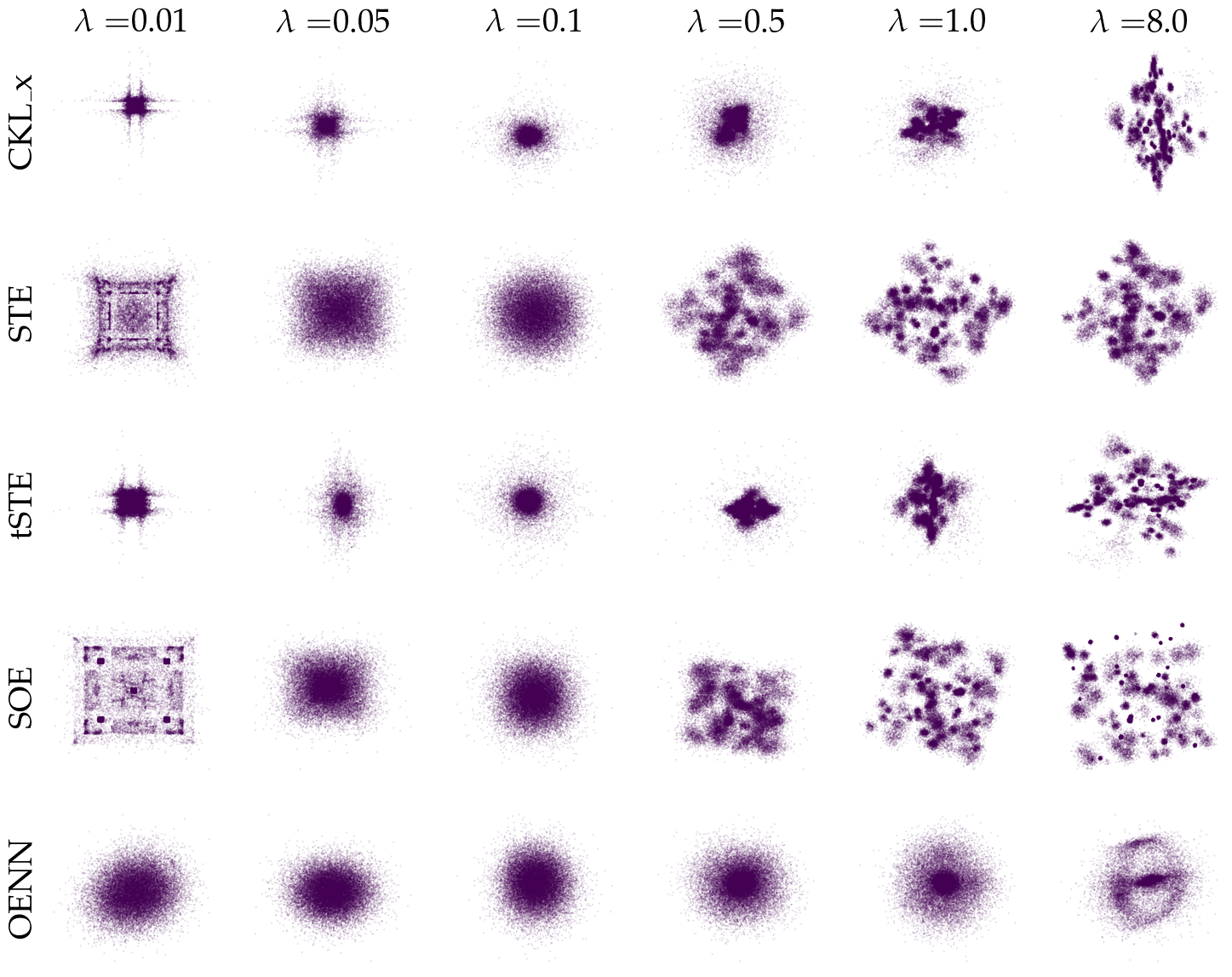}
\caption{\textbf{Birch3: }Increasing the number of triplets with a triplet multiplier $\lambda$ yieding $\lambda d n \log n$ triplets. }
\end{figure*}

\begin{figure*}[!htb]
\centering
\includegraphics[width=.95\textwidth]{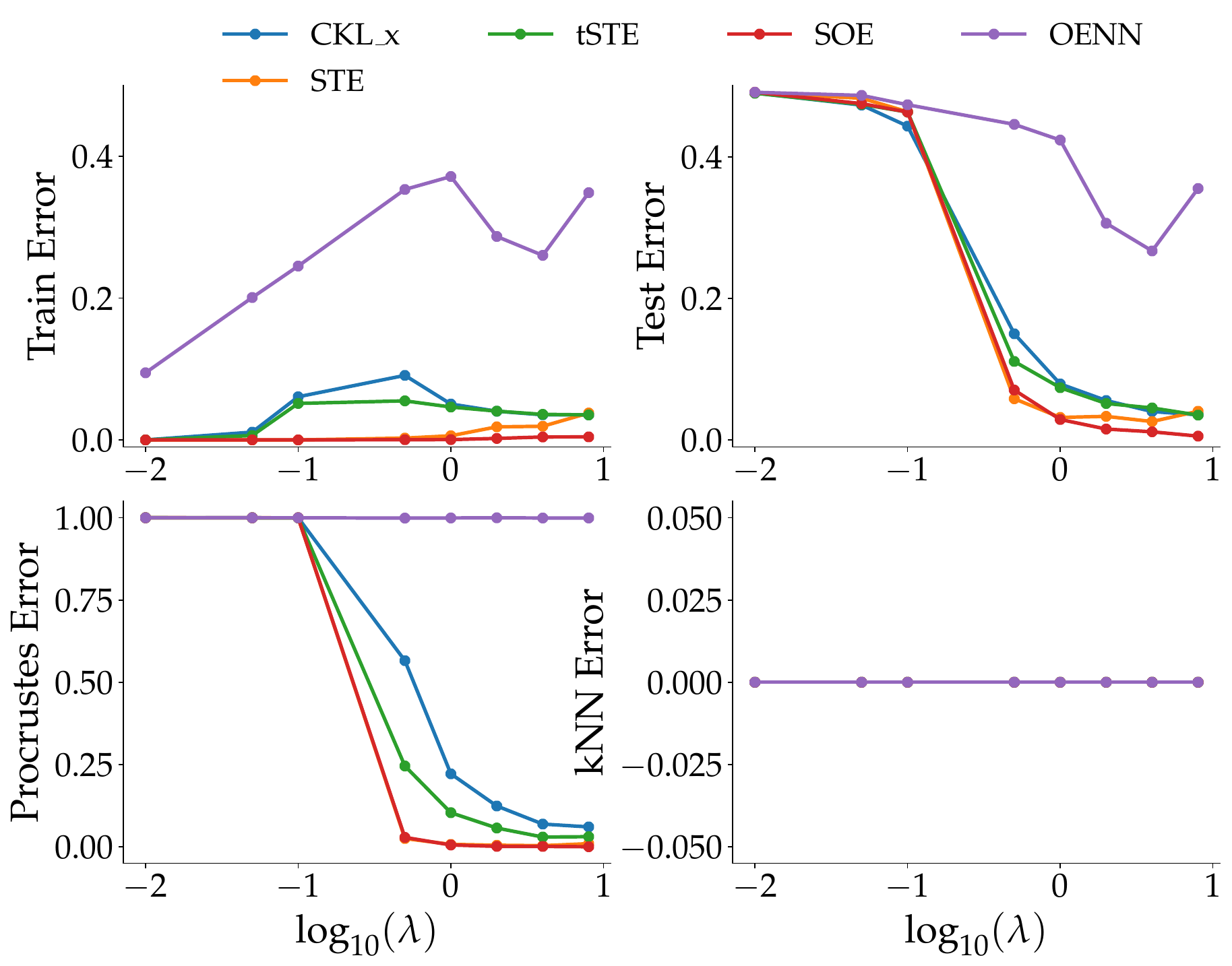}
\caption{{\bfseries Worms: }Increasing number of triplets. }
\end{figure*}

\begin{figure*}[!htb]
\centering
\subcaptionbox*{}[.4\textwidth]{
	\centering
	\includegraphics[width=.4\textwidth]{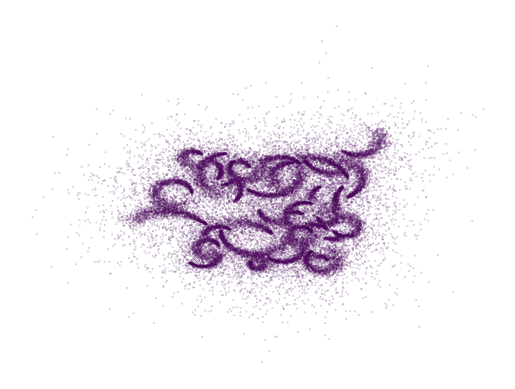}
}
\caption{Worms Original. }
\end{figure*}

\begin{figure*}[!htb]
\centering
{\bfseries Worms}\par\medskip
\includegraphics[width=.99\textwidth]{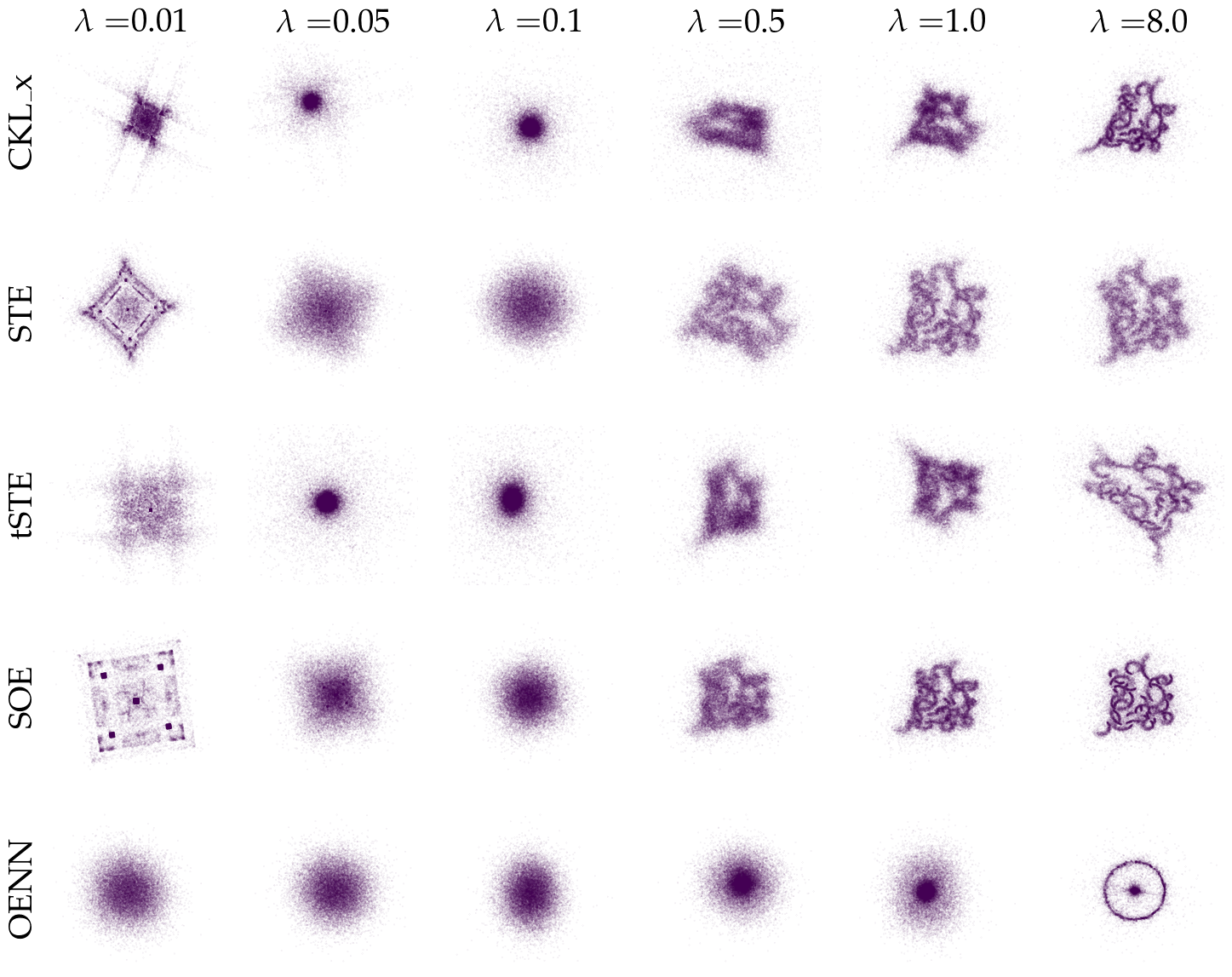}
\caption{\textbf{Worms: }Increasing the number of triplets with a triplet multiplier $\lambda$ yieding $\lambda d n \log n$ triplets. }
\end{figure*}

\begin{figure*}[!htb]
\centering
\includegraphics[width=.8\textwidth]{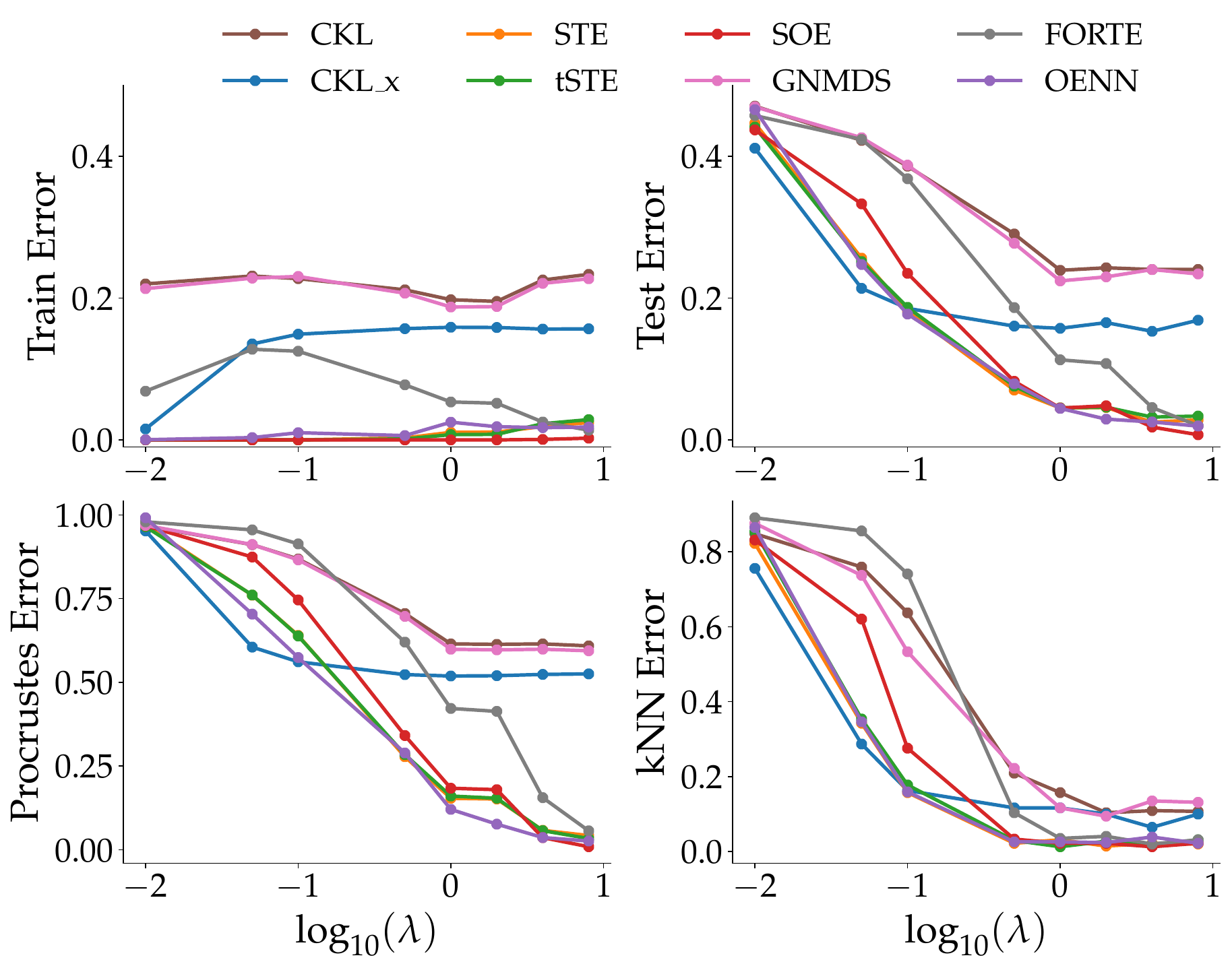}
\caption{{\bfseries Char: }Increasing number of triplets. }
\end{figure*}
%
%
%



\begin{figure*}[!htb]
\centering
\includegraphics[width=.8\textwidth]{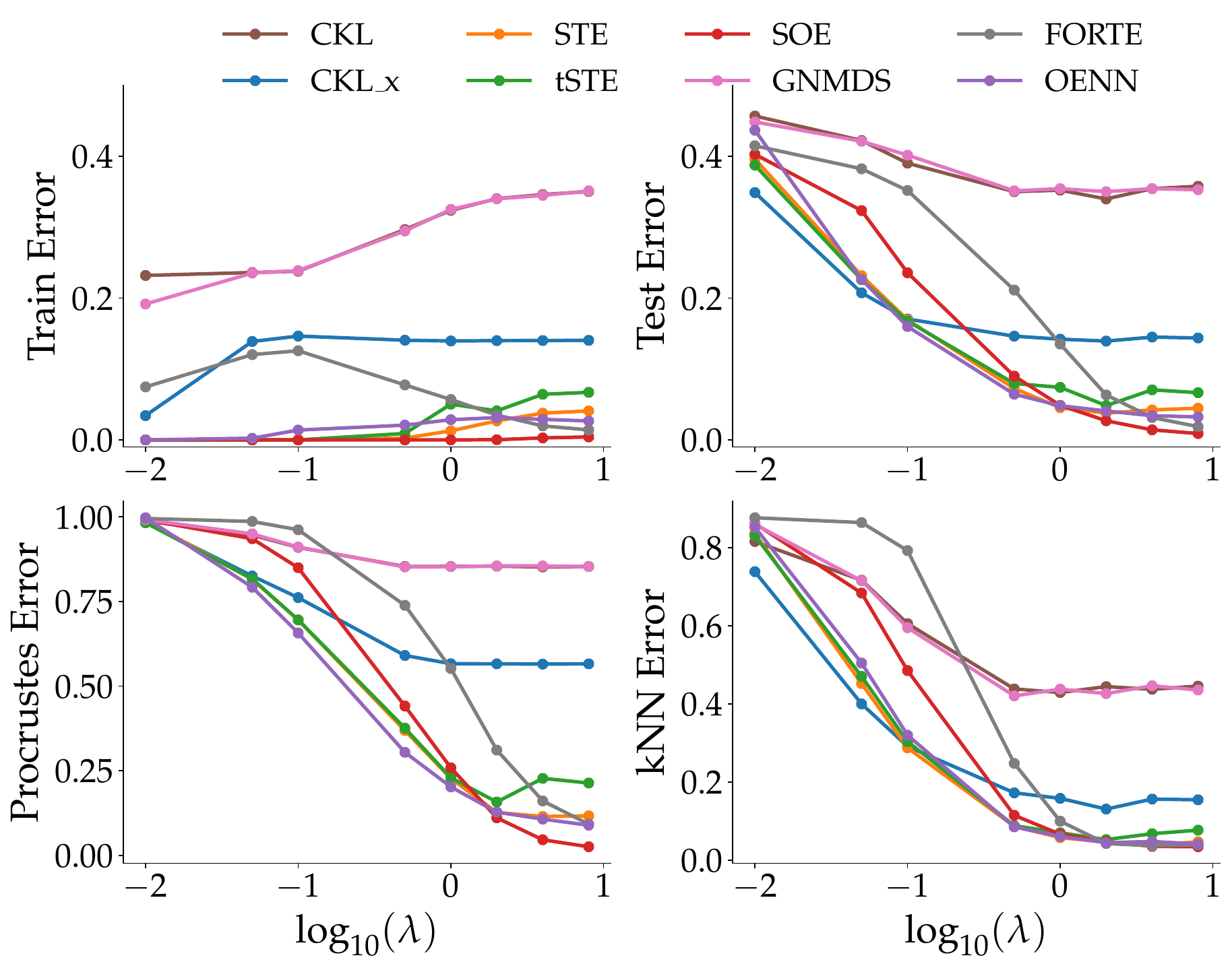}
\caption{{\bfseries usps: }Increasing number of triplets. The value of triplet multiplier $\lambda$ is mentioned above each column.}
\end{figure*}

%
%

\begin{figure*}[!htb]
\centering
\includegraphics[width=.8\textwidth]{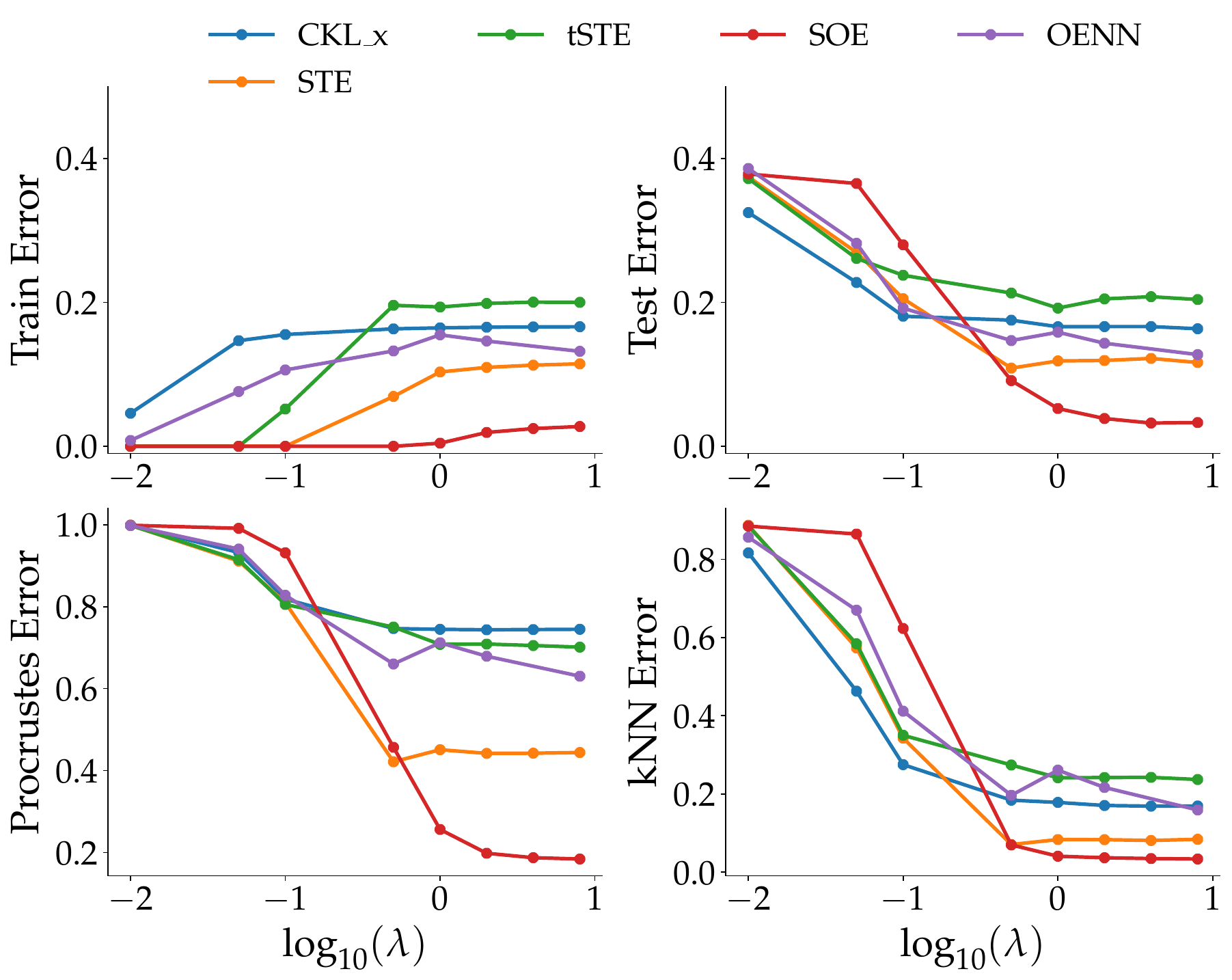}
\caption{{\bfseries MNIST: }Increasing number of triplets. The value of triplet multiplier $\lambda$ is mentioned above each column.}
\end{figure*}



\newpage
\FloatBarrier
\section{INCREASING NOISE}

\begin{figure*}[!htb]
\centering
\includegraphics[width=.7\textwidth]{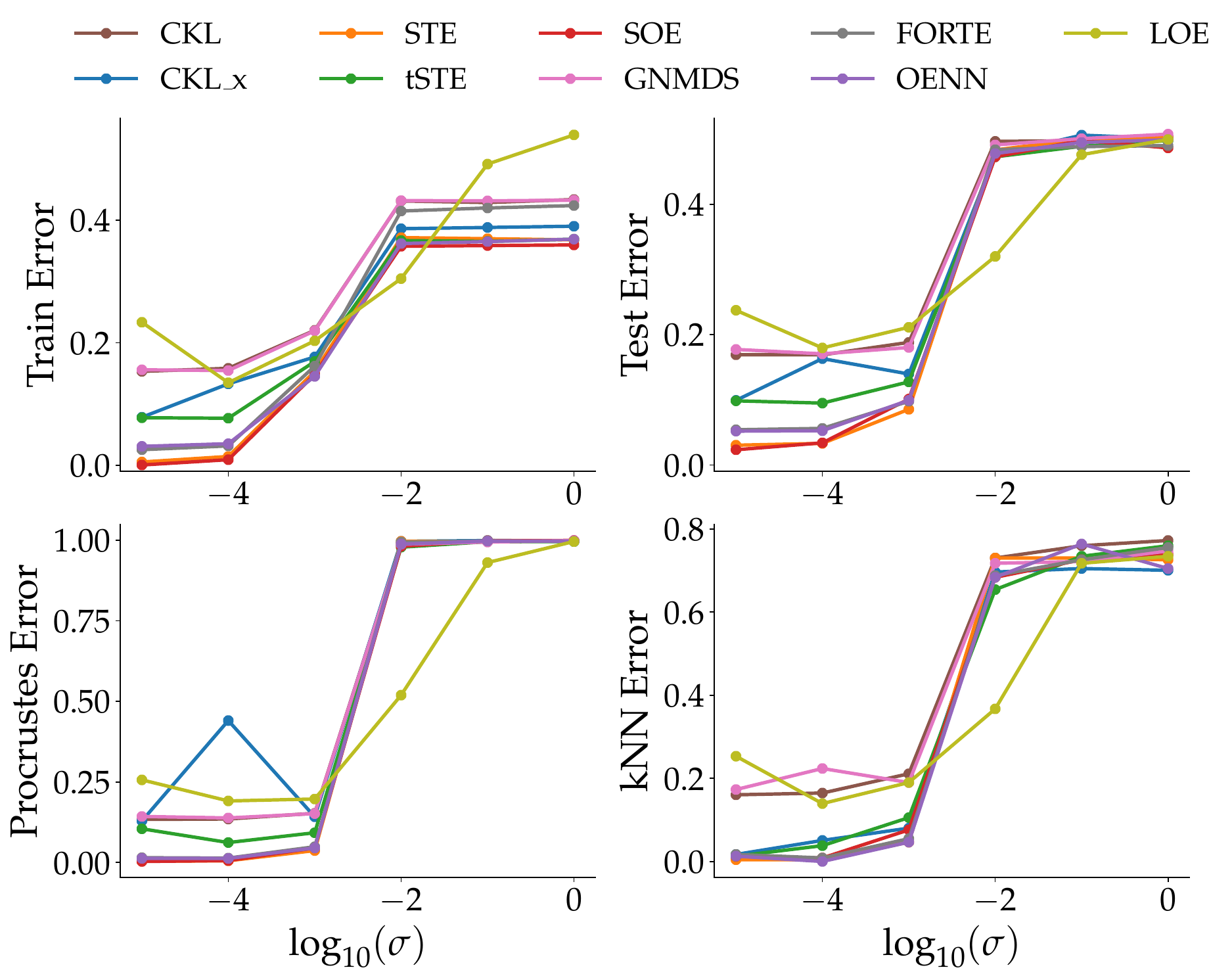}
\caption{{\bfseries Aggregation: }Increasing $\sigma$ of Gaussian Noise. The value of triplet multiplier $\lambda$ is $2$.}
\end{figure*}

\begin{figure*}[!htb]
\centering
\includegraphics[width=.7\textwidth]{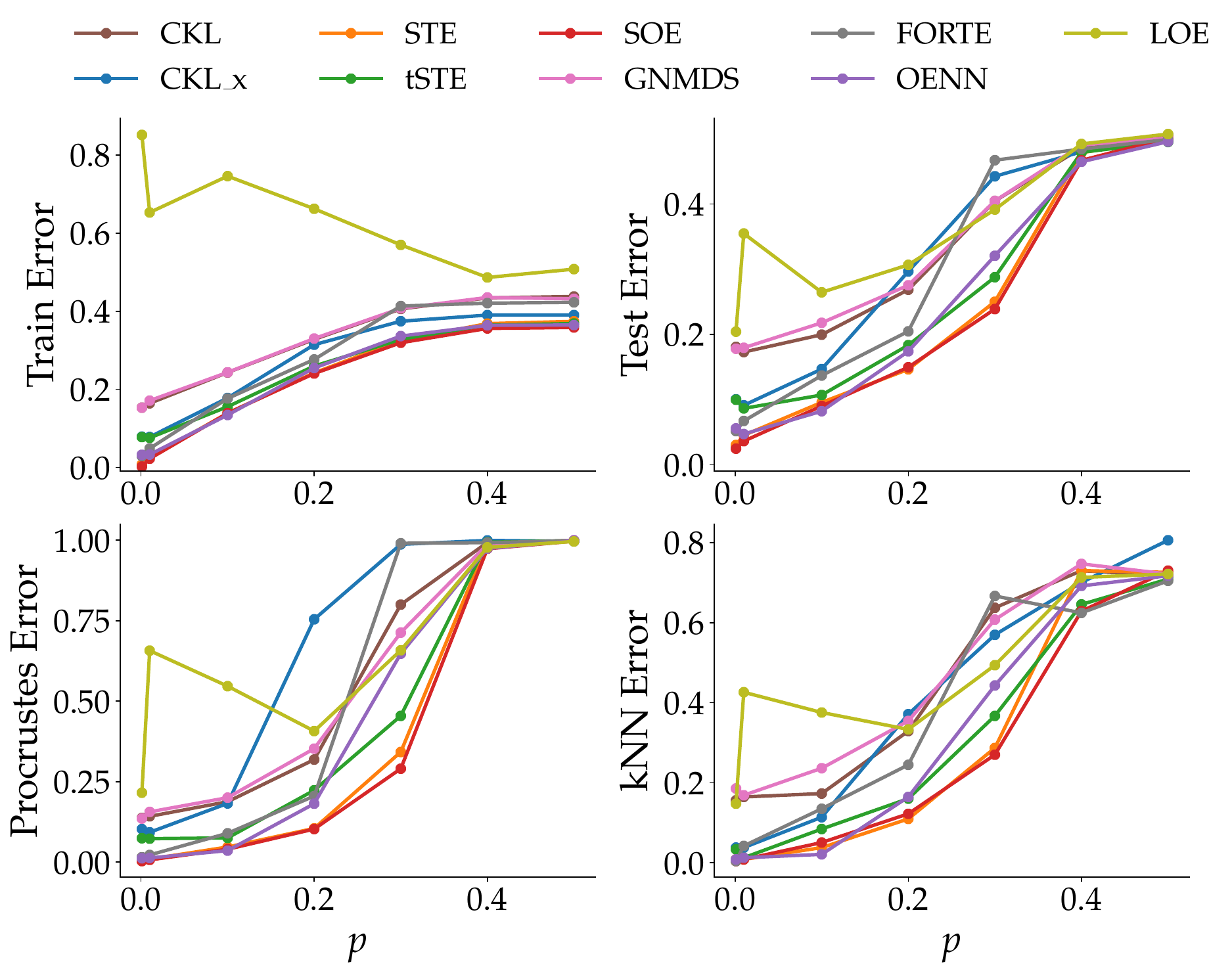}
\caption{{\bfseries Aggregation: }Increasing $p$ of Bernoulli Noise. The value of triplet multiplier $\lambda$ is $2$.}
\end{figure*}

\begin{figure*}[!htb]
\centering
{\bfseries Aggregation}\par\medskip
\subcaptionbox*{}[.4\textwidth]{
	\centering
	\includegraphics[width=.4\textwidth]{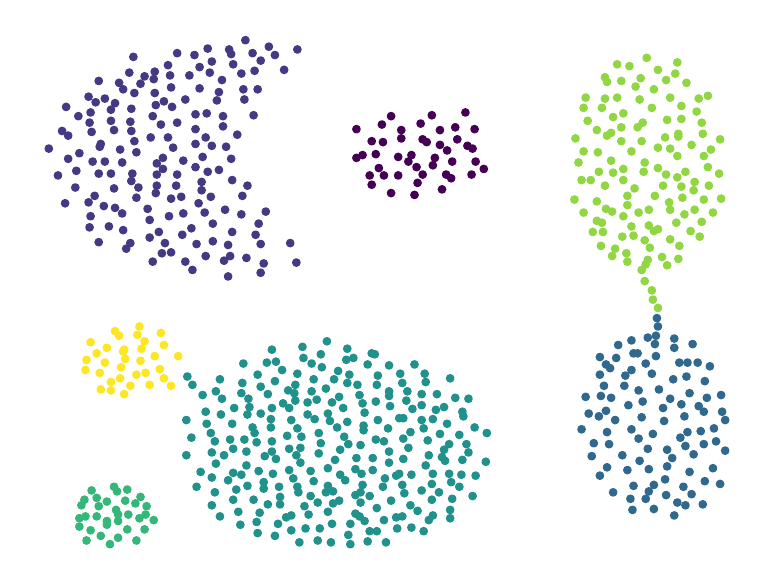}
}
\caption{A visualization of the original Aggregation dataset.}
\end{figure*}

\begin{figure*}[!htb]
\centering
\includegraphics[width=1.0\textwidth]{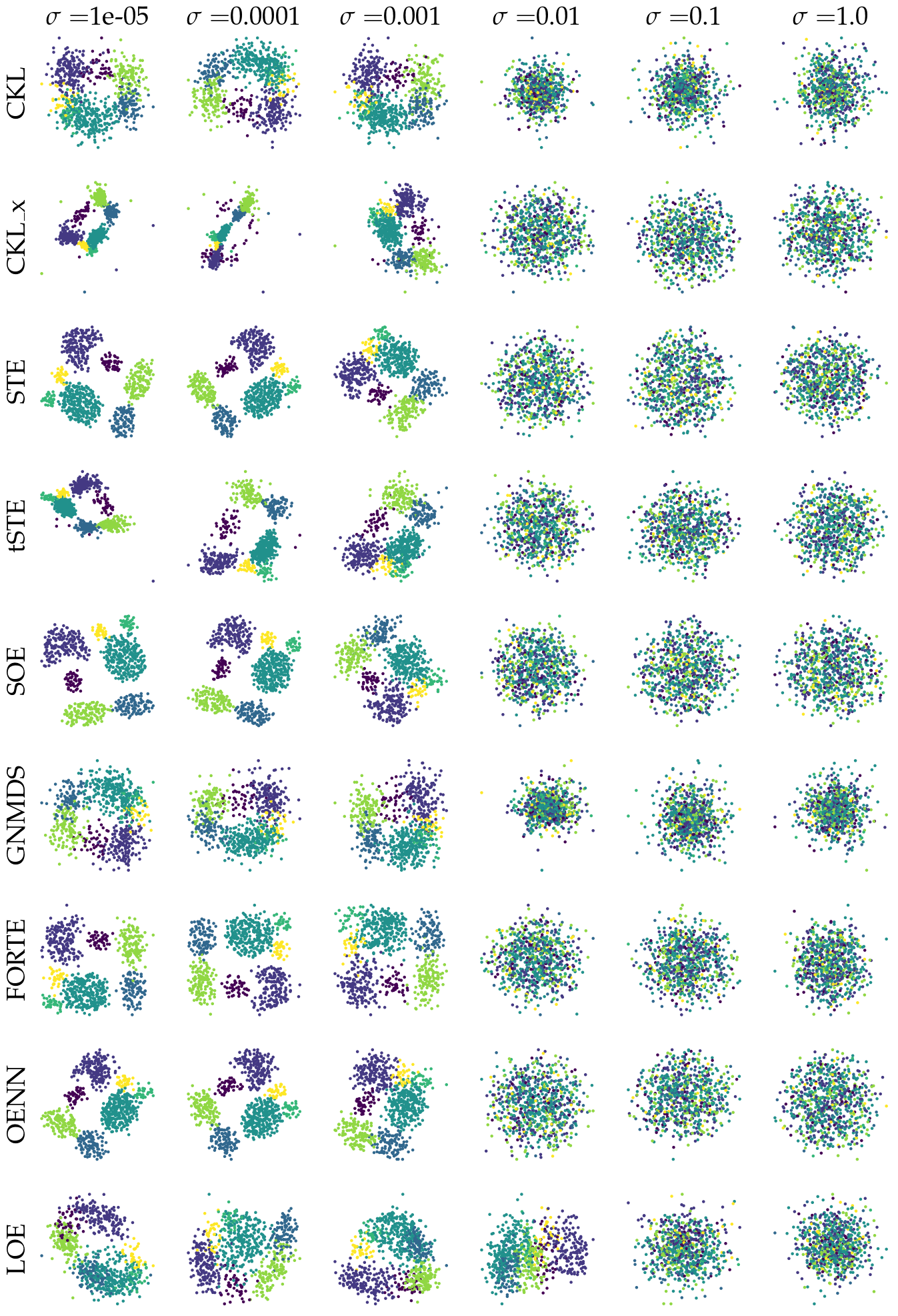}
\caption{{\bfseries Aggregation: }Increasing Gaussian Noise. The value of triplet multiplier $\lambda$ is $2$.}
\end{figure*}

\begin{figure*}[!htb]
\centering
\includegraphics[width=1.0\textwidth]{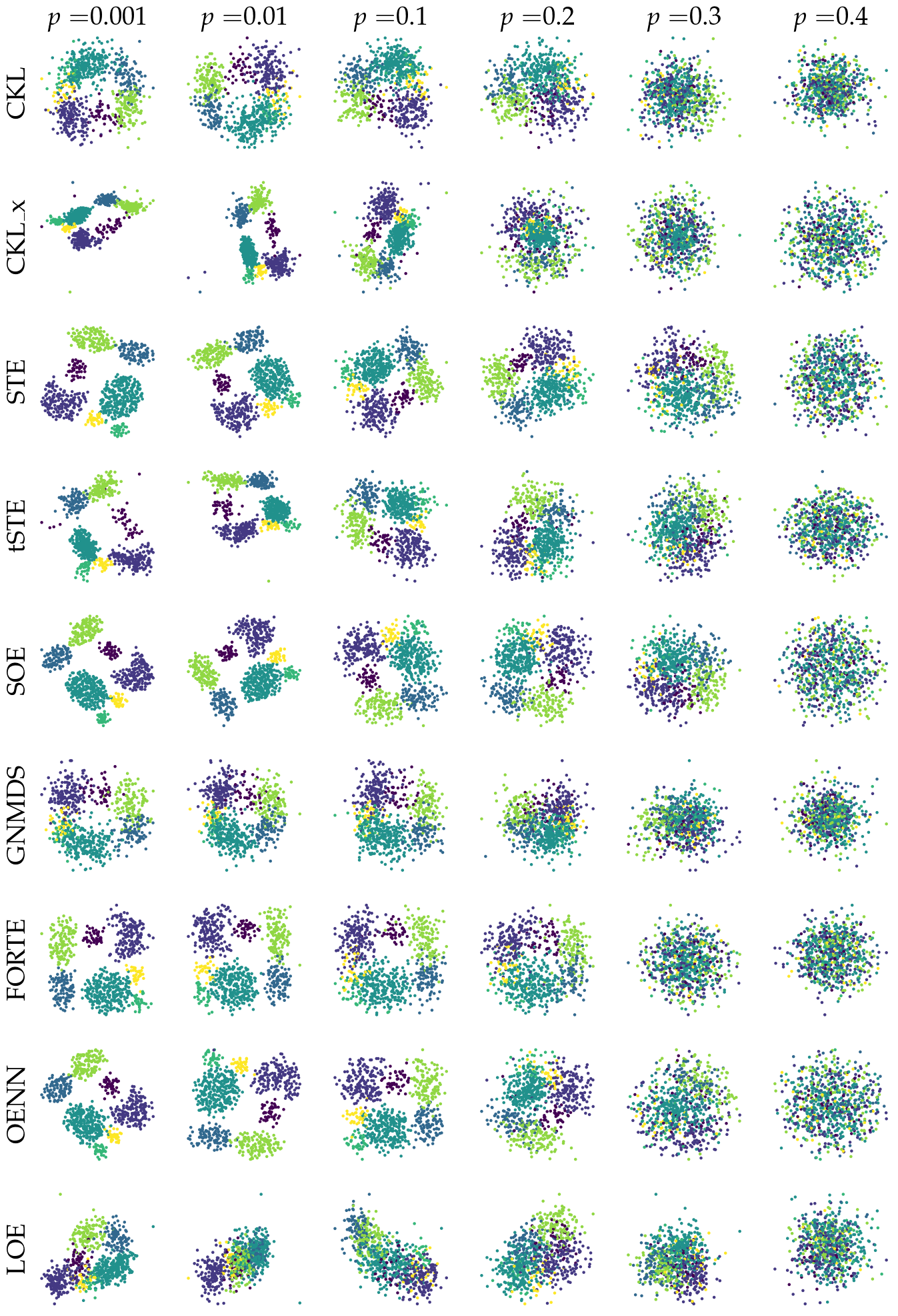}
\caption{{\bfseries Aggregation: }Increasing Bernoulli Noise. The value of triplet multiplier $\lambda$ is $2$.}
\end{figure*}


\begin{figure*}[!htb]
\centering
\includegraphics[width=.7\textwidth]{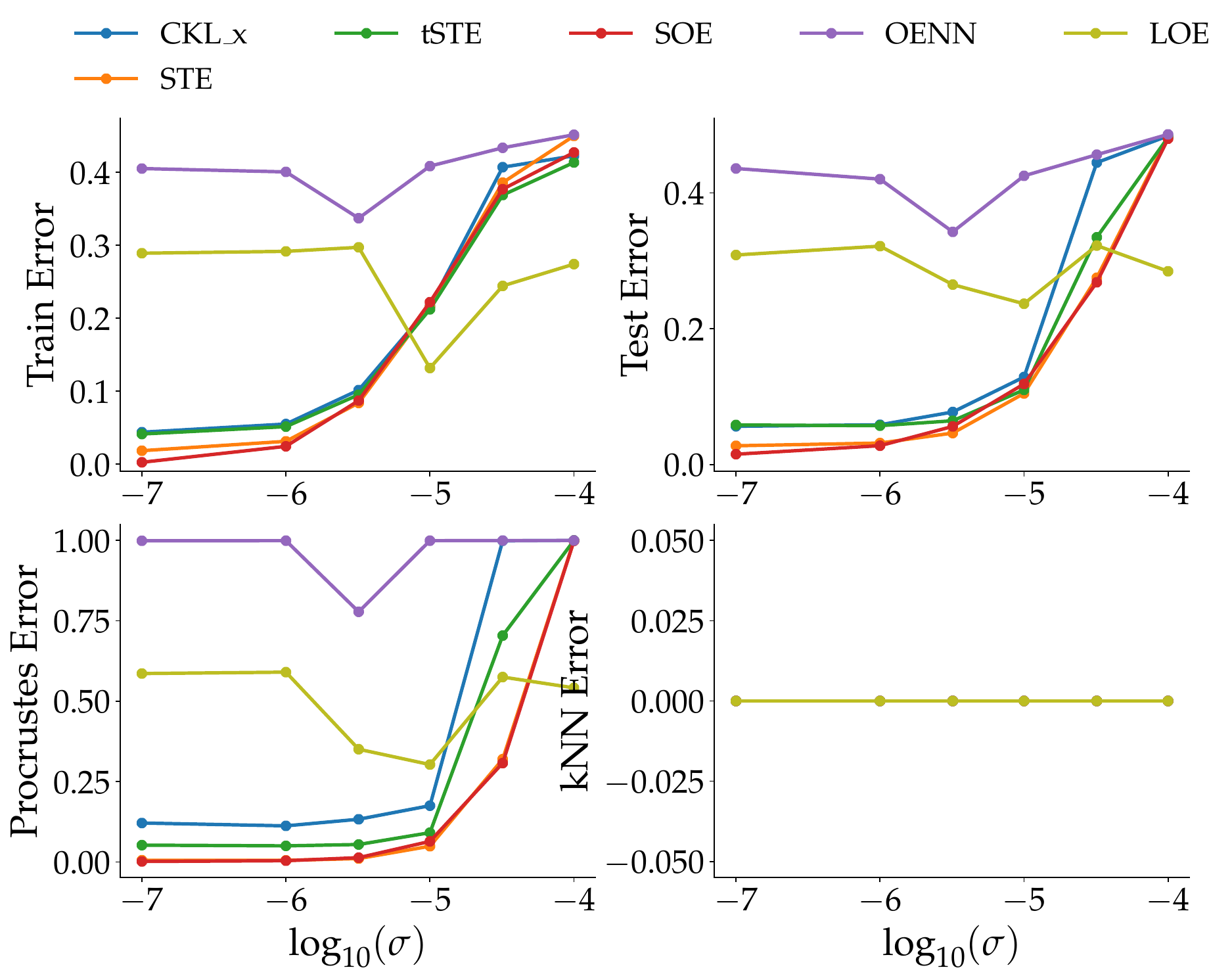}
\caption{{\bfseries Worms: }Increasing $\sigma$ of Gaussian Noise. The value of triplet multiplier $\lambda$ is $2$.}
\end{figure*}

\begin{figure*}[!htb]
\centering
\includegraphics[width=.7\textwidth]{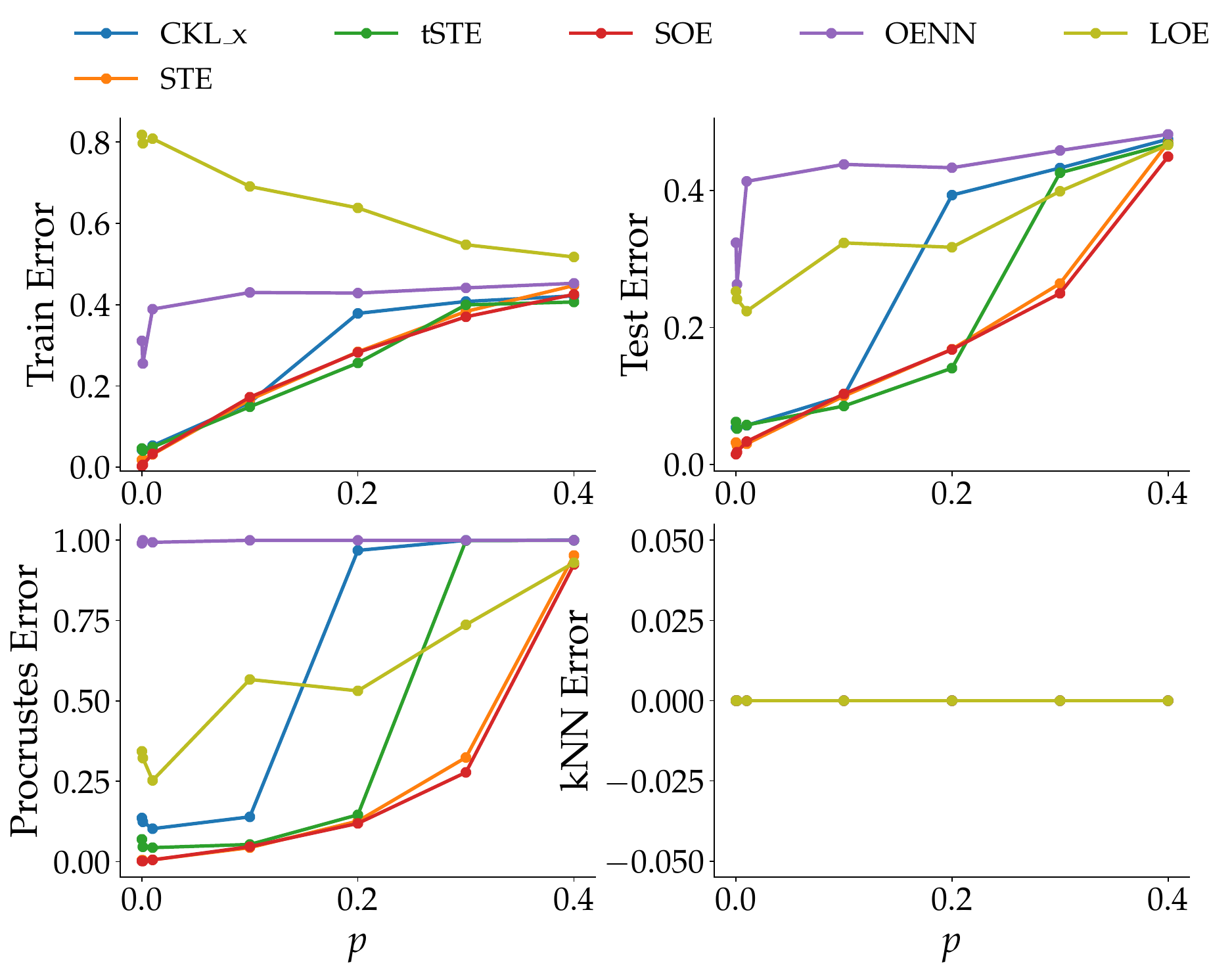}
\caption{{\bfseries Worms: }Increasing $p$ of Bernoulli Noise. The value of triplet multiplier $\lambda$ is $2$.}
\end{figure*}

\begin{figure*}[!htb]
\centering
{\bfseries Worms}\par\medskip
\subcaptionbox*{}[.4\textwidth]{
	\centering
	\includegraphics[width=.4\textwidth]{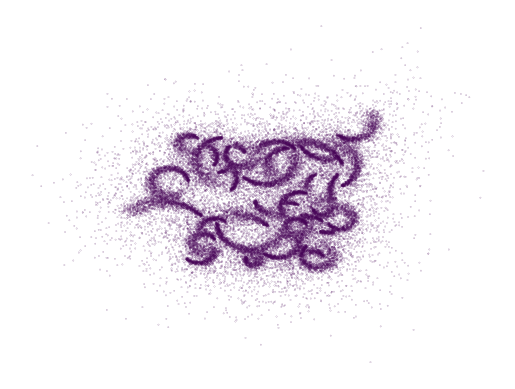}
}
\caption{A visualization of the original Worms dataset.}
\end{figure*}
\begin{figure*}[!htb]
\centering
\includegraphics[width=1.0\textwidth]{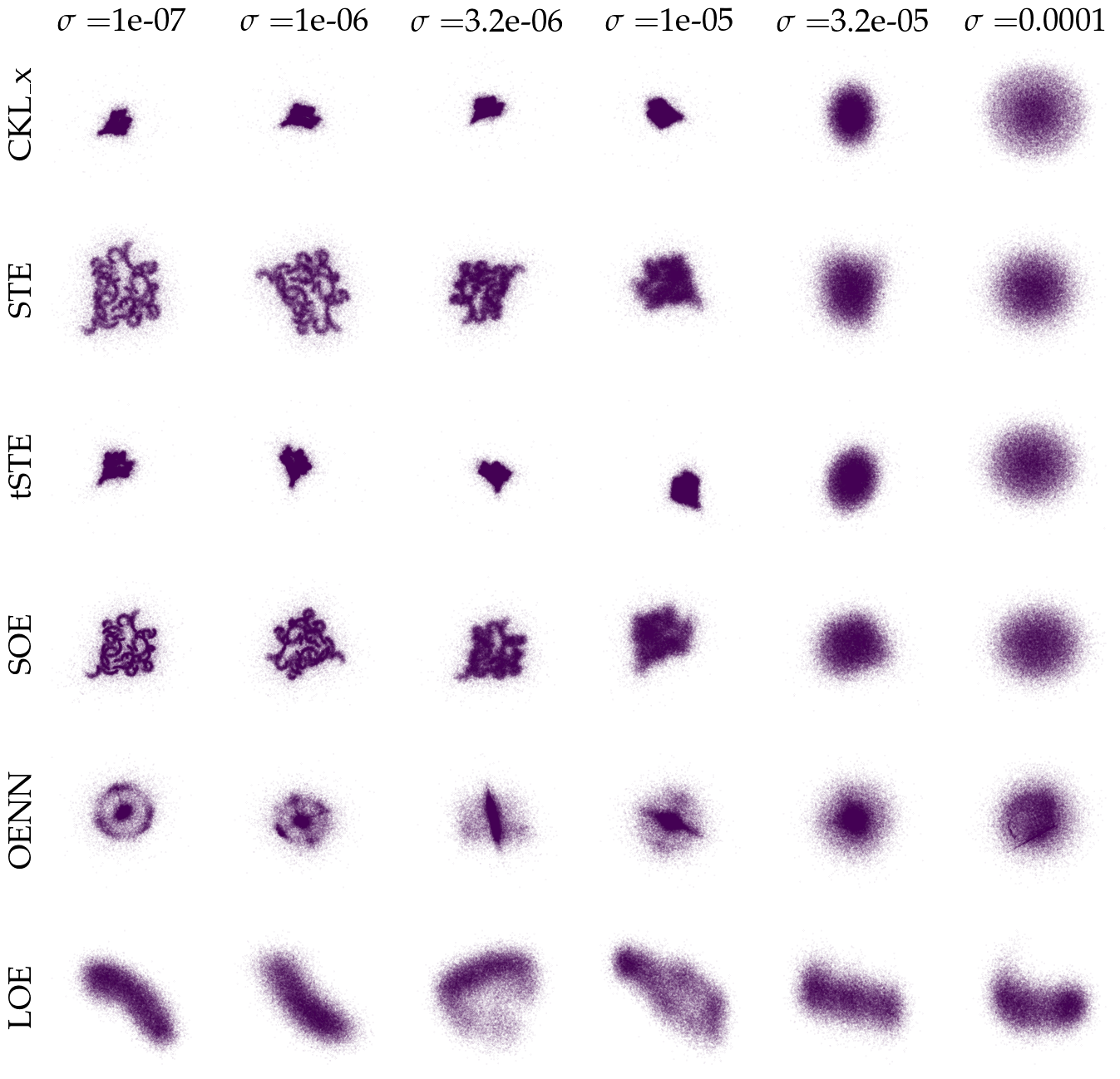}
\caption{{\bfseries Worms: }Increasing Gaussian Noise. The value of triplet multiplier $\lambda$ is $2$.}
\end{figure*}

\begin{figure*}[!htb]
\centering
\includegraphics[width=1.0\textwidth]{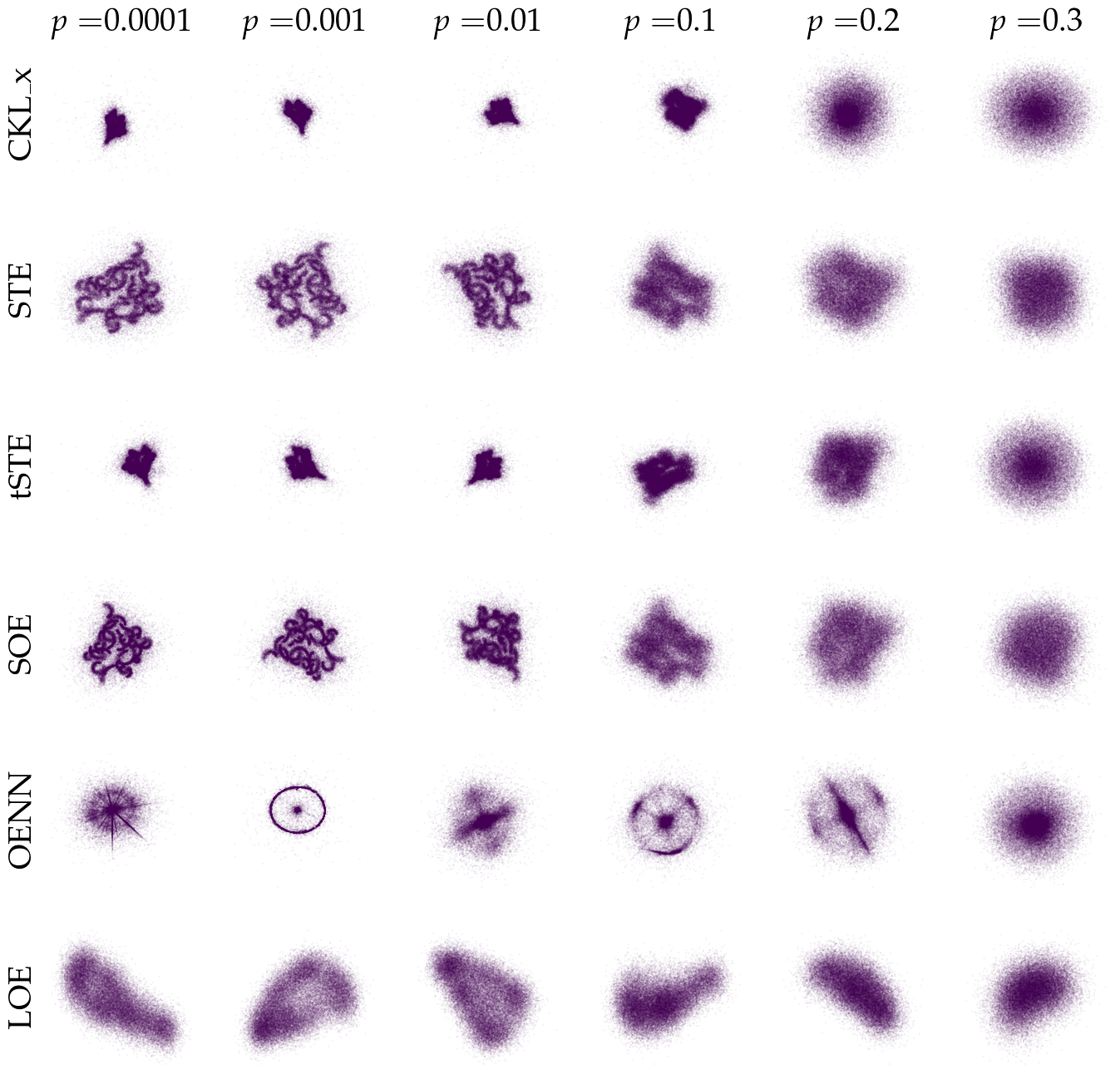}
\caption{{\bfseries Worms: }Increasing Bernoulli Noise. The value of triplet multiplier $\lambda$ is $2$.}
\end{figure*}


\begin{figure*}[!htb]
\centering
\includegraphics[width=.7\textwidth]{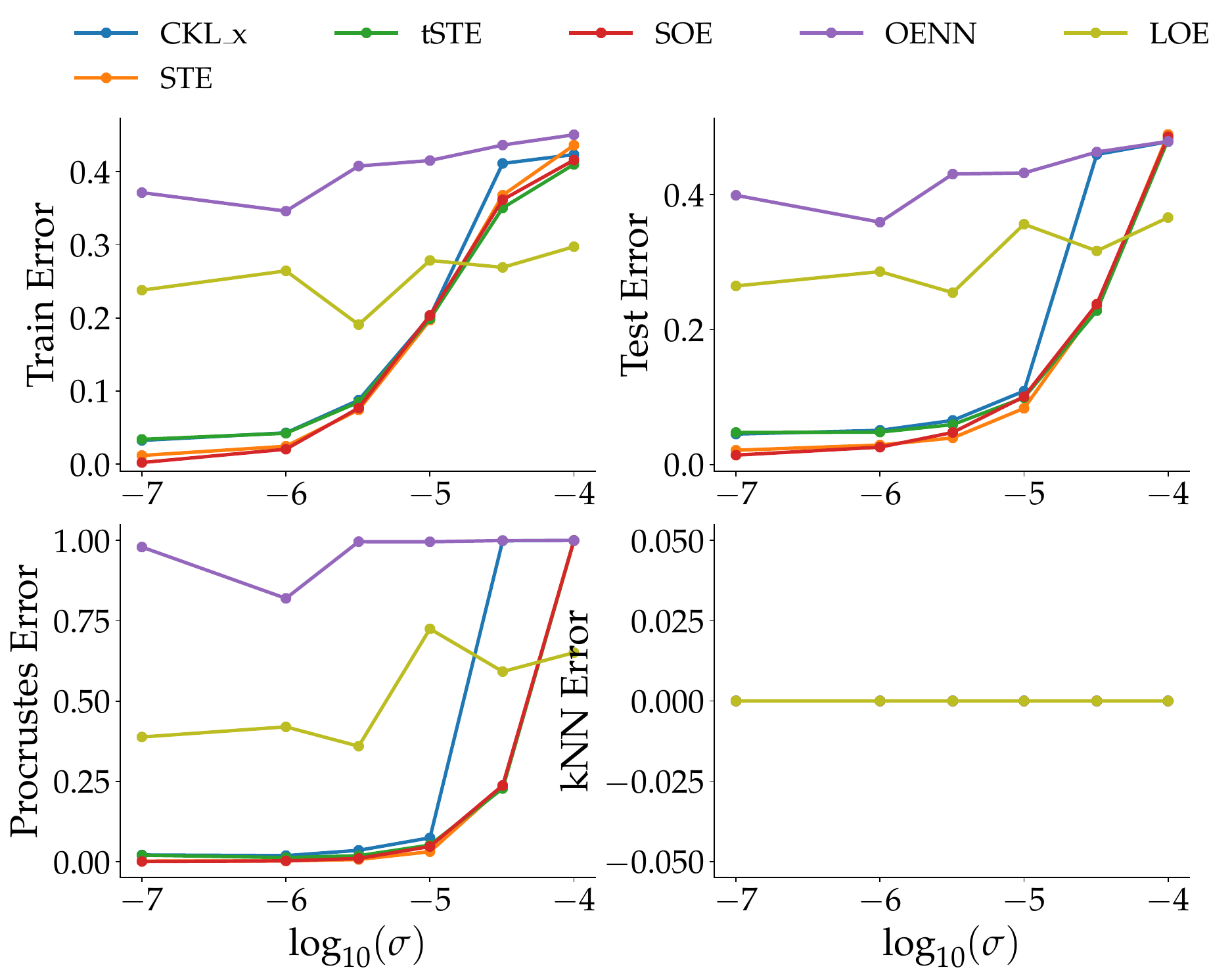}
\caption{{\bfseries Birch1: }Increasing $\sigma$ of Gaussian Noise. The value of triplet multiplier $\lambda$ is $2$.}
\end{figure*}

\begin{figure*}[!htb]
\centering
\includegraphics[width=.7\textwidth]{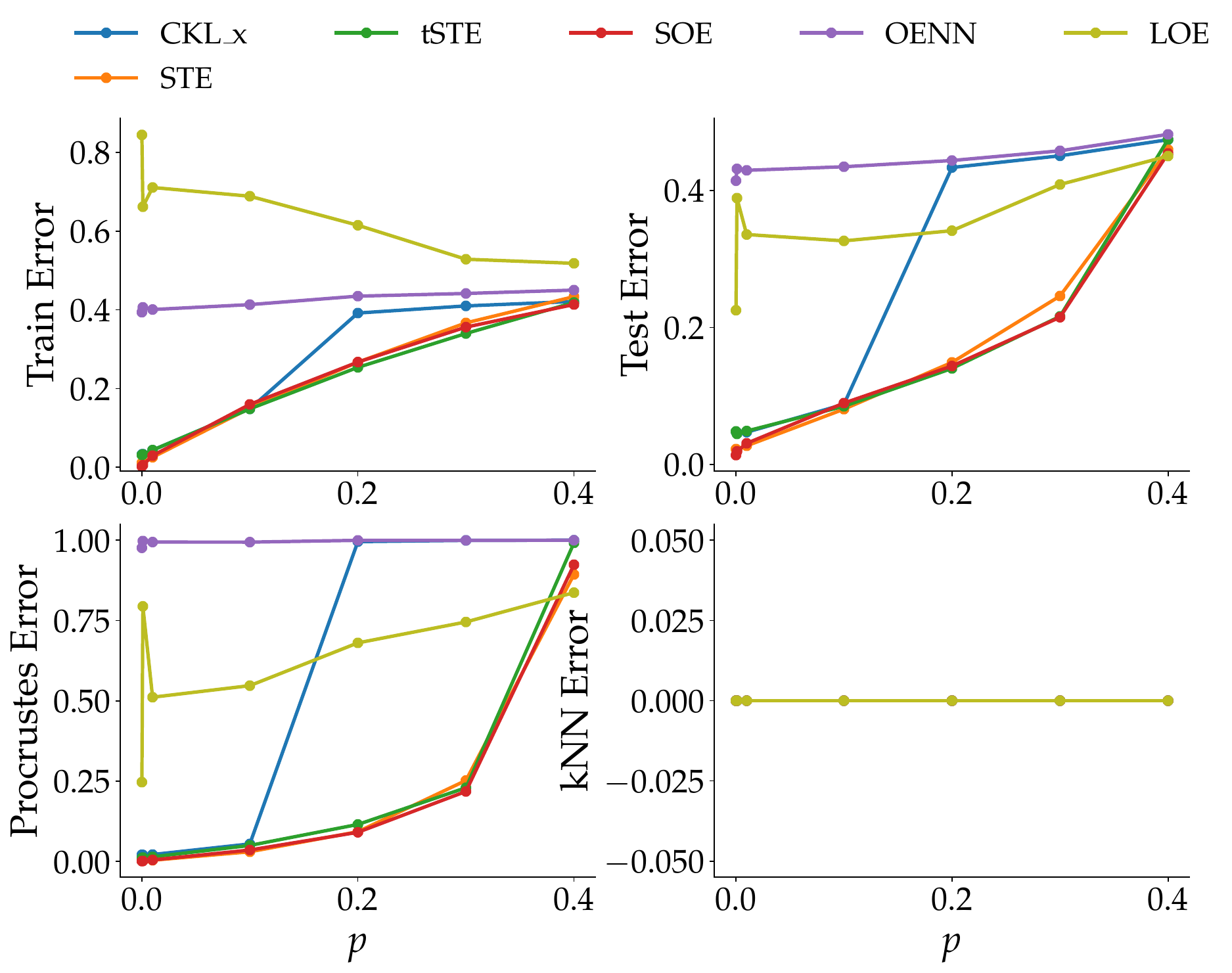}
\caption{{\bfseries Birch1: }Increasing $p$ of Bernoulli Noise. The value of triplet multiplier $\lambda$ is $2$.}
\end{figure*}


\begin{figure*}[!htb]
\centering
\includegraphics[width=.7\textwidth]{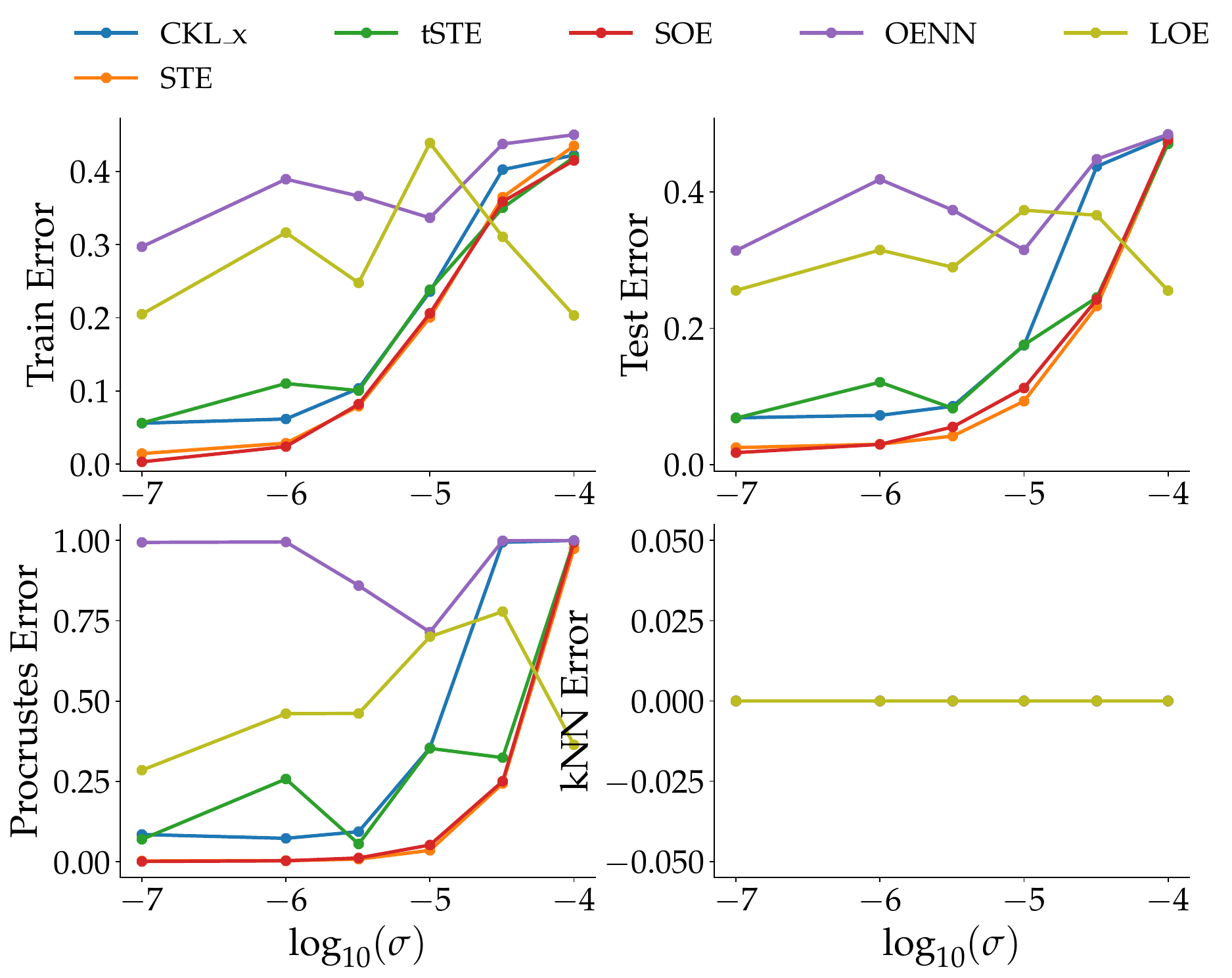}
\caption{{\bfseries Birch3: }Increasing $\sigma$ of Gaussian Noise. The value of triplet multiplier $\lambda$ is $2$.}
\end{figure*}

\begin{figure*}[!htb]
\centering
\includegraphics[width=.7\textwidth]{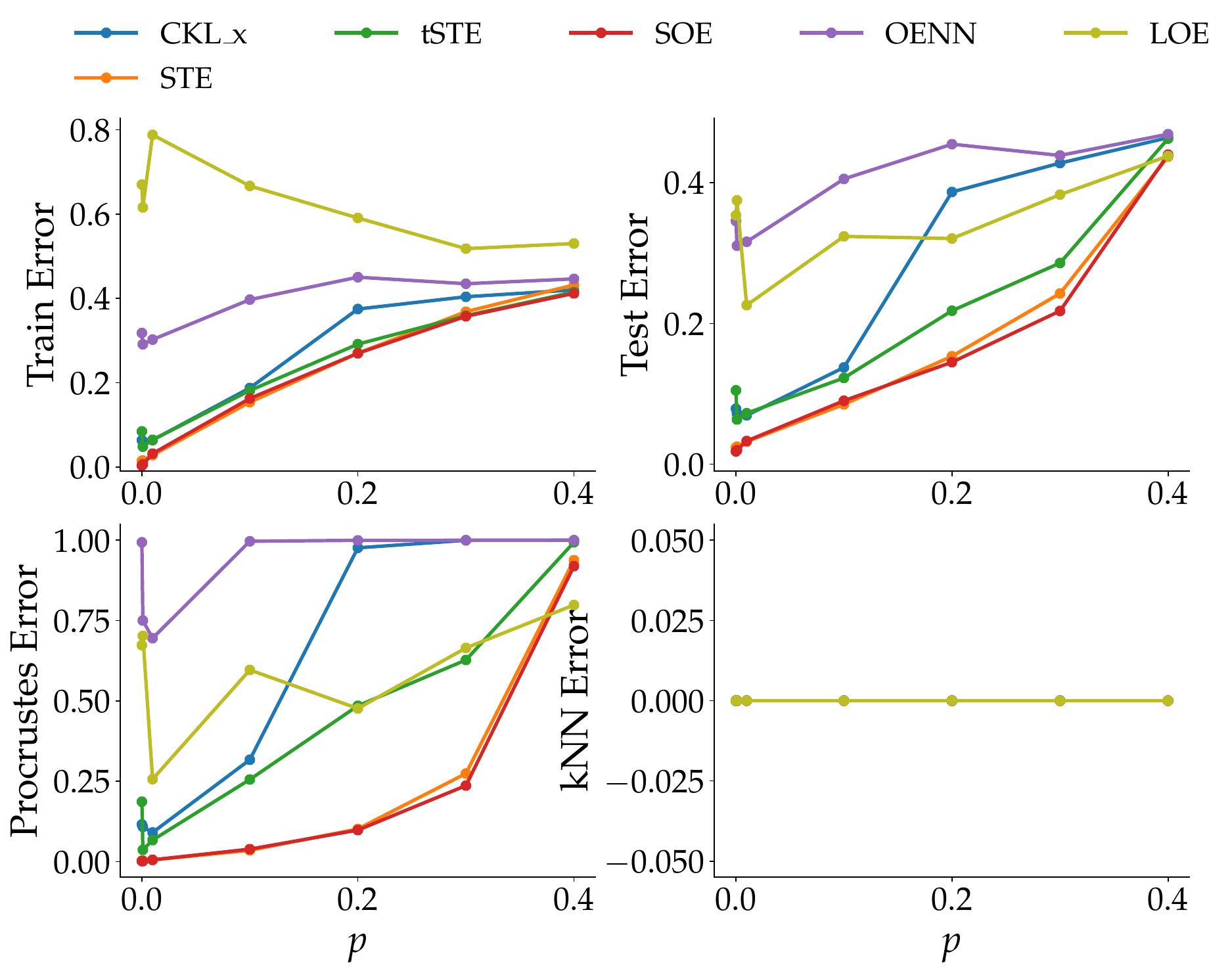}
\caption{{\bfseries Birch3: }Increasing $p$ of Bernoulli Noise. The value of triplet multiplier $\lambda$ is $2$.}
\end{figure*}


\begin{figure*}[!htb]
\centering
\includegraphics[width=.7\textwidth]{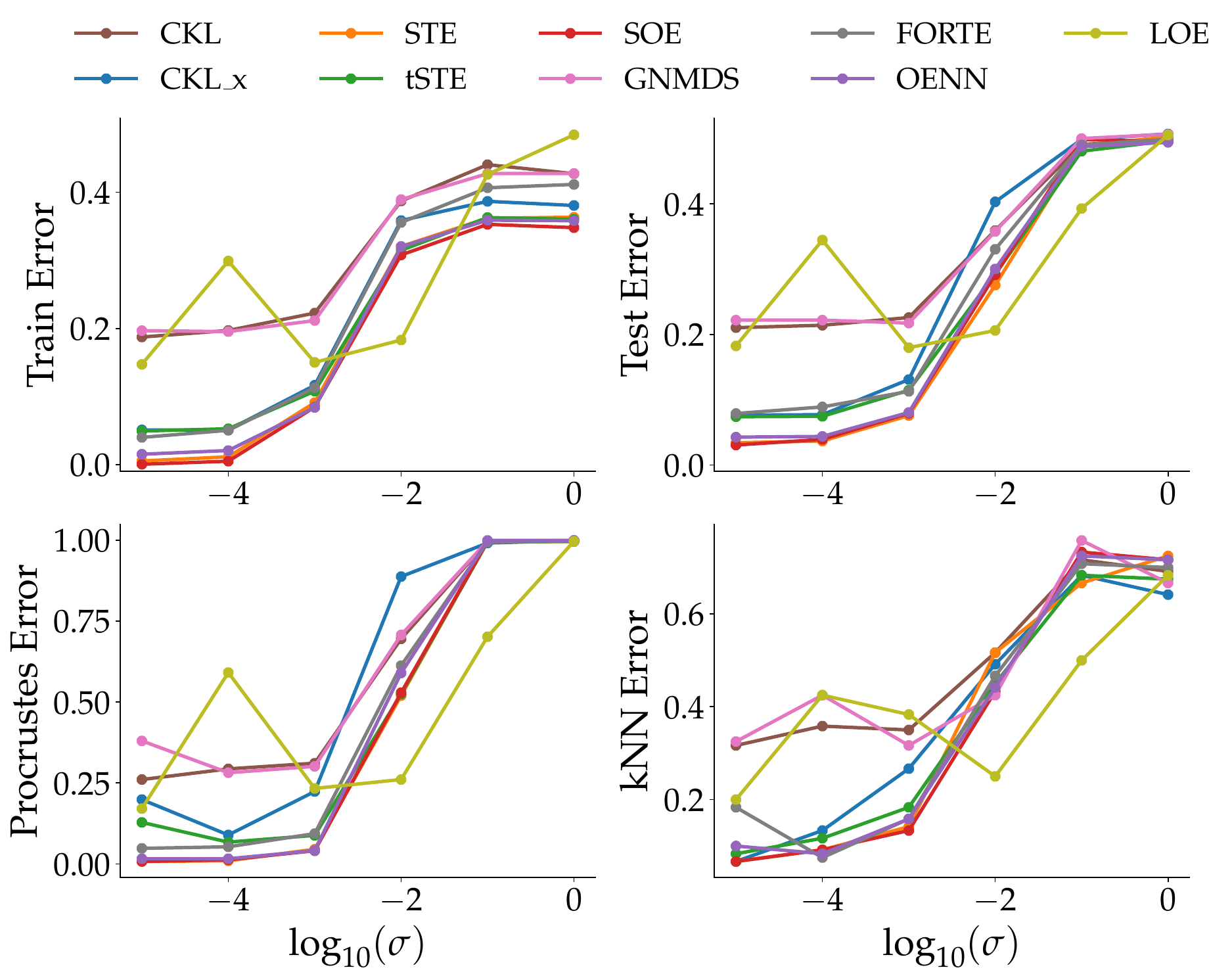}
\caption{{\bfseries Compound: }Increasing $\sigma$ of Gaussian Noise. The value of triplet multiplier $\lambda$ is $2$.}
\end{figure*}


\begin{figure*}[!htb]
\centering
\includegraphics[width=.7\textwidth]{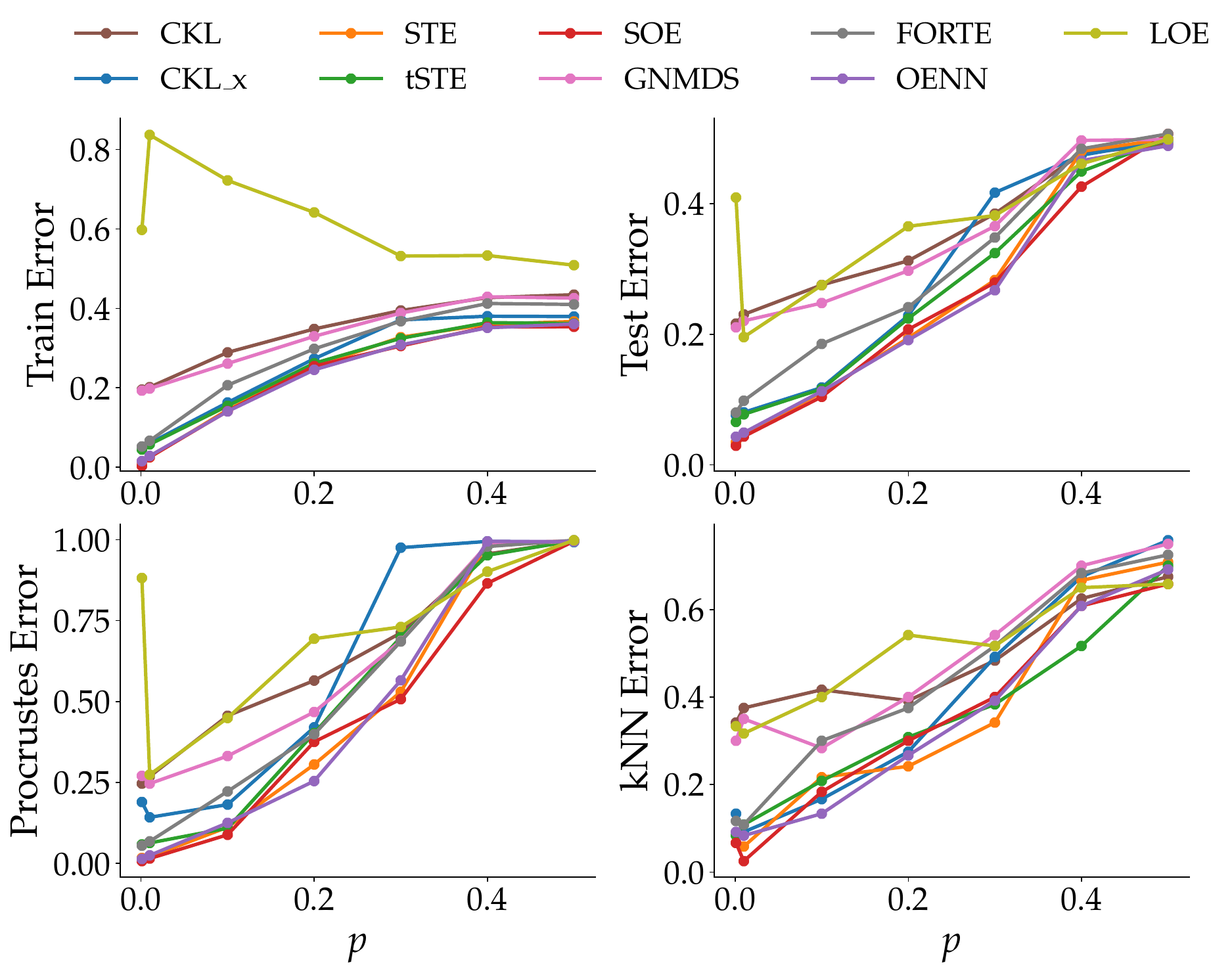}
\caption{{\bfseries Compound: }Increasing $p$ of Bernoulli Noise. The value of triplet multiplier $\lambda$ is $2$.}
\end{figure*}


\begin{figure*}[!htb]
\centering
\includegraphics[width=.7\textwidth]{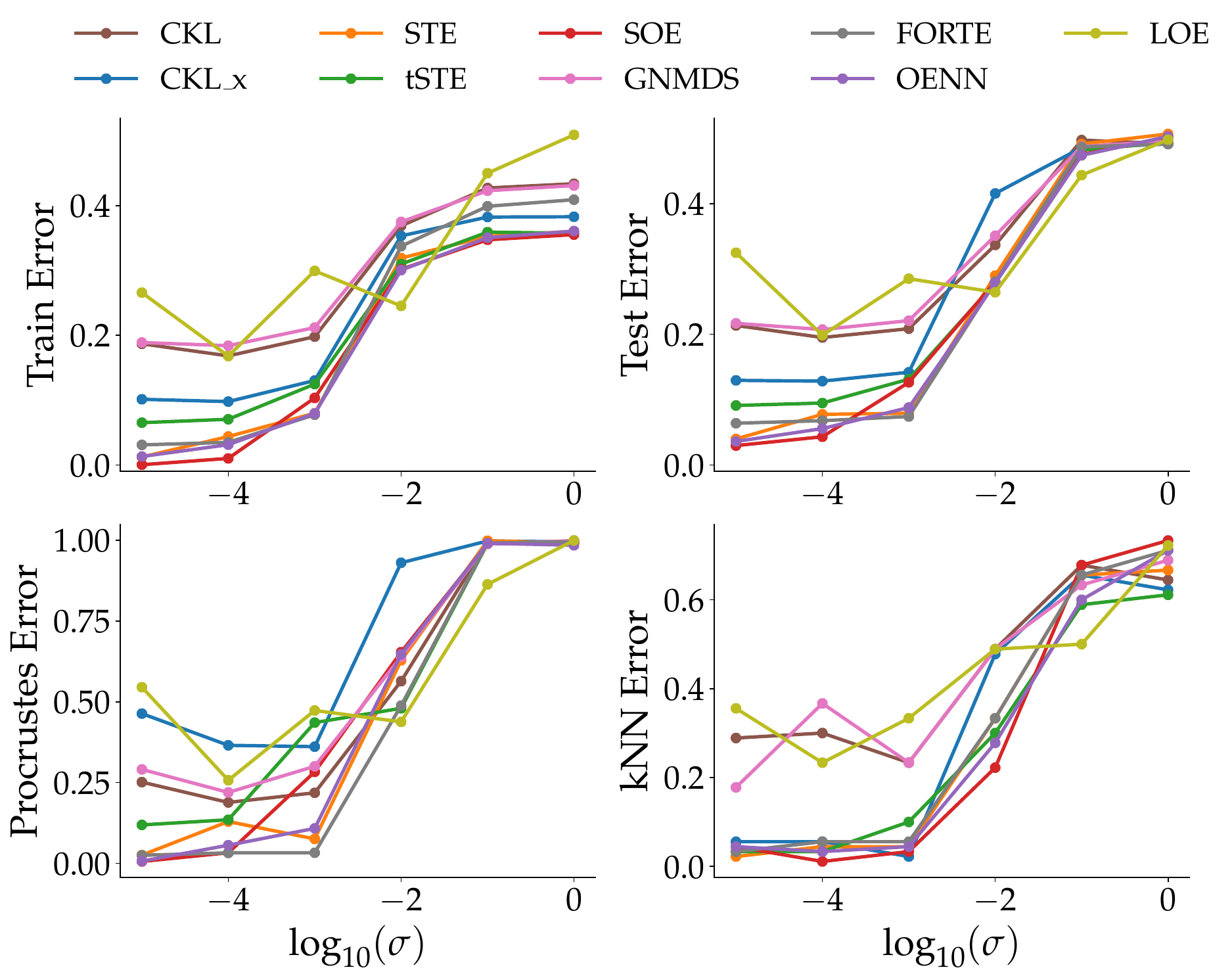}
\caption{{\bfseries Path-based: }Increasing $\sigma$ of Gaussian Noise. The value of triplet multiplier $\lambda$ is $2$.}
\end{figure*}


\begin{figure*}[!htb]
\centering
\includegraphics[width=.7\textwidth]{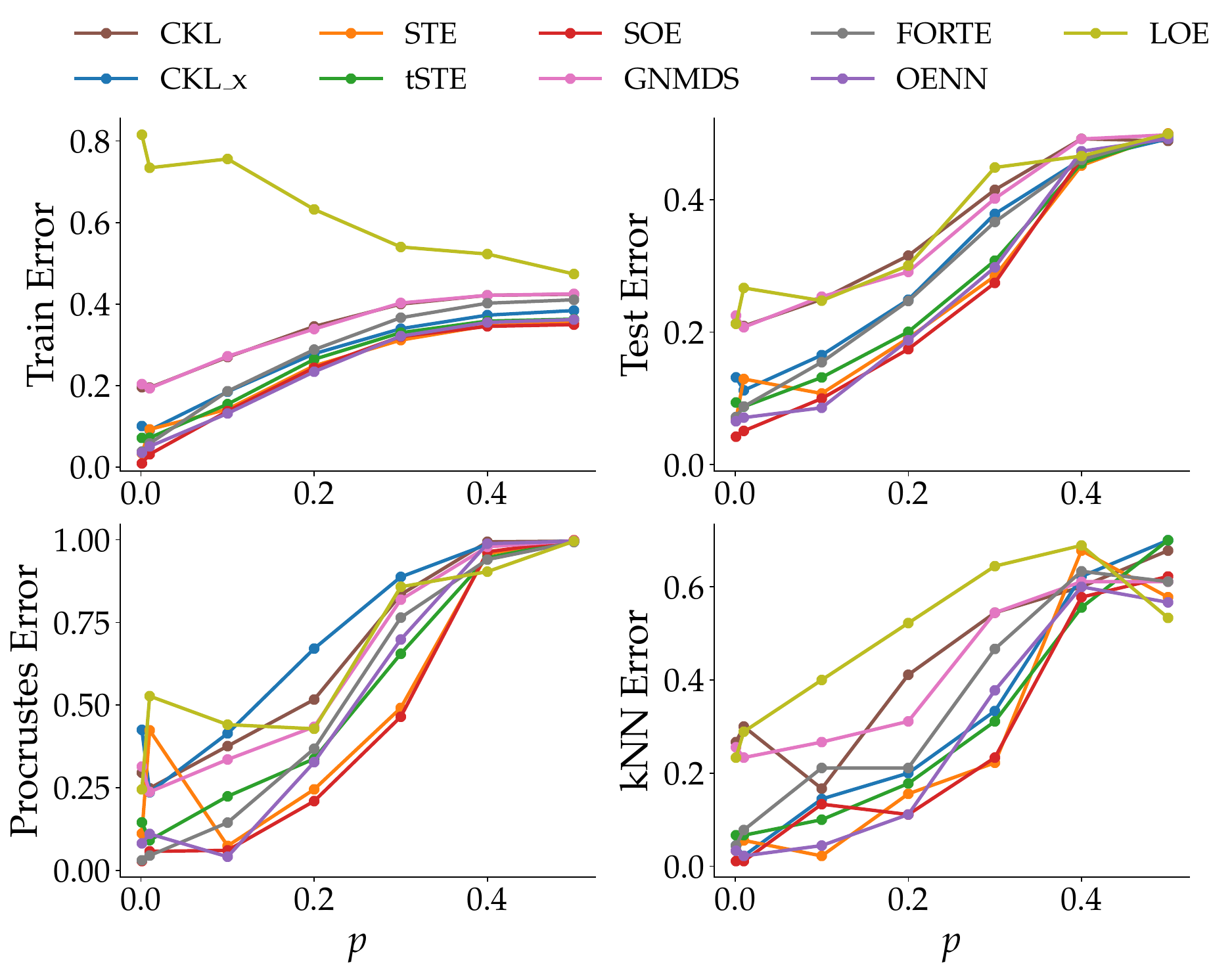}
\caption{{\bfseries Path-based: }Increasing $p$ of Bernoulli Noise. The value of triplet multiplier $\lambda$ is $2$.}
\end{figure*}


\begin{figure*}[!htb]
\centering
\includegraphics[width=.7\textwidth]{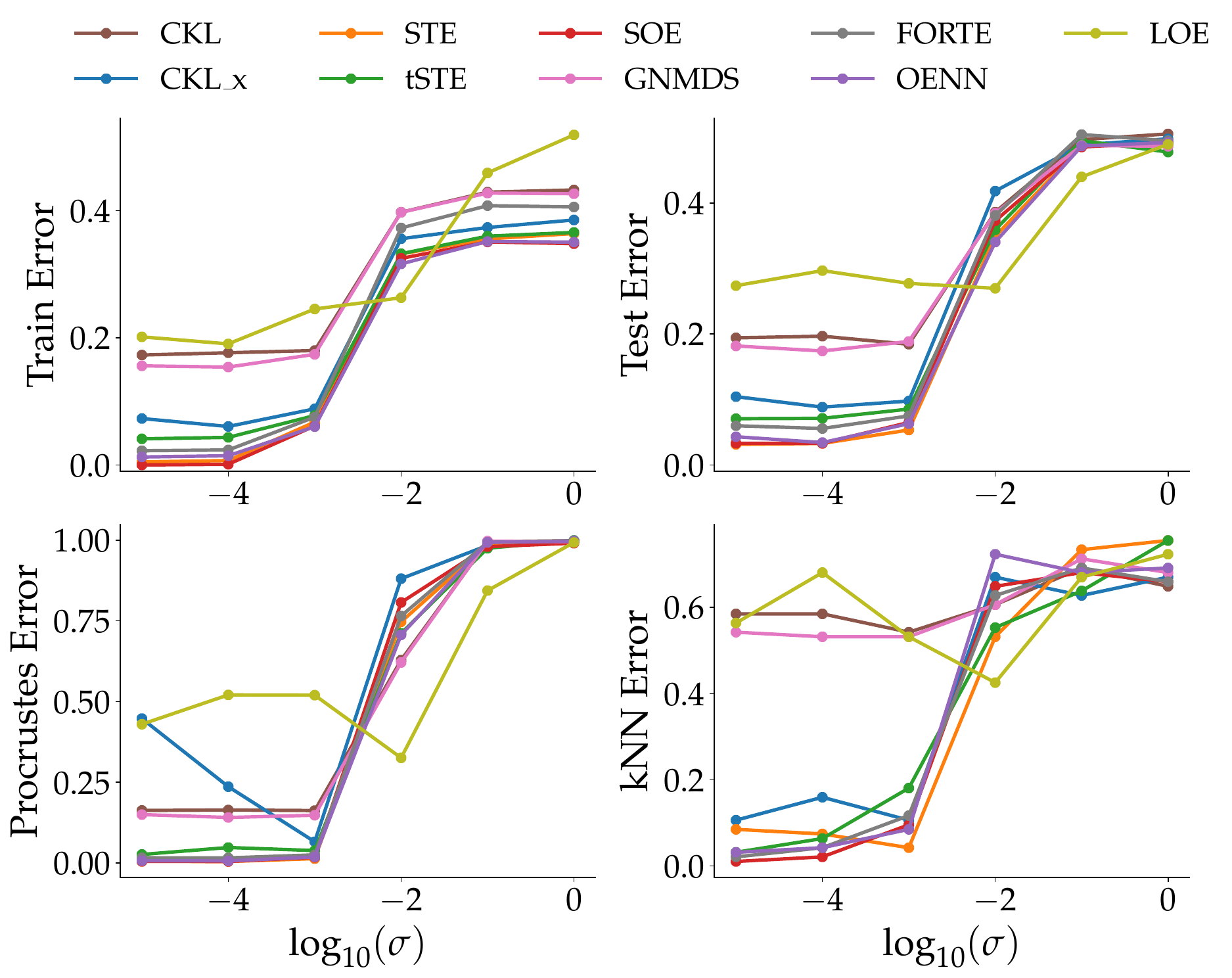}
\caption{{\bfseries Spiral: }Increasing $\sigma$ of Gaussian Noise. The value of triplet multiplier $\lambda$ is $2$.}
\end{figure*}


\begin{figure*}[!htb]
\centering
\includegraphics[width=.7\textwidth]{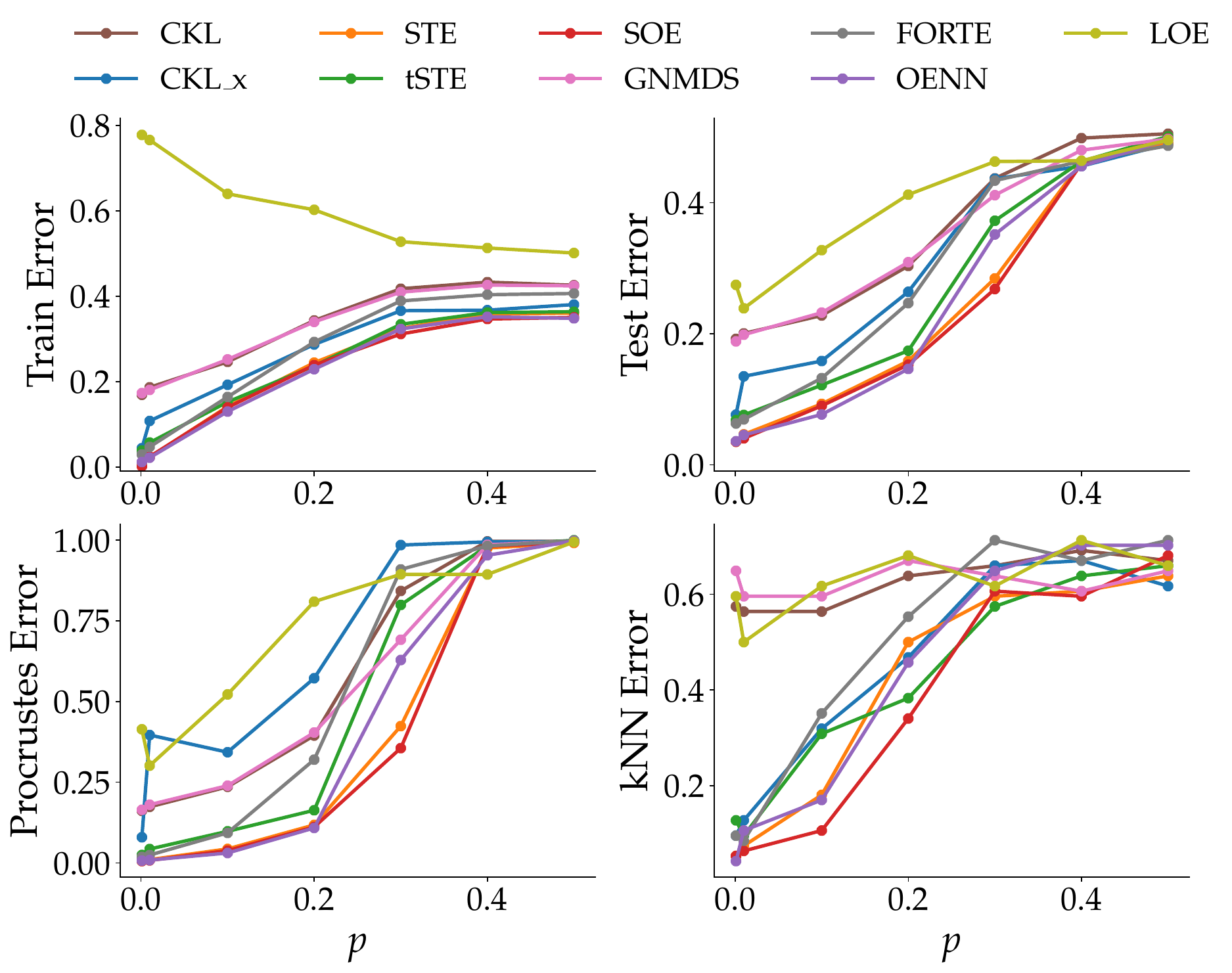}
\caption{{\bfseries Spiral: }Increasing $p$ of Bernoulli Noise. The value of triplet multiplier $\lambda$ is $2$.}
\end{figure*}


\begin{figure*}[!htb]
\centering
\includegraphics[width=.7\textwidth]{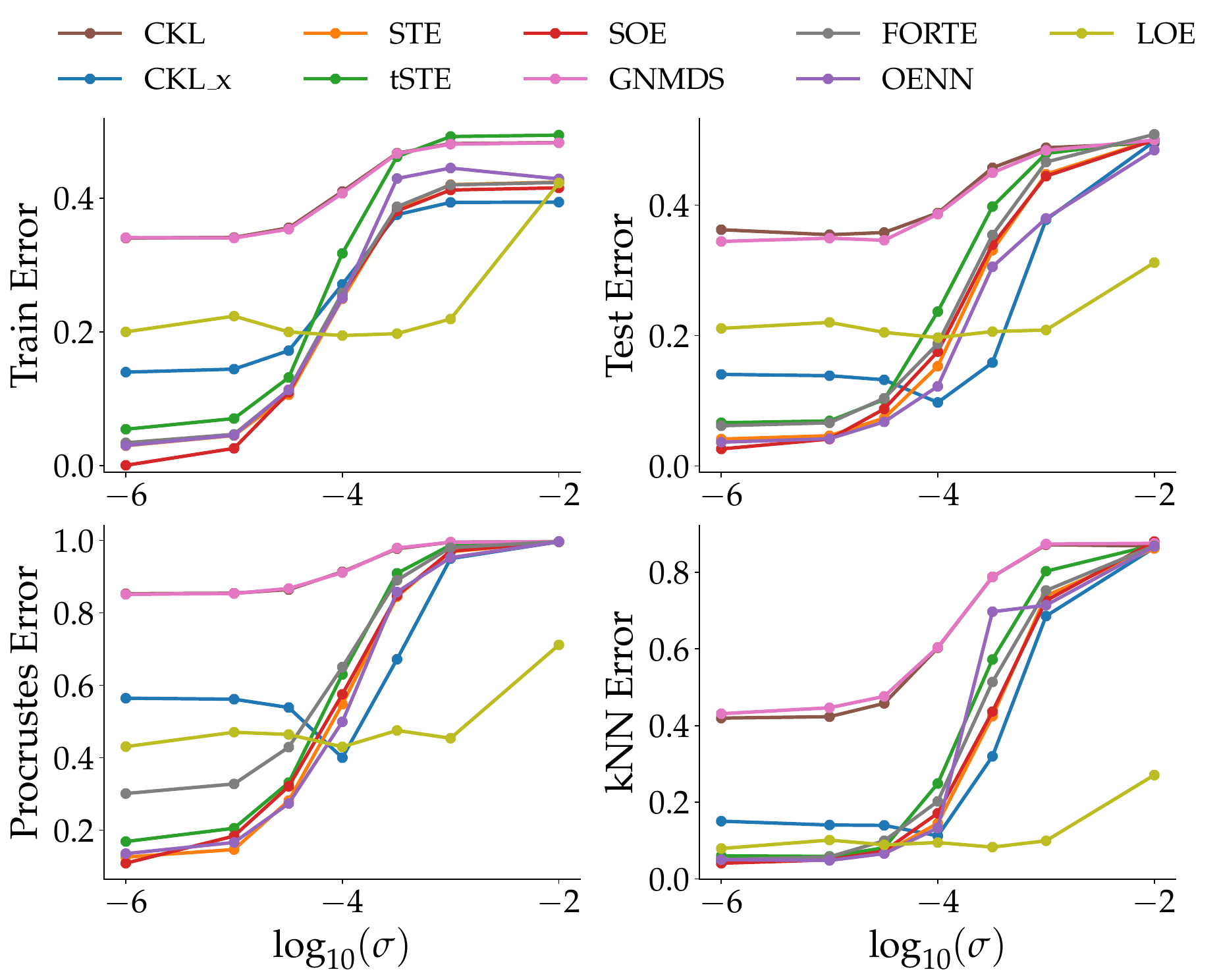}
\caption{{\bfseries USPS: }Increasing $\sigma$ of Gaussian Noise. The value of triplet multiplier $\lambda$ is $2$.}
\end{figure*}


\begin{figure*}[!htb]
\centering
\includegraphics[width=.7\textwidth]{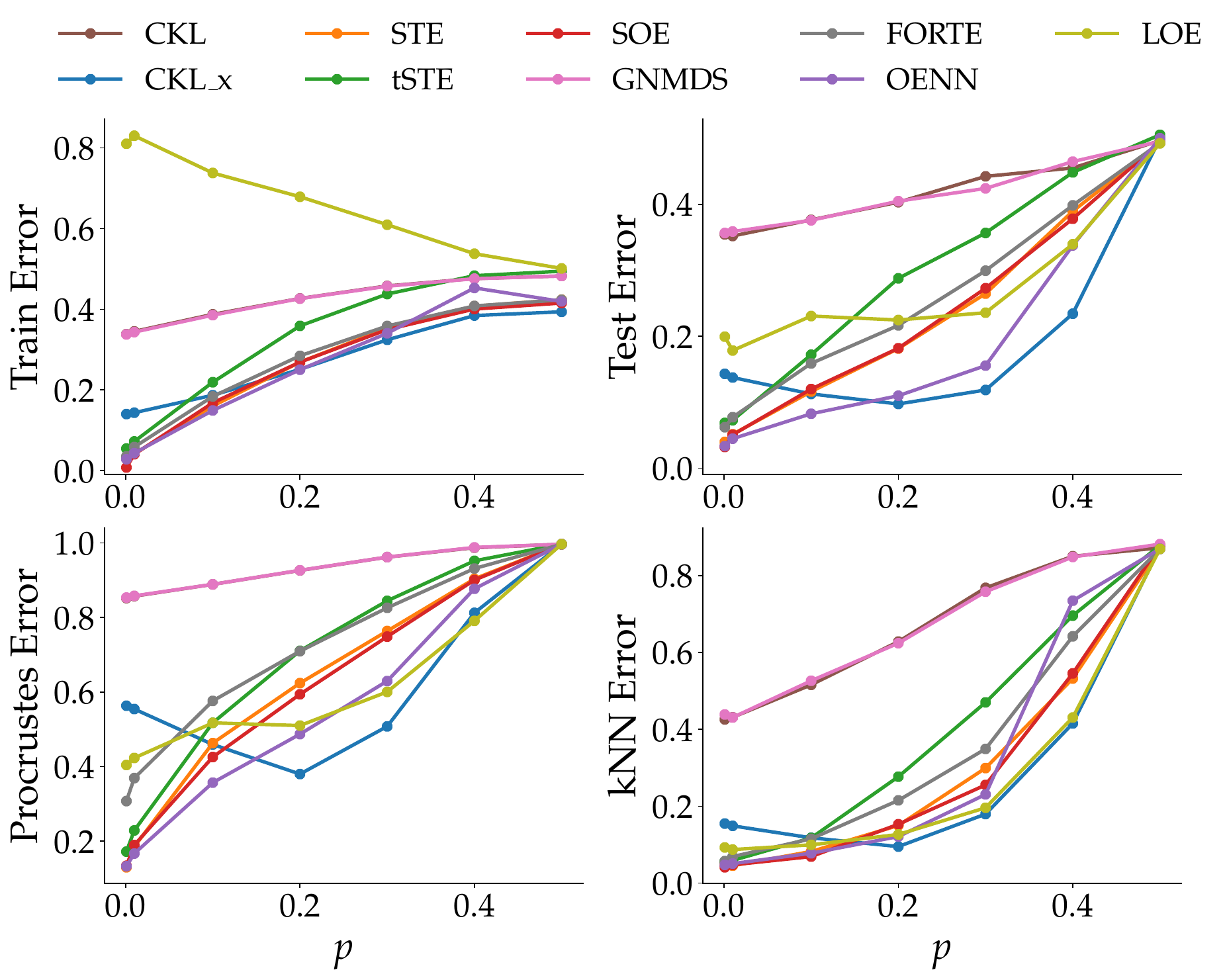}
\caption{{\bfseries USPS: }Increasing $p$ of Bernoulli Noise. The value of triplet multiplier $\lambda$ is $2$.}
\end{figure*}

\begin{figure*}[!htb]
\centering
\includegraphics[width=.7\textwidth]{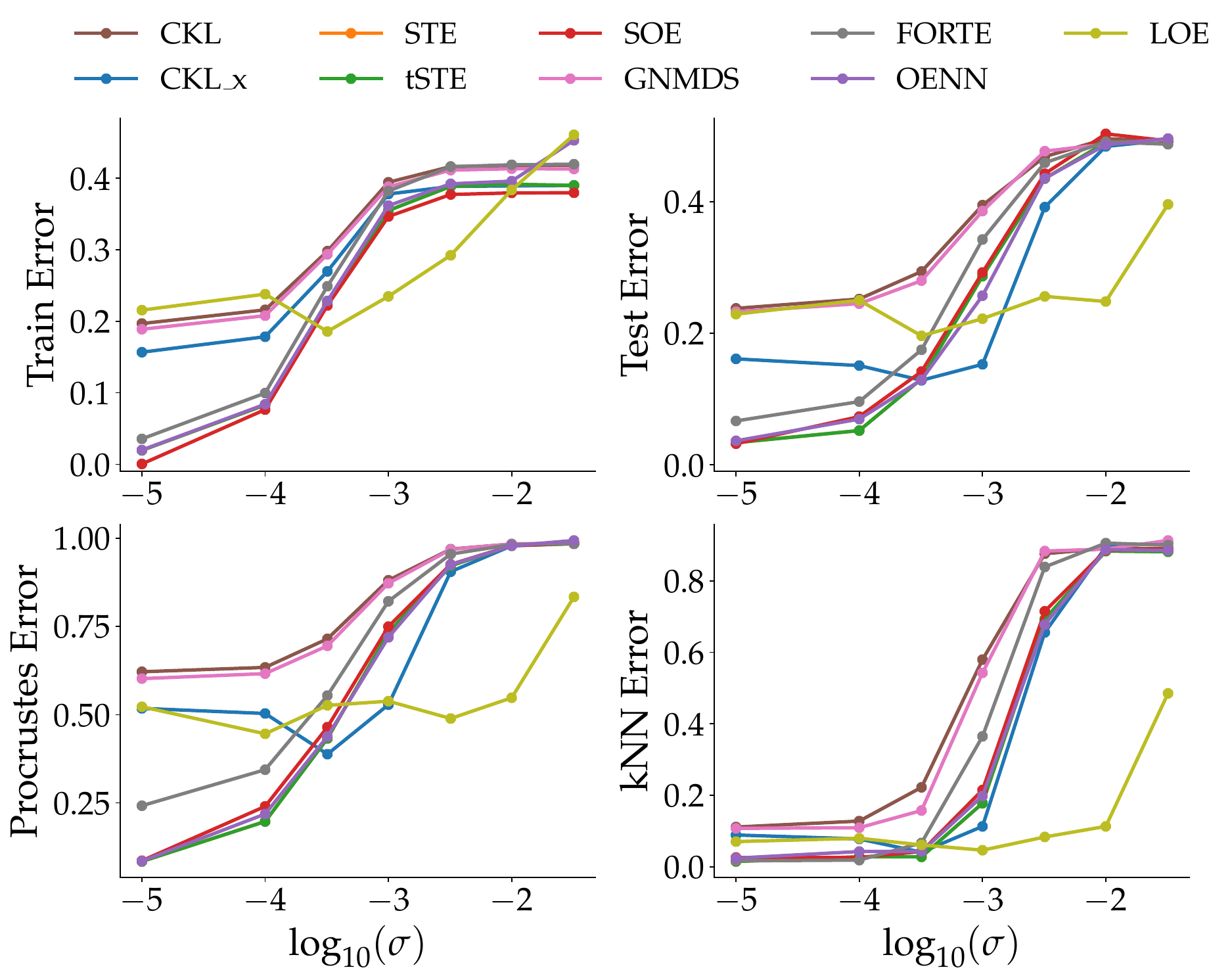}
\caption{{\bfseries CHAR: }Increasing $\sigma$ of Gaussian Noise. The value of triplet multiplier $\lambda$ is $2$.}
\end{figure*}

\begin{figure*}[!htb]
\centering
\includegraphics[width=.7\textwidth]{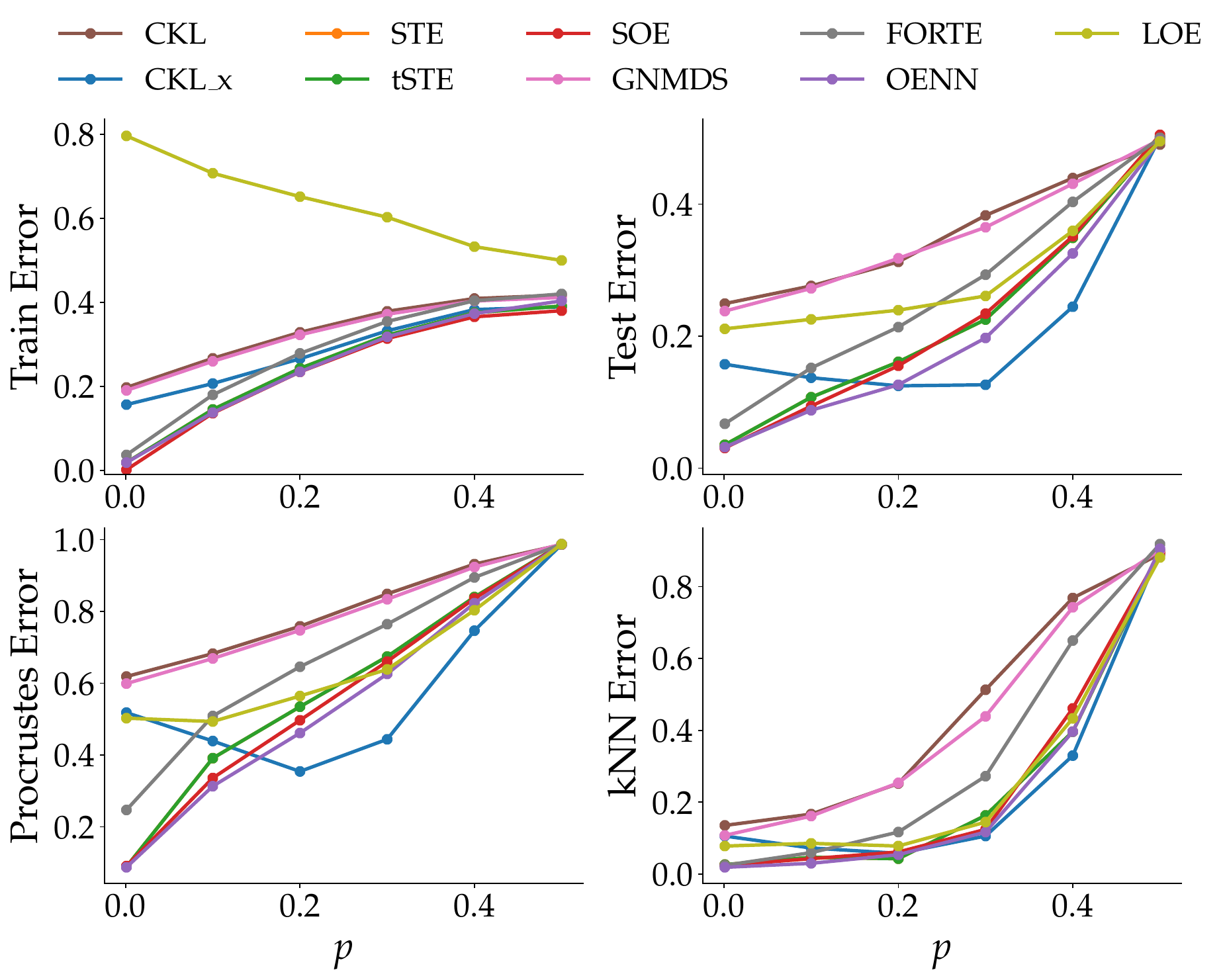}
\caption{{\bfseries CHAR: }Increasing $p$ of Bernoulli Noise. The value of triplet multiplier $\lambda$ is $2$.}
\end{figure*}

\newpage
\FloatBarrier
\section{INCREASING EMBEDDING DIMENSION}

\begin{figure}[!htb]
\centering
\includegraphics[width=.7\textwidth]{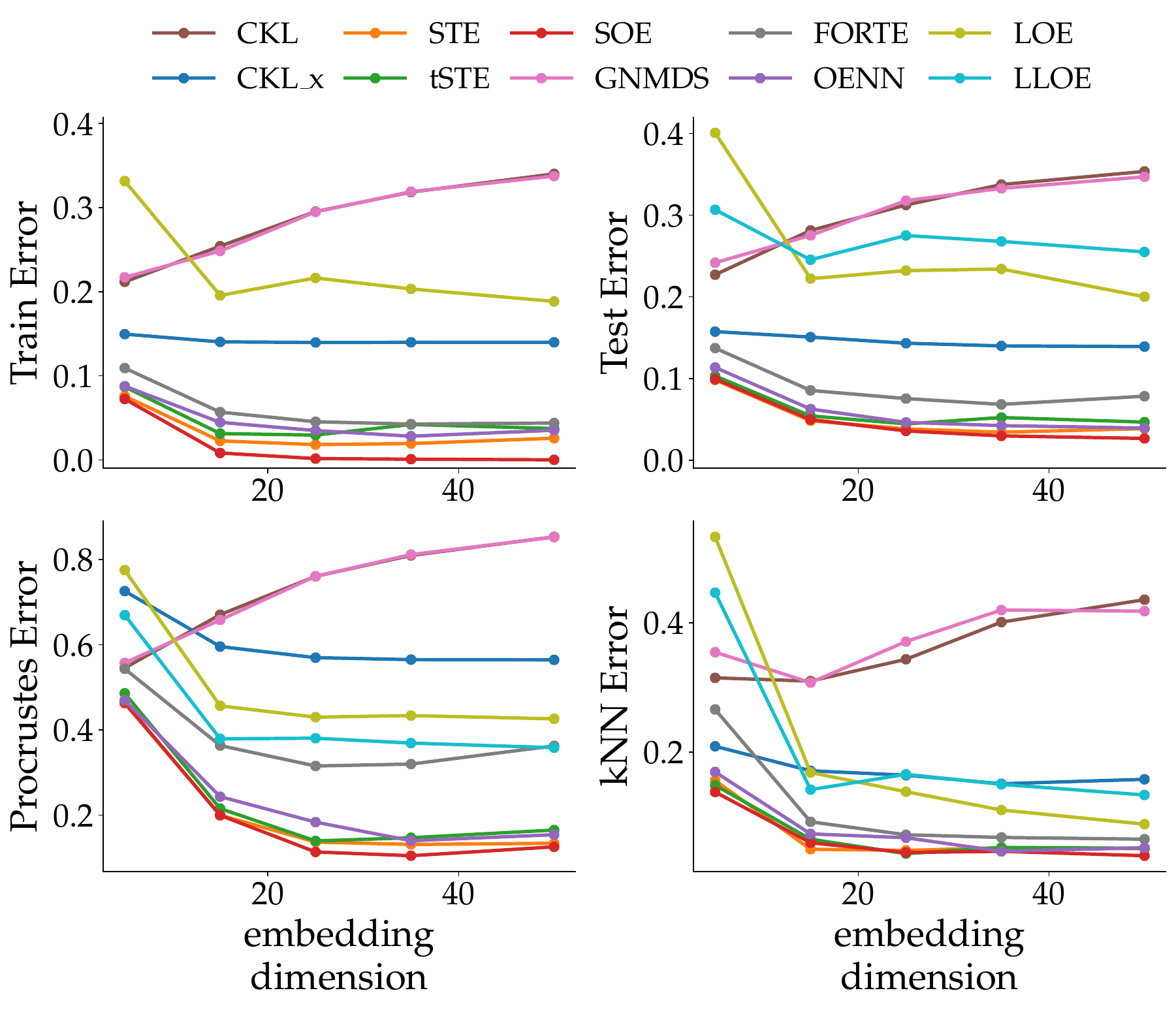}
\caption{\textbf{USPS:} Increasing Embedding Dimension. The quality of embeddings is evaluated using four quality assessment measures.}
\end{figure}


%

\begin{figure}[!htb]
\centering
\includegraphics[width=.7\textwidth]{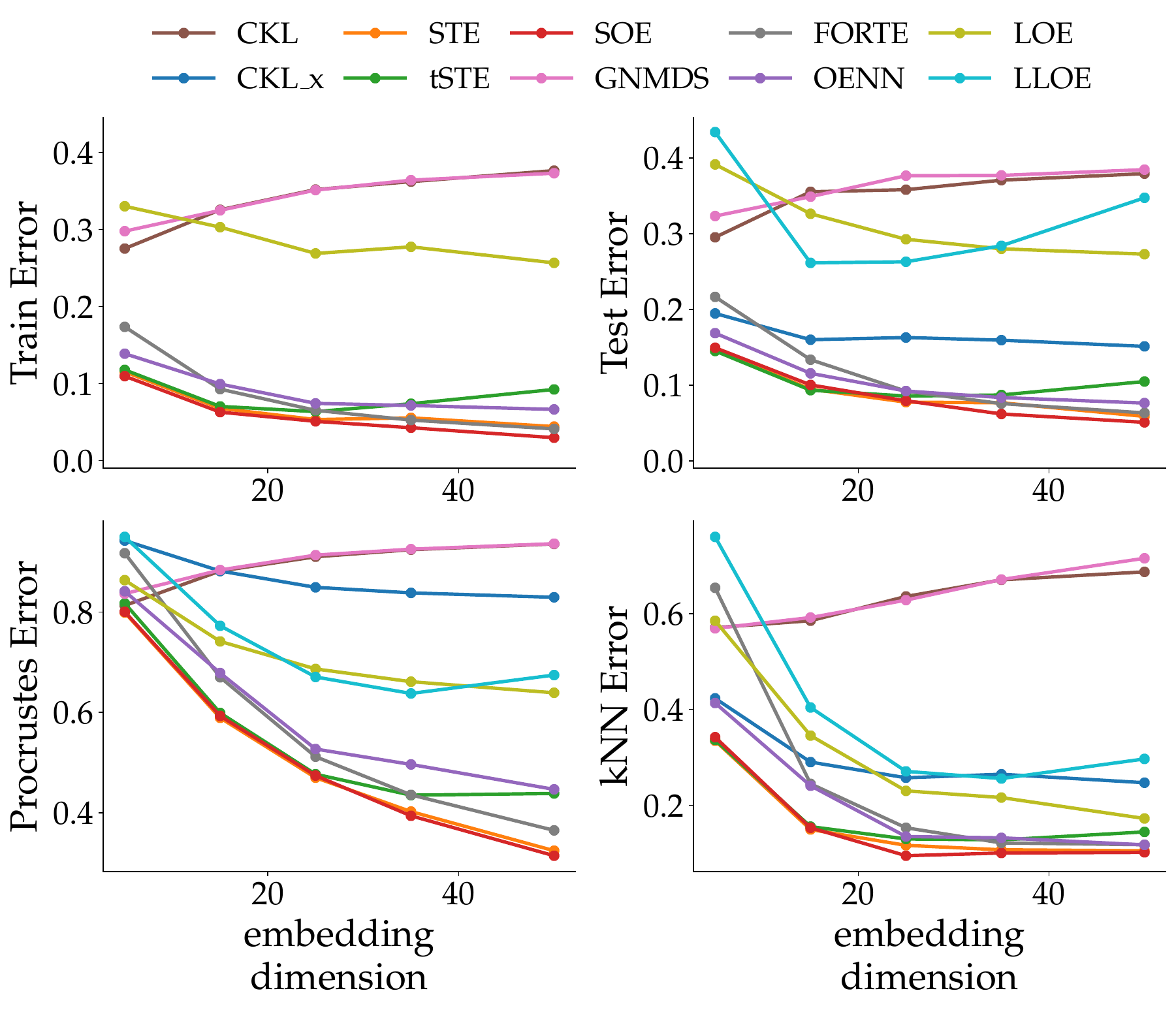}
\caption{\textbf{KMNIST: }Increasing Embedding Dimension. The quality of embedding output is evaluated using four evaluation criteria.}
\end{figure}




\begin{figure}[!htb]
\centering
\includegraphics[width=.7\textwidth]{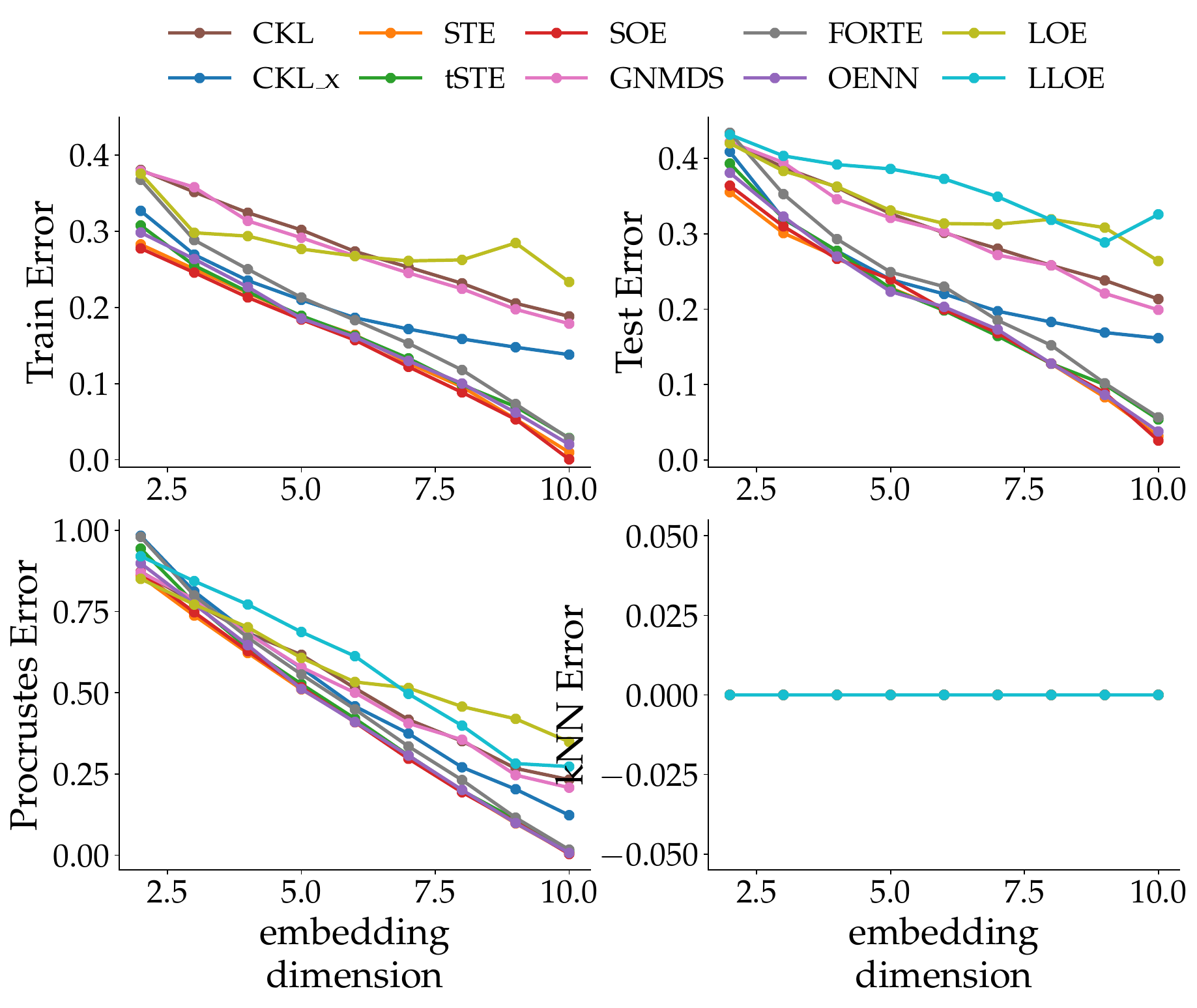}
\caption{\textbf{Uniform: }Increasing Embedding Dimension. }
\end{figure}


\newpage
\FloatBarrier
\section{INCREASING NUMBER OF POINTS}\label{supp:increasing_n}

\begin{figure}[!htb]
\centering
\includegraphics[width=.99\textwidth]{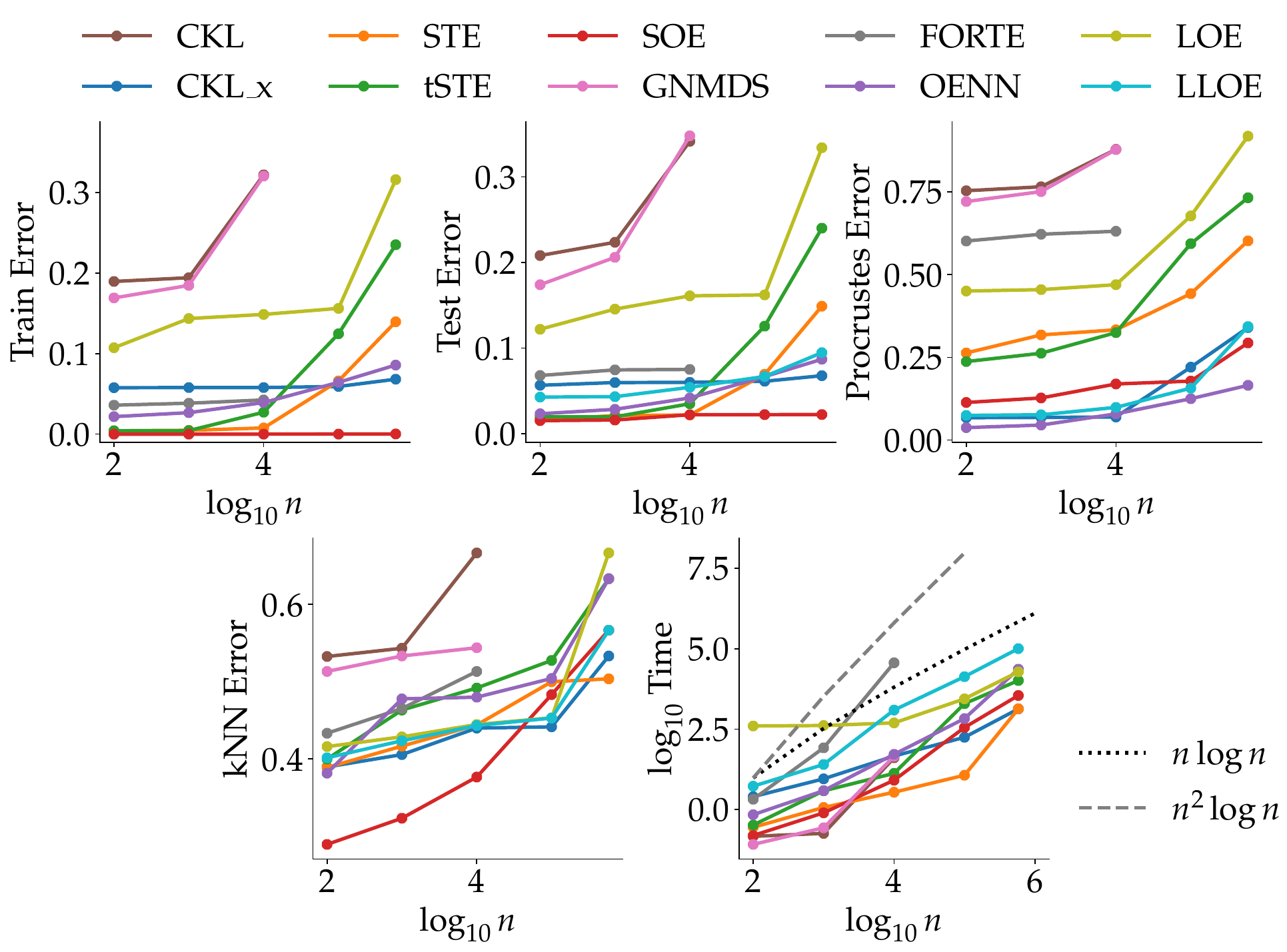}
\caption{\textbf{Covertype: }Increasing number of points. The quality of embedding and running time is reported with respect to the increasing number of points.}
\end{figure}
\begin{figure}[!htb]
\centering
\includegraphics[width=.99\textwidth]{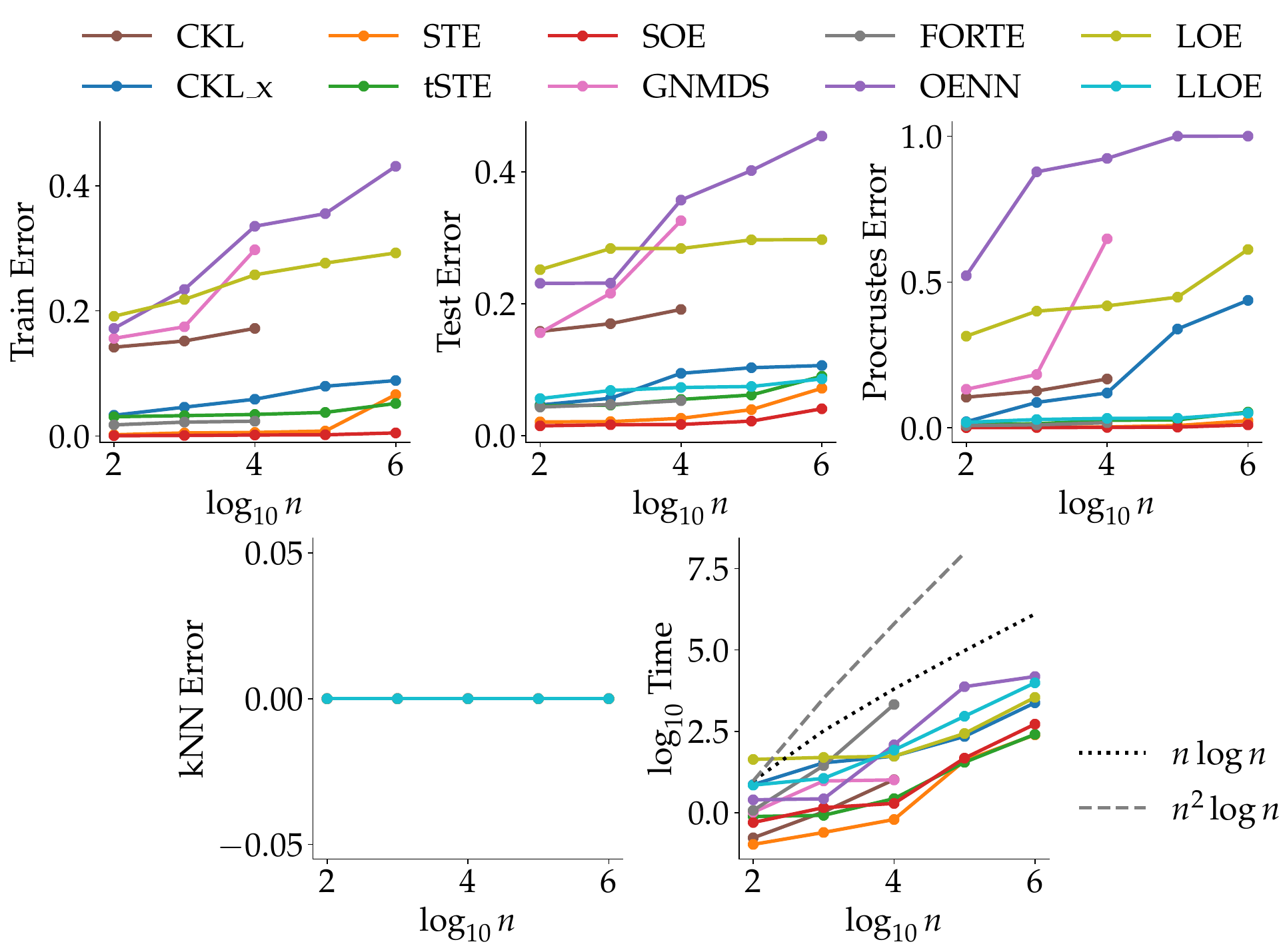}
\caption{\textbf{Uniform: }Increasing number of points. kNN Error is zero, since no labels are used in this dataset.}
\end{figure}

\newpage
\FloatBarrier
\section{INCREASING ORIGINAL DIMENSION}

\begin{figure*}[!htb]
\centering
\includegraphics[width=.99\textwidth]{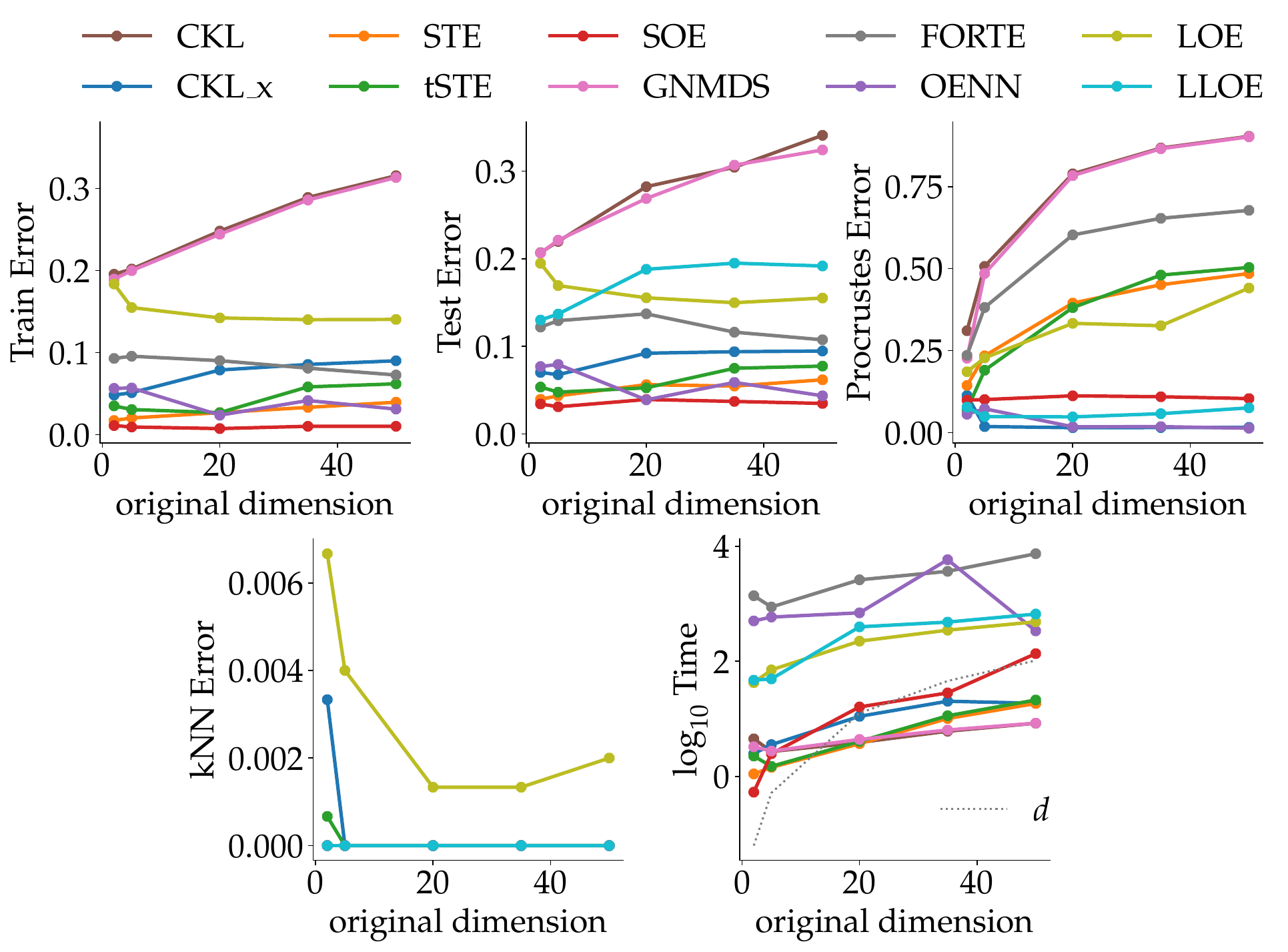}
\caption{\textbf{Gaussian Mixture: }Increasing Original Dimension. Surprisingly, LOE performs very badly on this dataset. Recall, that this dataset consists of two well separable normal distributions.}
\end{figure*}
\begin{figure*}[!htb]
\centering
\includegraphics[width=.99\textwidth]{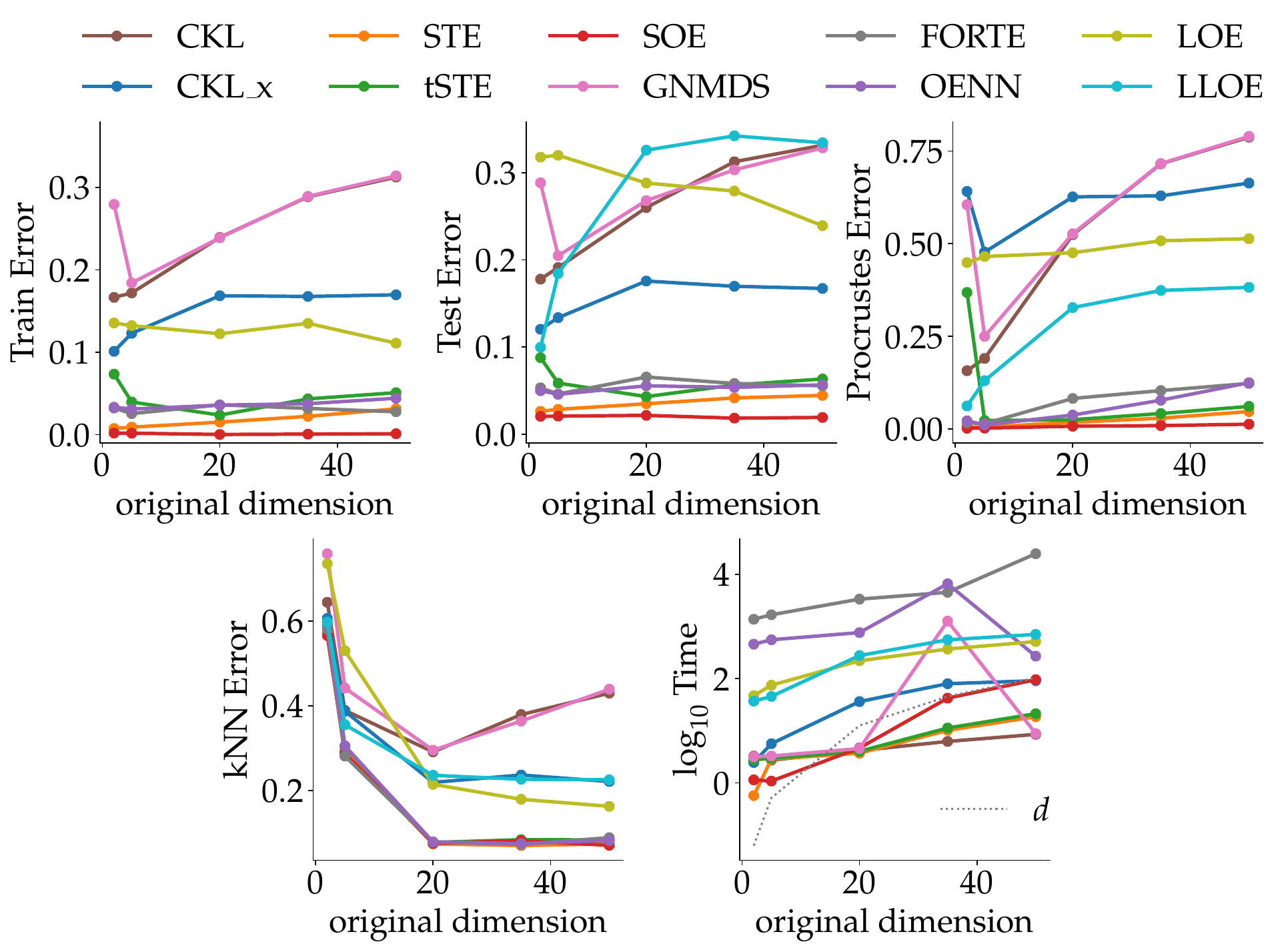}
\caption{\textbf{MNIST: } Increasing Original Dimension. For this experiment, we use a PCA projection of $5,000$ MNNIST points into the corresponding dimension as the ground-truth to generate the train triplets.}
\end{figure*}

\newpage

\FloatBarrier

\section{DETAILS ON ALL METHODS}
\label{sec:method_details}

The ordinal embedding problem has first appeared in the machine learning literature as Generalized Non-metric Multi Dimensional Scaling (GNMDS) \citep{agarwal2007generalized}. Non-Metric Multi-Dimensional Scaling (MDS) refers to the problem of finding a Euclidean embedding, when the rank order of pairwise similarities is given~\citep{shepard1962analysis, kruskal1964multidimensional}. The proposed method by Shepard and Kruskal is a very popular method in psychology and psychophysics. GNMDS provides a generalization to the Non-metric MDS by considering a set of quadruplet questions, instead of the full rank of distances, as the input. The quadruplet $(i,j,k,l)$ implies that the distance of $x_i$ to $x_j$ should be smaller than the distance of $x_k$ to $x_l$. Having access to $\bigO(n^2 \log(n))$ quadruplets, one can reconstruct the full rank of pairwise distances. Thus, assuming a smaller subset of quadruplet information is a generalization of the Non-Metric MDS. There have been various approaches to the problem in the last decade. However, they all share the same spirit as they optimize a loss function to ensure that triplet (or quadruplet) information are satisfied. We briefly discuss the various objective functions and their properties.

\subsection{Detailed Description of the Ordinal Embedding Neural Network (OENN)}
\begin{figure*}[t!]
    \centering
     \begin{subfigure}[t]{0.45\textwidth}
         \centering
         \includegraphics[width=0.7\textwidth, angle=90]{img/drawings/architectureRot90}
         \caption{}
         \label{fig:NNEM_1}
     \end{subfigure}
    ~
    \begin{subfigure}[t]{0.45\textwidth}
         \centering
         \includegraphics[width=0.7\textwidth, angle=90]{img/drawings/EmbNEtsvg1Rot90}
         \caption{}
         \label{fig:EmbNet}
     \end{subfigure}
     ~
     \begin{subfigure}[t]{0.45\textwidth}
    \centering
    \includegraphics[width=1.0\textwidth]{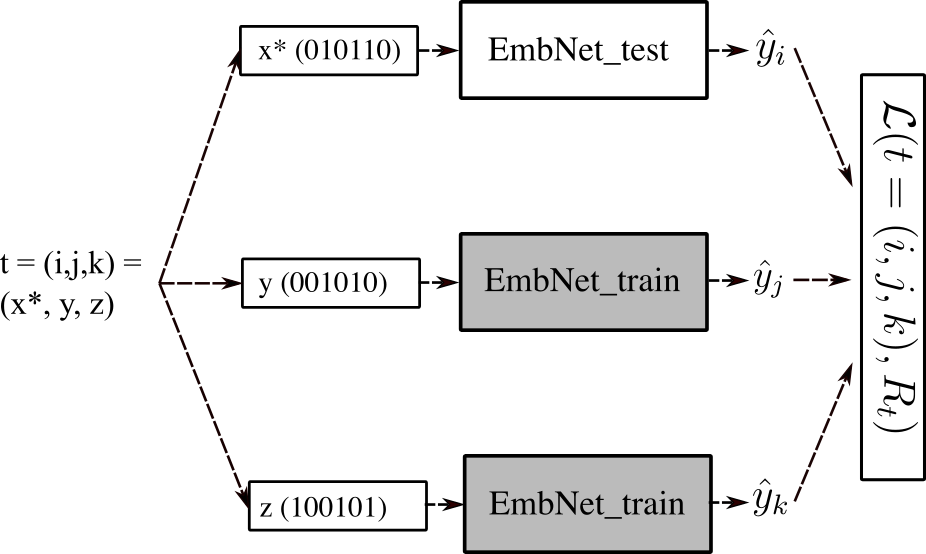}
    \caption{}
    \label{fig:inference}
\end{subfigure}
\caption{(a) The architecture of Ordinal Embedding Neural Network (OENN). As example a triplet (22, 10, 37) and its answer $R_t$ are  fed to the architecture. (b) The EmbNet neural network, which is used as a building blocks of ordinal embedding architecture. (c) Inference architecture - $x^*$ denotes an item from the test set, $y,z$ denote items from the train set. The parameters of $\text{EmbNet}_{\text{train}}$are frozen.}
\end{figure*}
 Our proposed architecture is inspired by the recent line of work on  contrastive learning
 ~\citep{wang2014learning,schroff2015facenet,cheng2016person,hoffer2015deep}. 
 Figure~\ref{fig:NNEM_1} shows a sketch of our  proposed network architecture. 
 The central sub-module of our architecture is what we call the embedding network (\textbf{\textit{EmbNet}}): one such network takes a certain encoding of a single data point $x_i$ as input (typically, an encoding of its index $i$, see below) and outputs a $d$-dimensional representation $\hat{y_i}$ of data point $x_i$. The EmbNet is replicated three times with shared parameters. The overall OENN network now takes the \textbf{\textit{indices}} $(i,j,k)$ corresponding to a triplet $(x_i,x_j,x_k)$ as an input. 
 It routes each of the indices $i,j,k$ to one of the copies of the EmbNet, which then return the $d$-dimensional representations $\hat{y_i}, \hat{y_j} , \hat{y_k}$, respectively (cf. Figure~\ref{fig:NNEM_1}). 
 The three sub-modules are trained jointly using the triplet hinge loss, as described by the following objective function:
\begin{equation}
    \label{eq:TripletLoss_1} 
      \gL(T)= \frac{1}{\vert T \vert}\sum_{(i,j,k)\in T} {\max \left \{ \Vert \hat{y_i} - \hat{y_j} \Vert^2  - 
\Vert \hat{y_i} - \hat{y_k} \Vert^2 +1, 0 \right \}}
\end{equation}
Note that this optimization problem is not a relaxation to the OE problem. Rather, its an equivalent one \cite{bower2018landscape}. Meaning that every global optima of this optimization problem is a feasible solution to the ordinal embedding problem and vice versa (up to a scale).

\textbf{\textit{On the choice of input representation:}}
Since we do not have access to any informative low-level input representations, the choice of input representations presents a challenge to this approach. However, we leverage the expressive power of neural networks~\citep{leshno1993multilayer,barron1993universal} and their ability to fit random labels to random inputs~\citep{zhang2016understanding} to motivate our choice of input encoding. Since our main goal is to find representations that minimize the training objective, we believe that completely arbitrary input representations are a viable choice.  

One such input representation could be the one-hot encoding of the index (where point $i$ is encoded by a string $\hat{x}_{i}$ of length $n$ such that $\hat{x}_{i}(l) = 1$ if $l = i$ and $0$ otherwise). The advantage of choosing such a representation is that it is memory efficient in the sense that there is no need to additionally store the representations of the items. However, under this choice of representation the length of the input vectors grows linearly with the number $n$ of input items. 
As one of the central contributions of our work is to perform ordinal embedding in large scales, we consider a more efficient way: we represent each item by the binary code of its index, leading to a representation length of  $\log{n}$.  Such a representation retains the memory efficiency of the one-hot encoding (in the sense as discussed above) but improves the length of the input representation from $n$ to  $\log{n}$. As we will see below, this representation works well in practice. However, note that there is nothing peculiar about this choice of binary code. Our simulations (see Subsection~\ref{sec:simWidth}) suggest that unique representations for items generated uniformly, randomly from a unit cube in $\mathbb{R}^{\alpha = \Omega(\log n)}$ can be used as the input encoding.

\textbf{\textit{Structure of EmbNet:}} Figure~\ref{fig:EmbNet} shows the schematic of the EmbNet. We propose a simple network with three fully-connected hidden layers and ReLu activation function. The final layer is a linear layer that takes the output of third hidden layer and produces the output embedding. The input size to the network is $\ceil{\log{n}}$ and each hidden layer contains $w$ nodes. 
The output layer has $d$ nodes to produce embeddings in $\sR^d$. 
The input and output size, $\ceil{\log{n}}$ and $d$, are pre-determined by the task. Thus the only independent parameter of the network is the width $w$ of hidden layers. Our experiments (see Subsection~\ref{sec:simWidth}) demonstrate that the hidden layer-width $w$ should grow logarithmically to the number of items $n$ to produce good embedding outputs. 

\textbf{\textit{Generalization to new data points:}} While the primary goal of embedding methods is to find a representation of the given training set, it is sometimes desirable to be able to extend the representation to new data points. 
A naive approach would be to re-train the OENN with the appropriate parameter settings (see Section \ref{sec:simWidth}) with the combined set of training and test triplets.
However, our approach features an elegant way to infer embeddings of a set of test items. Let $X' =~ \left \{ x^*_{1}, \cdots x^*_{k} \right \}$ denote a set of test items observed via a set of triplets $T'$ of the form $(x,y,z)$ where at least one of the items in each triplet comes from $X'$ and the rest could arise either from $X$ or $X'$. Such a triplet encodes the usual triplet relationship: $x$ is closer to $y$ than $z$.

Our inference procedure (shown in Figure \ref{fig:inference}) to obtain embeddings from $T'$ is as follows. Recall that our architecture is composed of 3 identical EmbNet components with shared weights. Consider a OENN trained on $T$ and denote its (trained) EmbNet component as $\text{EmbNet}^*$. We define $\text{EmbNet}_\text{test}$ to be an instance of $\text{EmbNet}^*$ with randomly re-initialized weights and $\text{EmbNet}_\text{train}$ to be an instance of $\text{EmbNet}^*$ with weights un-touched. The inference architecture, is then constructed dynamically for any triplet from $T'$ in the following manner: items from $X'$ in the triplet are provided as inputs to $\text{EmbNet}_\text{test}$, while items from $X$ are provided as inputs to $\text{EmbNet}_\text{train}$ . We then proceed to train the network in the same fashion as before over $T'$ except that we freeze the parameters of $\text{EmbNet}_\text{train}$. Note that the training examples are unaffected in the inference procedure. The procedure produces embeddings of test items that satisfy the triplet constraints containing items both from $X$ and $X'$. Our experiments (Section \ref{sec:gen_new_items_oenn}) indicate that the quality of embeddings provided by this procedure is quite satisfactory. We evaluate this using the triplet error. 


\textbf{Difference to previous contrastive learning approaches:}
As described earlier, our architecture is inspired by ~\citet{wang2014learning,hoffer2015deep}. However, there are fundamental differences: 1) we have no access to representations for the input items $x_1,..,x_n$ and our network takes completely arbitrary representations for the input items. 
2) In the previous work, triplets are always contrastive, which is of the form $(x, x^+, x^-)$ while simply gather triplets involving arbitrary sets of three points. Indeed, obtaining contrastive triplets requires additional information, either the class labels or more explicit similarity information between objects. 
\vspace{-2mm}

\subsection{On the choice of the loss function}
In our method, we use the hinge loss as the choice of our loss function. In what follows, we justify this choice by showing that the resulting optimization by using the hinge-loss does not constitute a relaxation to the ordinal embedding problem. Rather it's an equivalent one. 

The problem of ordinal embedding --- finding an embedding $X = \left \{ x_1, x_2, .., x_n \right \} \in \mathbb{R}^d$  that satisfies a set of given triplets, $\mathcal{T}$ - can be phrased as a quadratic feasibility problem \citep{bower2018landscape} as shown in Equation~\ref{eqn:OEasQF}. 
\begin{equation}
    \label{eqn:OEasQF}
    \textrm{find } X \textrm{ subject to } X^T P_{i,j,k} X > 0 \textrm{ } \forall (i,j,k) \in \mathcal{T}.
\end{equation}
Each matrix $P_{i,j,k}$ corresponds to a triplet constraint that satisfies, 
$$\vert \vert x_i - x_j \vert \vert^2  > \vert \vert x_i - x_k \vert \vert^2 \iff X^T P_{i,j,k} X > 0 $$

Every feasible solution to \ref{eqn:OEasQF} is a valid solution to the problem of ordinal embedding. 

An equivalent way to solve \eqref{eqn:OEasQF} (i.e., find a feasible solution of \ref{eqn:OEasQF}) is to find the global optima of the constrained optimization problem \citep{bower2018landscape} given by \eqref{eqn:OEwithHinge}: 
\begin{equation}
    \label{eqn:OEwithHinge}
    \min \limits_{X \in \mathbb{R}^{nd}}  \sum \limits_{(i,j,k) \in \mathcal{T}} \max \left \{ 0, 1 - X^T P_{i,j,k} X \right \}
\end{equation}

This it true because every feasible solution to (\ref{eqn:OEasQF}) can be scaled to attain global optima of (\ref{eqn:OEwithHinge}) and every global optima of (\ref{eqn:OEwithHinge}) is a feasible solution of (\ref{eqn:OEasQF}) \citep{bower2018landscape}. Moreover, in optimization (1), any positive scaling of a feasible point $X$ is a solution to (1) as well, whereas in optimization (2) this effect is eliminated. 

The hinge loss, therefore, satisfies the nice property that using the hinge loss to solve the ordinal embedding problem is not a relaxation but rather an equivalent problem. 

\subsection{Choice of parameters and hyper-parameters}
\label{sec:simWidth}
Our network architecture depends on a few parameters: the number of layers $l$, the width of the hidden layers $w$, length of the input encoding $\alpha$ and the dimension of the embedding space $d$. To reduce the number of independent parameters of the network, we simply fix the number of layers to $3$, where the number of units is the same in each of these layers. The embedding dimension is determined by the given task. Additionally, the training procedure requires the setting of a few hyper-parameters: choice of the optimizer, batch size, and learning rate. We use ADAM \citet{kingma2014adam} to train our network and use a batch size of $\min (\text{\# triplets}, 50,000)$ and a learning rate of $0.005$ for all the experiments. 

\textbf{On the width of the hidden layers $w$:}
The width $w$ of the hidden layers depends on the input dimension $d$ and the number of items $n$. To investigate this dependence we use toy datasets where the dimension of the input space could be controlled. We ran simulations on 5 different datasets, specifically, we used the $d$ principal components of MNIST\citep{mnist}, USPS\citep{usps}, CHAR\citep{char_covtype}, Mixture of Gaussian in $\mathbb{R}^d$ and data sampled uniformly from a unit cube in $\mathbb{R}^d$.

We perform two sets of experiments. First, we fix the number of points and generate datasets in an increasing number of dimensions, $d$. We generate random triples from these datasets and use our model with increasing $w$ to embed them back into $\mathbb{R}^d$. Our simulations demonstrate that the width of the hidden layers needs to grow linearly with the embedding dimension. Similarly, fixing the input and the embedding dimension, our simulations demonstrate that $w$ needs to grow logarithmically with $n$.
Therefore, in all of our experiments, we set, $$w = \max(120, 2 d \log n).$$ 

\textbf{Dependence of the layer width $w$ on the number $n$ of items.}
Recall that in our setting, the neural network is used to solve a memorization task. The goal of the network is to fit completely arbitrary inputs --- the index of the item --- to outputs in $\mathbb{R}^d$. Therefore, it is reasonable to expect that with an increasing number $n$ of items, a larger network is required to address the complexity of the task. We can either increase the representational power of the network by adding more layers or by increasing the width of layers. We fix the number of layers to $3$ and investigate the dependence of layer width $w$ on $n$. 

For each dataset, we choose an exponentially increasing number of items ($n$) in $\mathbb{R}^2$. Given a sample of $n$ points, we generate $n d\log{n}$ triplets (where $d=2$ is the embedding dimension). The embedding network (EmbNet) is constructed with 3 hidden layers, each having $w$ fully-connected neurons with ReLU activation functions. The width $w$ of the layers (= number of neurons in each layer) is increased linearly. For a fixed network and a fixed set of triplets, the experiment is executed 5 times in order to examine the average behavior of the model. 

\begin{figure*}[!htb]
    \centering
      \begin{subfigure}[t]{0.33\textwidth}
         \centering
         \includegraphics[width=1.0\textwidth]{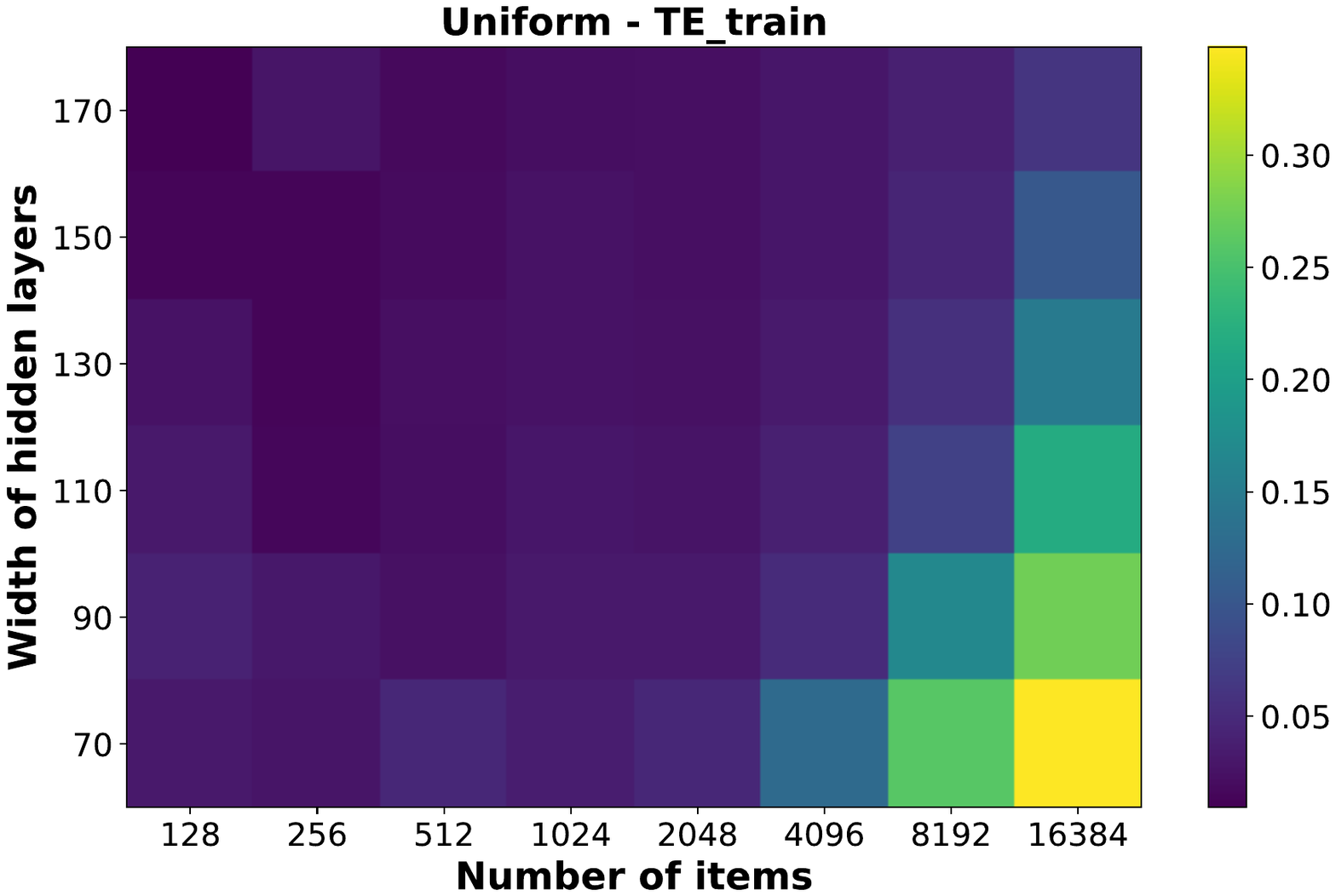}
     \end{subfigure}~\hfill
     \begin{subfigure}[t]{0.33\textwidth}
         \centering
         \includegraphics[width=1.0\textwidth]{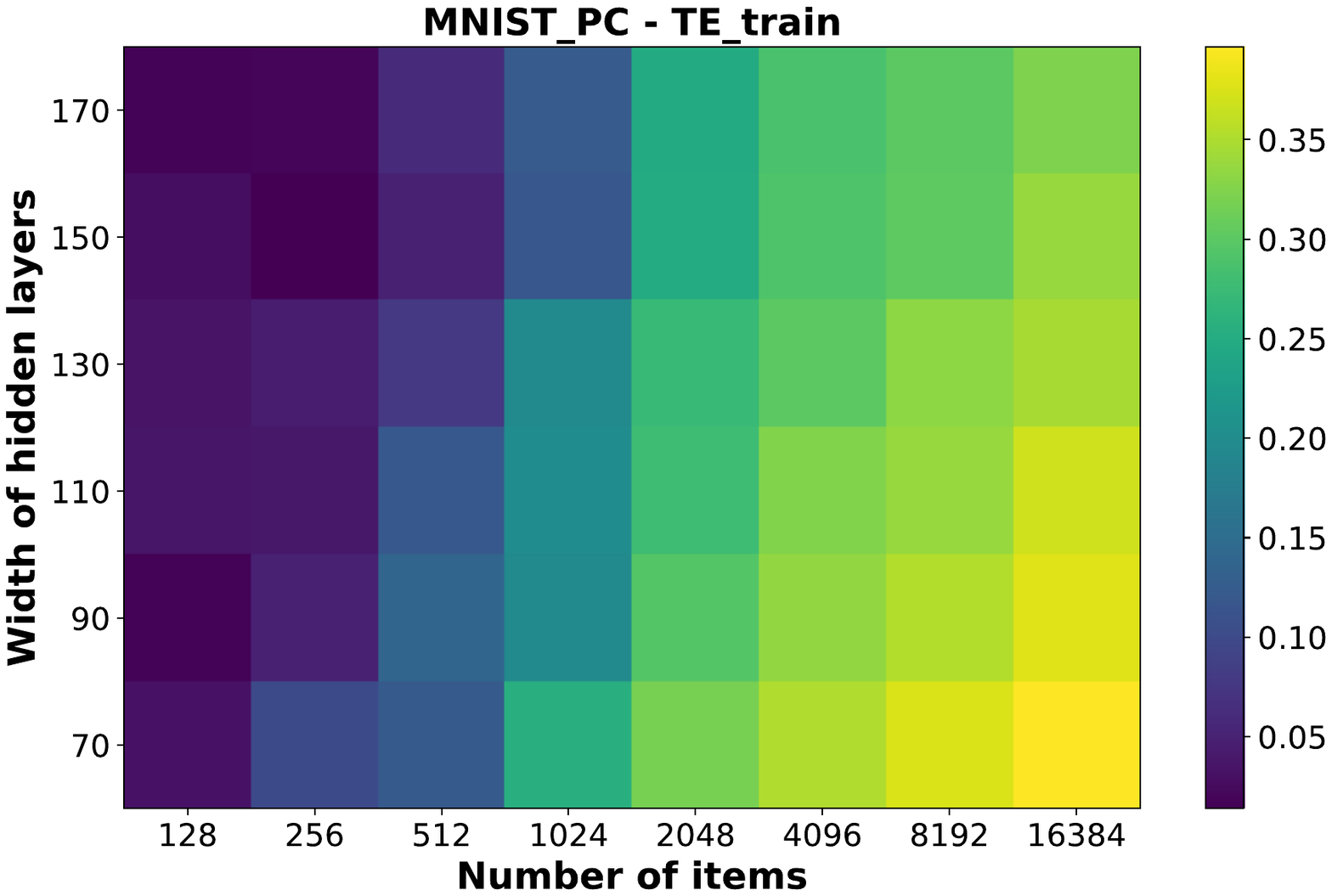}
     \end{subfigure}
    ~ \hfill
          \begin{subfigure}[t]{0.28\textwidth}
         \centering
         \includegraphics[width=1.0\textwidth]{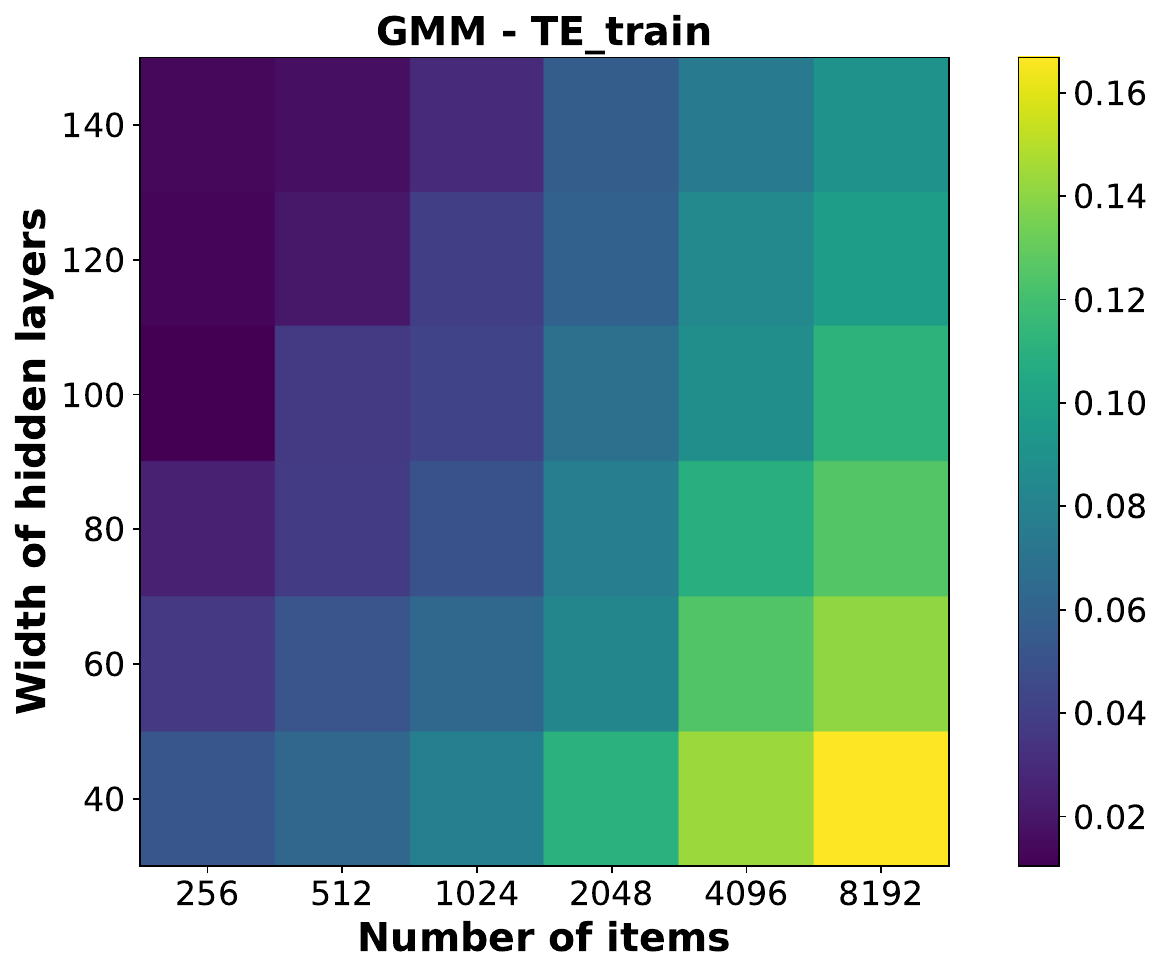}
     \end{subfigure}
    ~
     \caption{Triplet error (encoded by the heat map) with varying number of items and hidden layer size. The $x$ and $y$ axes correspond to the number of items ($n$) and the hidden layer size ($w$) respectively. The datasets used for the experiments is reported on the top of each figure. Note that $x-$axes grows exponentially, while the $y-$axes increases linearly. }
     \label{fig:NetN}
\end{figure*}

Figure~\ref{fig:NetN} shows the training triplet error ($\text{TE}_\text{train}$) of the ordinal embedding with varying number of items and width of the hidden layers for 3 different datasets. The error is reported by means of heat-maps, where warmer colors denote higher triplet error.
Note that the $x-$axis increases exponentially and the $y-$axis increases linearly. The plots clearly establish logarithmic dependence of the hidden layer width on the number of items $n$. 

\textbf{Dependence of the layer width $w$ on the embedding dimension $d$}

Besides the number of items, the embedding dimension is another factor that influences the complexity of ordinal embedding. We expect that the required layer width needs to grow with the embedding dimension. To investigate this dependence, as earlier, we sample $n=1024$ items in $\mathbb{R}^d$ from each dataset with varying input dimension $d$. We generate $nd \log{n}$ triplets based on the Euclidean distances of items. To construct a network, the width $w$ of the layers is chosen from a linearly increasing set. For a fixed network and a fixed set of triplets, the experiment is executed 5 times in order to examine the average behavior of the model.

\begin{figure*}[!htb]
    \centering
      \begin{subfigure}[t]{0.325\textwidth}
         \centering
         \includegraphics[width=1.0\textwidth]{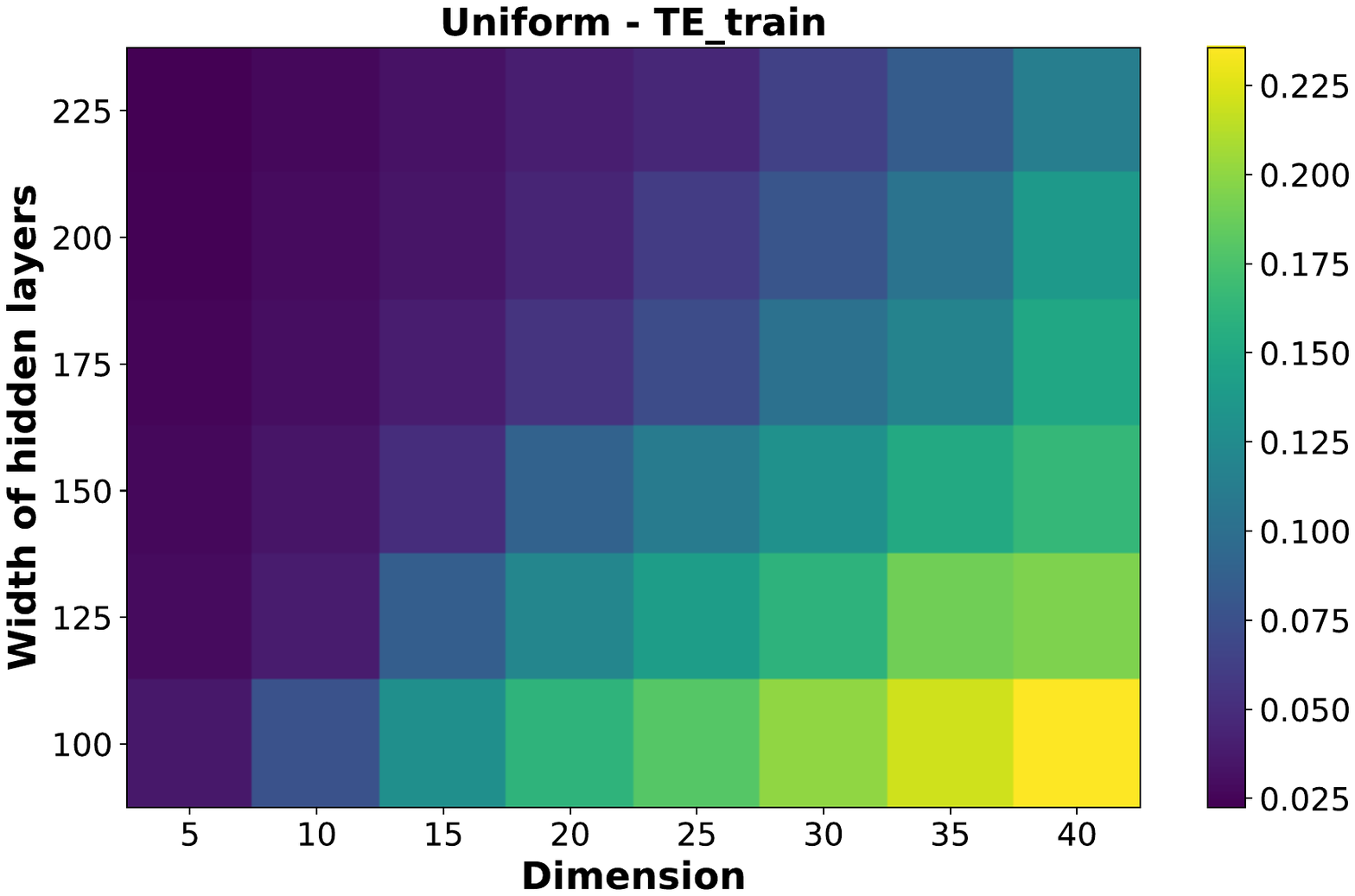}
     \end{subfigure}~\hfill
    \begin{subfigure}[t]{0.325\textwidth}
         \centering
         \includegraphics[width=1.0\textwidth]{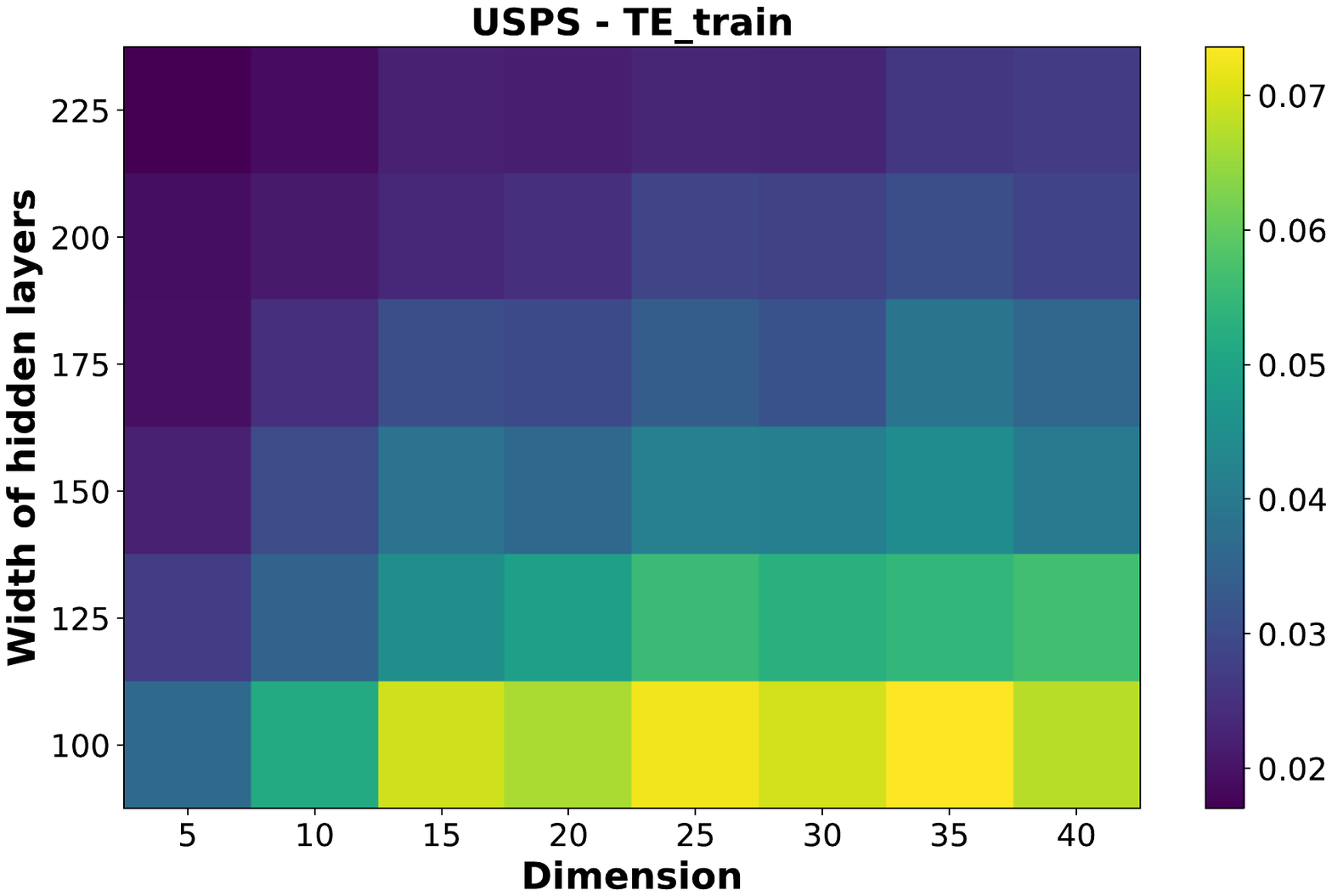}
     \end{subfigure}
    ~\hfill 
    \begin{subfigure}[t]{0.325\textwidth}
         \centering
         \includegraphics[width=1.0\textwidth]{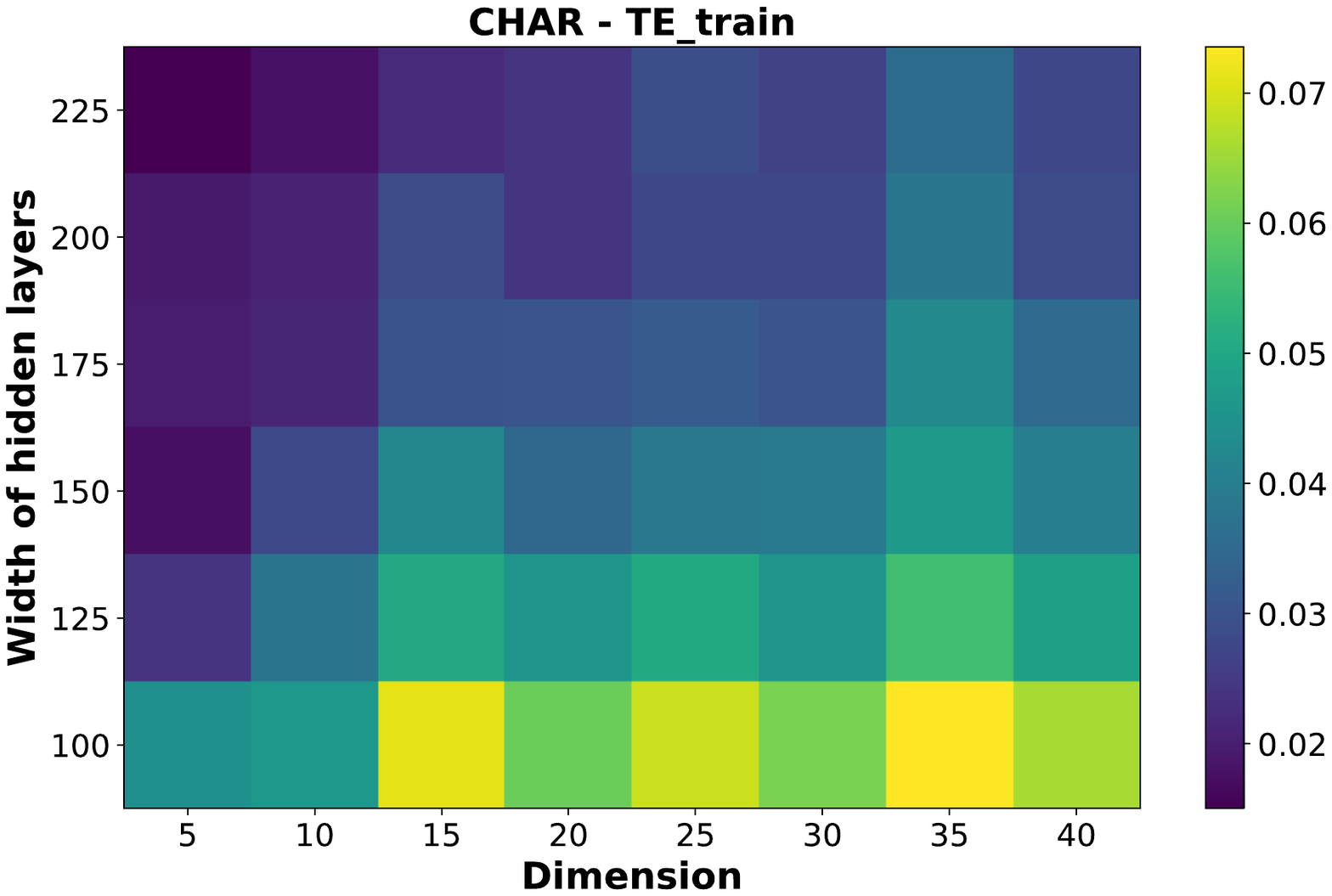}
     \end{subfigure}
    
     \caption{Triplet error (encoded by the heat map) with varying dimensions and hidden layer size. The $x$ and $y$ axes correspond to the number of dimensions ($d$) and the hidden layer size ($w$) respectively. The datasets used for the experiments is reported in the title of each figure. Note that both the axes scale linearly.}
     \label{fig:NetDim}
\end{figure*}

Figure~\ref{fig:NetDim} shows the training triplet error ($TE_{train}$) of the ordinal embedding with varying dimensions and width of the hidden layers for 3 different datasets. There is again a clear line of transition between low and high error regions. In all of the datasets, it can be observed that the layer width has a linear dependence on the embedding dimension.

\textbf{On the choice of the length of input encoding $\alpha$.}
The input to the OENN, as described earlier, can be chosen as a triplet of arbitrary encoding of the items. We ran simulations to establish the dependence of the length of such encoding with the number of items. These simulations show that the size of the input encoding needs to grow logarithmically with the number of items. In all the experiments, we set, $$\alpha = \ceil{\log n}.$$

\begin{figure}[!htb]
    \centering
 \includegraphics[width=0.4\textwidth]{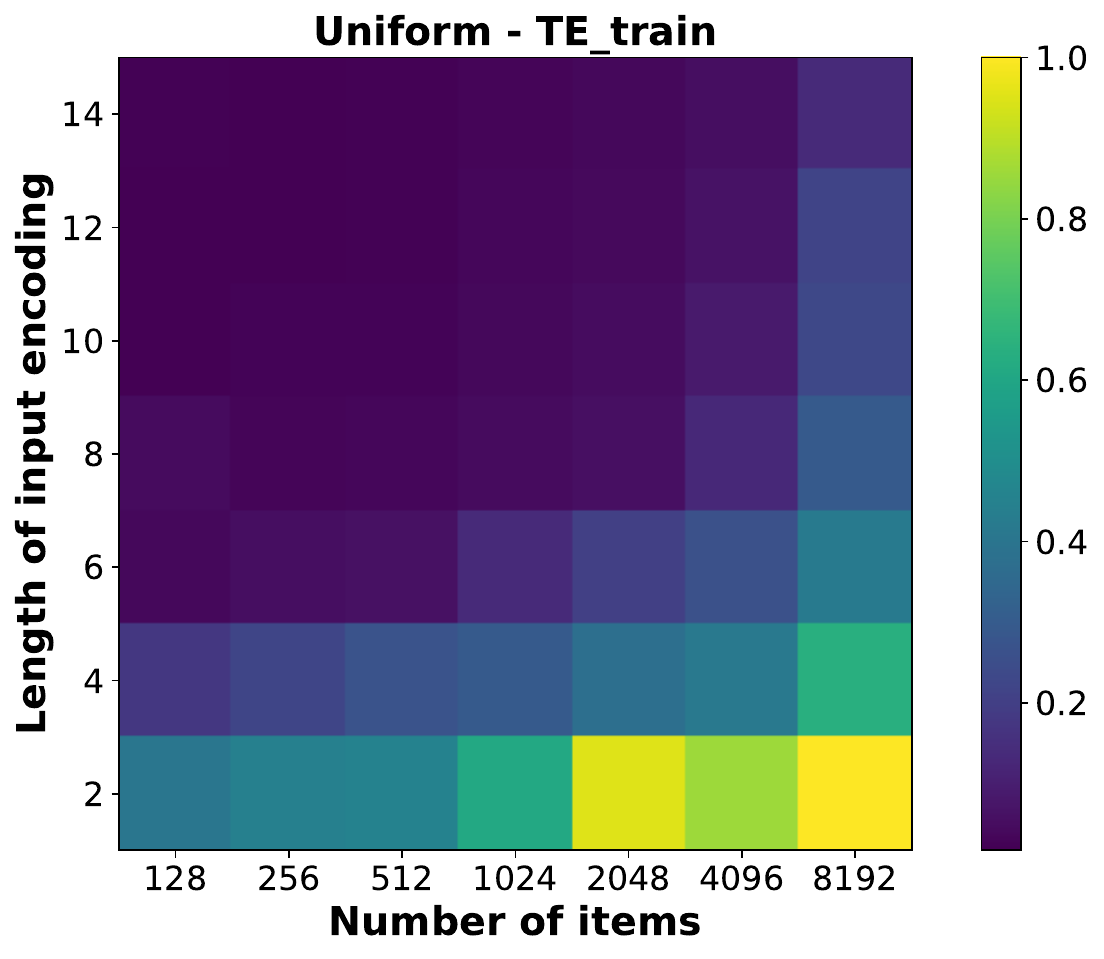}
     \caption{Triplet error (encoded by the heat map) with varying number of items and length of the input encoding. The $x$ and $y$ axes correspond to the number of items ($n$) and the length of the input encoding ($q$) respectively. Note that $x-$axes grows exponentially, while the $y-$axes increases linearly. The color of the heat map represents the triplet error obtained in the training procedure.}
     \label{fig:NetInp}
\end{figure}

\textbf{On the length of input encoding $q$.}
Our algorithm takes random encoding of triplets of items as input and learns a transformation from the input encoding to vectors in Euclidean space of a dimension specified by the task. We run simulations with random input representations of a fixed length $q$ where we choose  each component of the vector uniformly at random from the unit interval. Our simulations show that the size of the input encoding needs to grow logarithmically with an increasing number of items and using arbitrary representations of length $\log n$ suffices to achieve a small training triplet error ($5\%$). To conduct the simulations, we sampled an exponentially increasing number of items $n$ uniformly from a unit square in $R^{2}$. For a fixed set of $n$ items, we generate $nd\log{n}$ triplets (where $d=2$ is the dimension). To isolate the effect of the length of the input encoding, we construct our EmbNet by fixing the width of the hidden layer ($w = 100$). Figure \ref{fig:NetInp} shows the training triplet error ($TE_{train}$) of the ordinal embedding with varying number of items and the size of the input encoding. 
The result shows that logarithmic growth of the size of the input encoding with respect to $n$ suffices to obtain desirable performance. 

\subsection{Experiments on Generalization to new items}
\label{sec:gen_new_items_oenn}
\begin{figure*}[!htb]
    \centering
     \begin{subfigure}[b]{0.39\textwidth}
         \centering
         \includegraphics[width=1.0\textwidth]{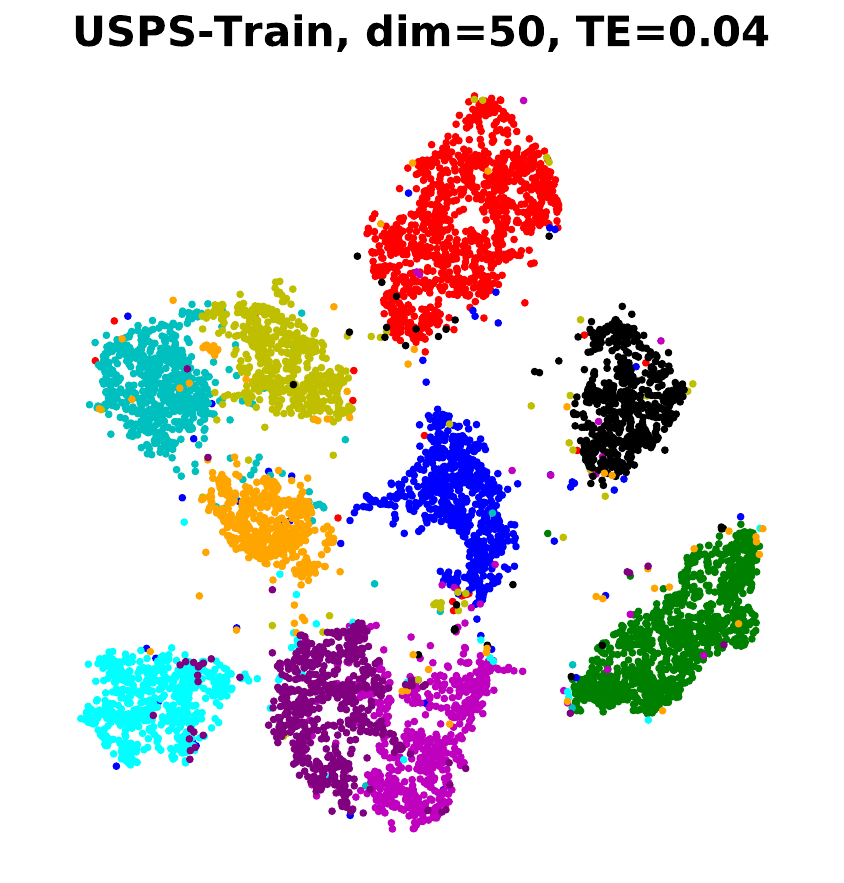}
     \end{subfigure}
      \begin{subfigure}[b]{0.39\textwidth}
         \centering
         \includegraphics[width=1.0\textwidth]{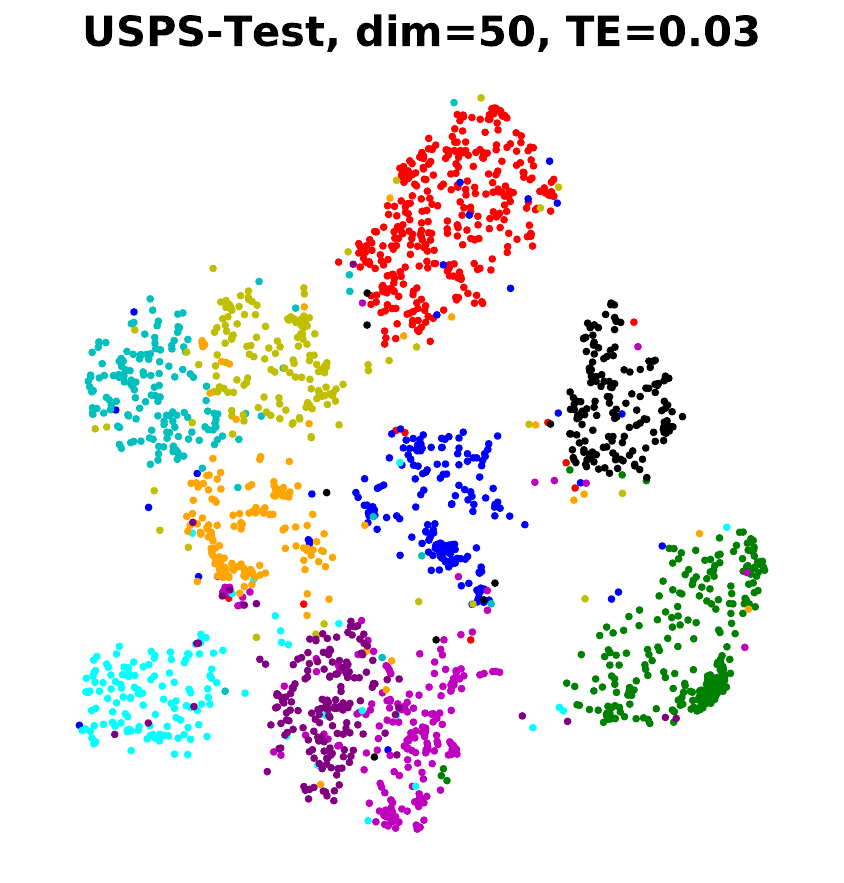}
     \end{subfigure}
     \begin{subfigure}[b]{0.39\textwidth}
         \centering
         \includegraphics[width=1.0\textwidth]{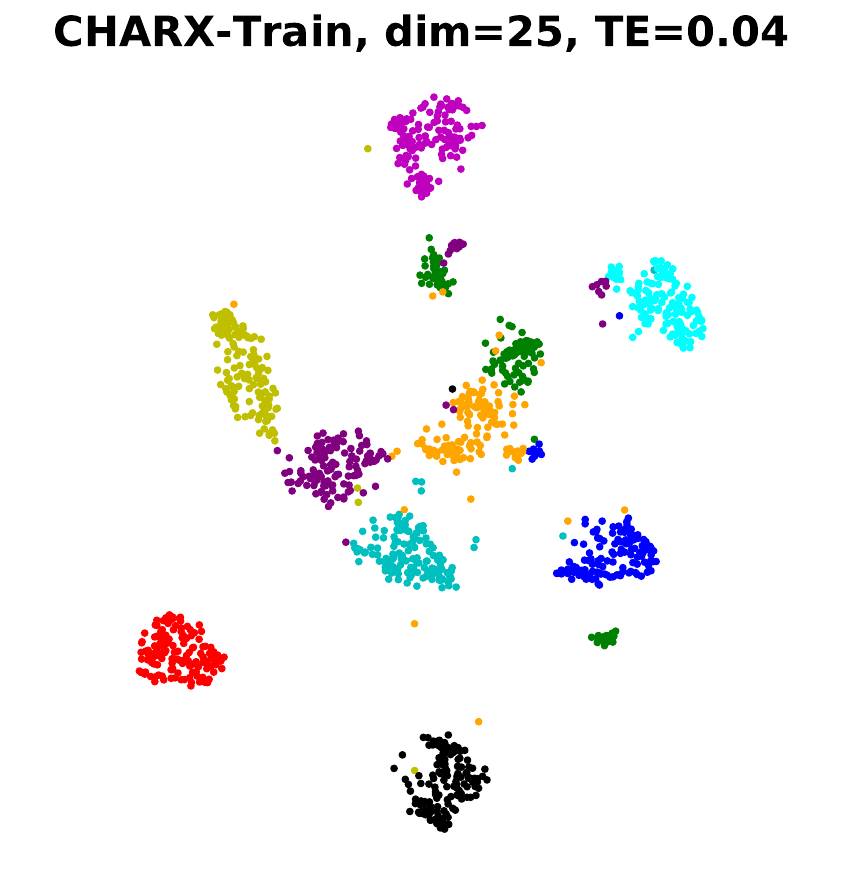}
     \end{subfigure}
      \begin{subfigure}[b]{0.39\textwidth}
         \centering
         \includegraphics[width=1.0\textwidth]{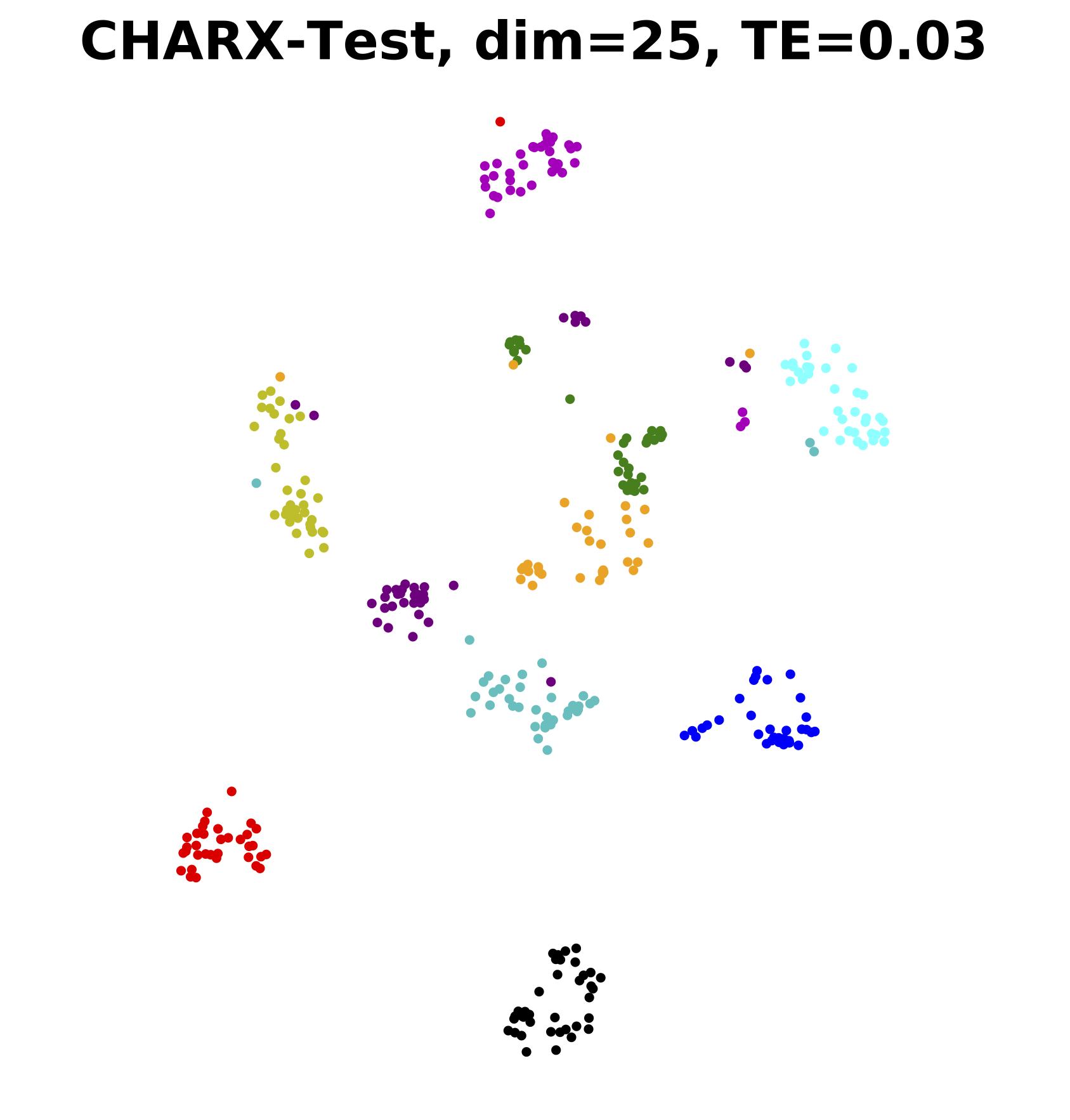}
     \end{subfigure}
     \caption{On the left, we show the embeddings generated by $\textrm{OENN}_\textrm{train}$ on the training set (visualized using t-SNE). On the right, we show the corresponding embeddings generated by $\textrm{OENN}_\textrm{test}$. The title of each figure reports the dataset used, the embedding dimension,  and the fraction of triplets that are violated (TE). Colors indicate the labels of the items (which are only used for visualization, not for training or testing). To generate these t-SNE embeddings, we set the perplexity parameter to $30$. }
     \label{fig:generalization}
\end{figure*}

In this section, we present the performance of our method on out of sample extensions. To conduct this experiment, we choose the datasets USPS and CHAR. The USPS dataset has already a pre-defined training and test set. In the case of CHAR, we create the test set by randomly splitting the data into a train (80\%) and a test set (20\%). Subsequently, we generate $2 n d \log n$ triplets by drawing points uniformly at random from the train set $X = \left \{ x_1, x_2, \ldots, x_n \right \}$ of items. These triplets are then used to train the training network $\textrm{OENN}_\textrm{train}$. Now we are going to study the out-of-sample extension. To this end, we randomly generate $2$ million triplets of items $(x^{*}_i, x_j, x_k)$, where $x^{*}_i$ is always chosen from the test set $X^* = \left \{ x^*_1, x^*_2, \ldots, x^*_k \right \}$ and $x_j, x_k$ are drawn from the train set. The input representation of an item $x^{*}_i$ from the test set is chosen as the binary encoding of the index $i$ in length $\ceil{\log n}$. The input representations for an item $x_j$ from the train set is chosen as the binary encoding of the index $j$ in length $\ceil{\log n}$. Note that, the same indices are allowed to appear in the test set as well as the training set since different EmbNet components are used to generate the embeddings of items from the train set and the test set. Let $T^*$ denote the set of all triplets containing items from $X^*$. The generalization network $\textrm{OENN}_\textrm{test}$ is then trained on $T^*$ (while keeping $\textrm{OENN}_\textrm{train}$ fixed as explained in the main paper). The results are demonstrated in Figure \ref{fig:generalization} by reporting both the fraction of triplets in $T^*$ that are not satisfied by $\textrm{OENN}_\textrm{test}$ and also by visualizing the output embeddings of both the training data and the test data in $2$ dimensions. The t-SNE algorithm is used to obtain the 2D visualization of the combined set of embeddings of both the train and the test items. The results clearly show that, in addition to satisfying all the triplets in $T^*$, the quality of the obtained embeddings is high: the train and test embeddings look pretty much the same, and the triplet error is very low. 


\vskip 0.2in
\bibliography{references}

\end{document}